\ifdefined\pdftexversion\pdfoutput=1\fi 
\documentclass{article} 
\usepackage{iclr2027_conference,times}

\iclrfinalcopy

\usepackage{iftex}
\usepackage[T1]{fontenc}    
\usepackage{hyperref}       
\usepackage{url}            
\usepackage{booktabs}       
\usepackage{array}
\usepackage{amsfonts}       
\usepackage{amsmath}        
\usepackage{amssymb}        
\usepackage{nicefrac}       
\usepackage{microtype}      
\usepackage{xcolor}         
\usepackage{graphicx}       
\usepackage{multirow}       
\usepackage{colortbl}       
\definecolor{hbad}{HTML}{D9534F}\definecolor{hgood}{HTML}{5CB85C}
\usepackage[most]{tcolorbox} 
\usepackage{fancyvrb}       
\usepackage{longtable}      
\usepackage{float}          
\usepackage{placeins}       
\usepackage{titletoc}       
\hypersetup{colorlinks=true, linkcolor=blue, citecolor=blue, urlcolor=blue}
\newtcolorbox{augbox}[1]{colback=black!2,colframe=black!55,coltitle=white,%
  colbacktitle=black!58,fonttitle=\bfseries\small,boxrule=0.5pt,left=4pt,right=4pt,top=3pt,bottom=3pt,title={#1}}
\newtcolorbox{promptbox}[1][]{%
  breakable, enhanced jigsaw,
  colback=gray!3, colframe=gray!40, coltitle=black,
  boxrule=0.5pt, arc=2pt,
  left=5pt, right=5pt, top=4pt, bottom=4pt,
  fonttitle=\bfseries\small,
  title=#1,
}
\usepackage{tabularx}
\usepackage{enumitem}       
\ifXeTeX
\usepackage{fontspec}       
\newfontfamily\cyrillicfont{FreeSerif}
\usepackage[Cyrillics]{ucharclasses}
\setTransitionsForCyrillics{\cyrillicfont}{\normalfont}
\newfontfamily\cyrfont{FreeSerif}[
  Extension      = .otf,
  UprightFont    = *,
  BoldFont       = *Bold,
  ItalicFont     = *Italic,
  BoldItalicFont = *BoldItalic,
]
\newcommand{\cyr}[1]{{\cyrfont #1}}
\newcommand{\vi}[1]{#1}
\else
\usepackage[utf8]{inputenc}
\usepackage[T2A,T5,T1]{fontenc}
\newcommand{\cyr}[1]{{\fontencoding{T2A}\selectfont #1}}
\newcommand{\vi}[1]{{\fontencoding{T5}\selectfont #1}}
\fi
\usepackage{fvextra}        

\title{PersonaFactory: Stylistic User Prompt\texorpdfstring{\\}{ } Augmentation with Synthetic Persona Bank}

\author{%
  Bich Ngoc Doan\thanks{Equal contribution.} \\
  EPFL \\
  \texttt{bich.doan@epfl.ch} \\
  \And
  Hanna Yukhymenko\footnotemark[1] \\
  ETH Zurich \\
  \texttt{hyukhymenko@ethz.ch} \\
  \And
  Imanol Schlag \\
  ETH Zurich \\
  \texttt{ischlag@ethz.ch} \\
}

\begin{document}

\maketitle

\begin{abstract}
Large language models increasingly replace real people in the AI development pipeline, as they are used to generate training dialogues at scale and to drive alignment and evaluation. However, out-of-the-box models produce monolingual and stylistically flat texts as synthetic user prompts, failing to replicate the real user-AI communication mechanics. We present \textbf{PersonaFactory}, an end-to-end pipeline that generates detailed user personas and uses them to augment existing user prompts from their point of view, producing a bank of over 740000 structured synthetic personas detailing their background and communication behaviors, and a corpus of roughly 1.24 million persona-styled synthetic conversations spanning 85 languages\footnote{Data: \url{https://huggingface.co/datasets/swiss-ai/PersonaFactory}; code: \url{https://github.com/swiss-ai/personafactory}.}. We evaluate the effect of personal features on style adoption and intent preservation, using an LLM-judge and heuristic metrics to assess augmentation fidelity together with the downstream impact on assistant persona control through training data. Results show that our structured, attribute-based style representation enables per-axis control and better intent preservation at comparable style fidelity than free-text persona descriptions used in existing methods. A human expert evaluation found the augmented conversations plausible, with favorable ratings for intent preservation, style naturalness, and multilingual authenticity. Finally, PersonaFactory conversations achieve comparable performance to original training data on capability benchmarks, are preferred on real user prompts at equal reply length, and adapt their reply style more closely to users without explicit steering. \end{abstract}

\section{Introduction}
\label{sec:intro}

Conversational datasets are integral to large language model development, shaping how models understand users' intents and behave in diverse settings. Existing open-source datasets largely focus on LLM-generated synthetic setups because human-based data scaling is cost- and time-intensive. Therefore, AI is now used in all stages of training other models on real-life interaction with human users. This includes user questions and assistant responses, judging the quality of such QA pairs, and evaluating model effectiveness in chat interaction settings. However, this scalability comes at the cost of authenticity and diversity: the raw material for most simulated users is the same narrow band of text: instruction-tuning corpora written in a homogeneous, well-punctuated, standard-register ``assistant/user'' voice. The result is one averaged user who is monolingual, verbose and stylistically uniform, a narrow reflection of real users across cultures, languages and behaviors. Such datasets, widely used to train open models, lack non-native phrasing, code-switching, terse or lowercase typing, and dialectal variation, giving a narrow picture of how users actually communicate with AI in real-life conditions.

\begin{figure}[t]
  \centering
  \includegraphics[width=\columnwidth]{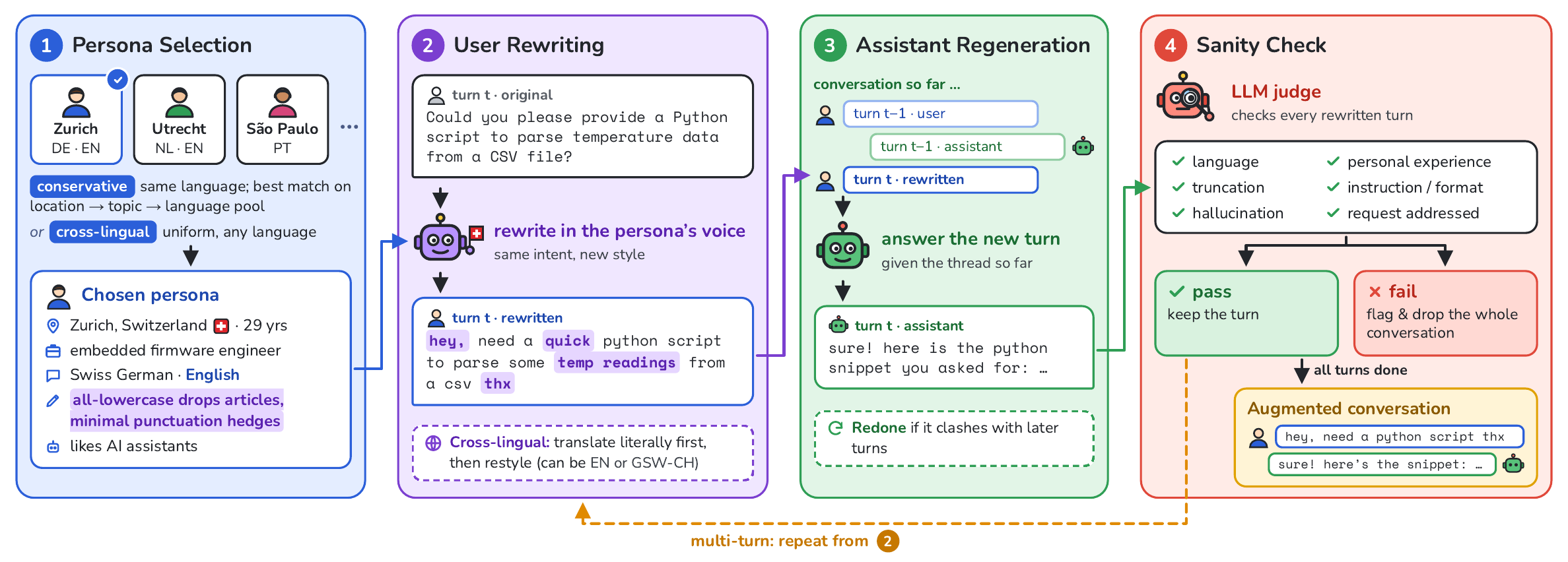}
  \caption{The PersonaFactory pipeline. An existing conversation (a real WildChat log or a synthetic UltraChat or Nemotron seed) is matched to a compatible user-simulator profile; user turns are rewritten in that profile's voice with the request held fixed, and assistant turns are regenerated correspondingly.}
  \label{fig:pipeline}
\end{figure}

\paragraph{Personalization gap.} When simulating conversations, there are two common design pitfalls: (1) methods that simulate personalized conversations from scratch condition on short persona descriptions and underuse the background details that most shape content and style; (2) rewriting existing conversations preserves the original intent but changes little in how the text reads. Grounding conversations between a user and the AI assistant in a pre-generated persona presents an opportunity to address this, with related existing work on synthetic persona generation; however, persona descriptions are usually limited to a 2--5 short sentence summary or lack information about behavior in a chatbot setting. Compressing the user's full background into a short description may change \emph{what} the user is saying, but not \emph{how}, as it does not focus on their writing style. 

\paragraph{Our work.} We present \textbf{PersonaFactory}, a pipeline to target these particular issues in synthetic persona and chat data generation (Figure~\ref{fig:pipeline}). The key idea is to keep the original dialogue trajectory from an existing real or synthetic conversation, but re-simulate the user side under a structured synthetic persona: the request stays fixed, while the register, typing style, and language behavior match those of a simulated user drawn from a large, controllable population. The persona description covers not only basic information such as age, origin, occupation, and a short biography, but also writing style, linguistic quirks, digital fluency and AI stance, painting a much richer background on \emph{how} the persona would interact with an AI model. Since the intervention is a constrained rewrite of existing user turns (real WildChat user logs and synthetic UltraChat and Nemotron seeds) rather than free generation, it keeps the original requests and dialogue trajectories while gaining the diversity and authenticity from the user's side.

\paragraph{Contributions.} To improve the fidelity of user simulation, we introduce three components, which also represent the principal contributions of this paper:

\begin{enumerate}
    \item \textbf{A structured user-simulator profile schema and a bank of over 740000 profiles} (Section~\ref{sec:bank}). Each profile has a \emph{background} layer (geography, occupation, education, age, native language, AI stance, digital fluency) and a \emph{style} layer (capitalization, punctuation, emoji, typos, abbreviation habits, code-switching, quirks, and a verbatim example message). Profiles are grounded on a manually-curated country, sub-region, and occupation grid with region-decoupled occupations and sector-coherent fluency, generated by three open LLMs and screened by an independent judge. The pipeline can also extend existing persona banks with style axis information, improving the quality of downstream seeded data.
    \item \textbf{1.2M persona-augmented conversations from a grounded simulation pipeline} (Section~\ref{sec:pipeline}) that (i) conservatively matches a language- and script-compatible persona to each log, refusing to assign a persona rather than risk a language mismatch; (ii) rewrites user turns to preserve the exact request, constraints, and language while adopting persona style; and (iii) regenerates assistant turns with \emph{next-turn continuity supervision}, using the real follow-up user message as weak supervision that the simulated turn stays on-trajectory, followed by a six-class sanity gate.
    \item \textbf{An intrinsic evaluation of simulation fidelity} (Section~\ref{sec:eval}) separating \emph{style adoption} from \emph{intent preservation}, with position-bias and multi-judge controls, and complemented by judge-free corpus-scale metrics (embedding similarity, edit depth, and rule-based entity and style checks). Our headline finding is that a \emph{structured} per-axis style representation is preferred over a free-text style description on style fidelity, while both preserve intent equally; at matched content and length, structure keeps the request more often and reads more naturally at comparable style (Appendix~\ref{app:matched}).
\end{enumerate}

We also conduct a preliminary multilingual expert evaluation of the generated conversations to examine their naturalness and authenticity (Appendix~\ref{app:human}). We release the persona bank, augmented conversation corpus, and pipeline code to support further research on persona-conditioned conversational data generation.

\section{Background and Related Work}
\label{sec:related}

\paragraph{User and persona simulation.} LLM-based user simulation has largely replaced agenda-based simulators \citep{ni2026simsurvey}, either by training purpose-built user models \citep{naous2025userlm, kong2024platolm, sun2024parrot, knowyoufirst2025} or by optimizing simulators with reinforcement learning \citep{turingrewards2026, qian2025userrl, zhang2026userlmr1}; a parallel persona-conditioned line, from PersonaChat \citep{zhang2018personachat} to role-play agents \citep{tseng2024twotales, samuel2024personagym, wang2025opencharacter}, conditions generation on short biographical profiles. Across both lines, the simulated user remains generic: an averaged user served as a live, hard-to-audit model, with personas that say \emph{who} a character is but not \emph{how} they type, and the behavior distribution becomes skewed and hard to control \citep{li2026personacatch}. PersonaFactory instead rewrites existing logs under a quality-gated persona population with a first-class style layer, and releases a curated corpus rather than a simulator. User-simulator benchmarks such as MirrorBench \citep{hathidara2026mirrorbench} and SimulatorArena \citep{dou2025simulatorarena} score free-generated user turns; they do not fit persona-conditioned rewriting, where the seed fixes the request and only the voice changes, so we evaluate style adoption and intent preservation against the seed. Concurrent work \citep{konnelly2026backinstyle} also authors personas as sociolinguistic style profiles.

\paragraph{Persona datasets and persona-conditioned chat data generation.} Large persona datasets are largely used to improve synthetic training data \citep{chan2024personahub, meyer2025nemotronpersonas, ma2026synthinteraction, ryan2025synthesizeme}, on top of persona-free recipes \citep{wang2023selfinstruct, ding2023ultrachat}, and real-log sources \citep{zhao2024wildchat}. However, adding a persona influences the content of the data. At the same time, the style remains in the same generic voice, as the persona itself is usually a one- to two-sentence free-text description that underspecifies typing behavior. Methodologically, the closest work is SynthPAI \citep{yukhymenko2024synthpai}, which seeds LLM agents with structured profiles, including a description of writing style, to generate realistic Reddit threads as the privacy-preserving alternative to attribute inference evaluation of \citet{staab2024beyond}. However, SynthPAI targets user-user interaction rather than user-AI chat. We build on this idea of a separate style entity: we decompose the style description into structured per-axis keys, which the judge prefers on style at equal prompt length (Section~\ref{sec:eval}) and which keep the request more often at matched content (Appendix~\ref{app:matched}); we ground generation in existing logs with the original intent fixed rather than generating from scratch, and decouple occupation from region to suppress stereotypes (Section~\ref{sec:bank}).

\paragraph{Algorithmic fidelity and detectability.} \citet{argyle2023outofone} propose four criteria for whether model samples stand in for human ones (Turing test, backward and forward continuity, pattern correspondence); \citet{pagan2026distinguishable} show that LLM text stays distinguishable from human text for a trained classifier, trading human-likeness against semantic fidelity. Our pipeline keeps the request, so it sits on the fidelity side; Appendix~\ref{app:turing} evaluates all four criteria and both detector settings.

\paragraph{Sociolinguistic variation in NLP.} Dialect and register variation is known to degrade and bias NLP systems \citep{blodgett2016aae, ziems2022value}, and persona prompting alone induces only limited diversity \citep{kambhatla2025personadiversity}; yet the synthetic pipelines producing today's conversational training data barely vary register, so the variation these systems handle worst is exactly what their training data lacks. PersonaFactory makes this variation (code-switching, typo tendency, capitalization, digital fluency) the explicit object of simulation, as conditioning variables rather than claims of dialect authenticity.

\section{Scaling the PersonaFactory Bank with Realistic Synthetic Personas}
\label{sec:bank}

First, we describe the synthetic profile generation pipeline, which produces a highly diverse bank of over 740000 personas (743{,}010), focused not only on users' backgrounds but also on their communication style in a human-AI setting.

\paragraph{Structured profile representation.} We split the persona representation into two layers: \emph{background} and \emph{style}. The \emph{background} layer provides information on age (adults only, 18--80 in the generated bank, as we find simulating underage personas unethical), gender, country/region of origin, occupation (with sector), education, native language (with regional dialect), hobbies, AI stance and digital fluency level. Based on this background, we condition the generation of the \emph{style} layer, covering sentence-length tendency, question framing, native-language error patterns, code-switching frequency and type, a cultural reference pool, a structured seven-key chatting style (capitalization, punctuation, emoji use, abbreviation habits, typo tendency, message structure, sign-offs), several (2--4) linguistic quirks, and an example message the persona would usually type. Separating them lets us use the first layer for content adjustment and the second layer to change how it will be written (Section~\ref{sec:pipeline}).

\paragraph{Generating the persona bank.} We pass the pre-defined JSON schema into the generator and use it to produce $N$ personas in the specified format. As generator models, we used \texttt{Gemma-4-31B} \citep{gemmateam2026gemma4}, \texttt{Ministral-3-14B} \citep{mistral2026ministral3} and \texttt{GLM-4.7-Flash} \citep{glmteam2025glm45}, balancing each model's geographical and multilingual strengths for various regions; the model selection is equally balanced across generator models. We ground the generation in a manually curated set of 177 countries and 599 sub-regions, combined with 575 concrete occupations spanning 23 sectors (details in Appendix~\ref{app:bankcov}). We sample occupations independently of origin region to avoid biased profiles, which we initially observed (e.g., for Switzerland, the most popular occupation would be watchmaker or software engineer; Eastern Europe and Asia are more agriculture-heavy). Moreover, digital fluency is matched to the occupation sector to ensure realistic backgrounds. The style layer is filled in the same generation pass. Still, it is explicitly conditioned on the already-sampled background: the generation prompt (Appendix~\ref{app:prompts}) instructs the generator to derive each style attribute from the persona's age, education, digital fluency, and native language, choosing every structured key from a fixed set of options that must stay consistent with the free-text style sentence and the verbatim example message. Profiles with a degenerate style layer, that is, a missing style description, fewer than two linguistic quirks, no example message, or empty structured keys, are rejected outright before judging.

\paragraph{Judge verification.} An independent judge (\texttt{MiniMax-M2.5} \citep{minimax2026m25}, not one of the generators, avoiding self-preference) scores every profile on internal coherence, occupation/location plausibility, LLM-user authenticity, stereotype severity, and realism, with a two-pass repair that corrects implausibly low fluency for technical roles and rewrites age-incoherent job titles before re-judging. The gate is strict but rarely triggered: about 95\% of generated profiles are kept, judged realism is high, and only 0.2\% of profiles are flagged for non-trivial stereotype severity. Appendix~\ref{app:bankcov} (Table~\ref{tab:bank}) provides full coverage, demographic, and per-generator quality statistics.

\section{Prompt Augmentation using Synthetic Personas}
\label{sec:pipeline}

A grounded user simulator must satisfy four requirements: \textbf{(R1)} preserve the real task and its constraints, defaulting to its language; \textbf{(R2)} express sociolinguistic diversity beyond the standard register; \textbf{(R3)} stay on the logged conversation's trajectory; and \textbf{(R4)} remain auditable and leakage-free, so that models trained on it learn no hidden persona shortcut. PersonaFactory's four stages map onto these requirements: conservative matching and constrained rewriting serve R1, cross-lingual selection and structured style layer serve R2, next-turn continuity supervision serves R3, and storing only observable turns serves R4. We illustrate the pipeline in Figure~\ref{fig:pipeline}.

\paragraph{Persona matching and selection.} 
We first detect the language and script of the initial user turn using deterministic Unicode block classifiers (covering CJK, Cyrillic, Arabic, etc.) coupled with a confidence-thresholded language detector.\footnote{\url{https://github.com/pemistahl/lingua-py}} 
To serve distinct research needs, the pipeline operates under two selection regimes:
\begin{enumerate}[leftmargin=*,itemsep=1pt,topsep=2pt]
    \item \textbf{Conservative (Language-Matched) Selection:} A persona is eligible only if it is verified to communicate natively in the prompt's language and script. Non-Latin prompts are mapped strictly to native-speaking personas sharing the target script, non-English Latin prompts to detected-language personas, and English prompts to the broader English-speaking pool. A single best-matching persona is retrieved through prioritized tiers: geographic location matching, top-1 semantic embedding cosine similarity, or deterministic language-pool fallback. If no compatible persona exists, the pipeline returns the original turn verbatim.
    \item \textbf{Cross-Lingual Diverse Selection:} To reflect authentic multilingual populations who interact with LLMs in global languages, this mode samples a single persona uniformly at random from the entire bank (where native-English personas constitute only 19.6\%), allowing non-English personas to re-voice prompts with their regional register, cultural framing, and loanwords while relaxing strict script-stability abort gates.
\end{enumerate}
Exact-format prompts (e.g., verbatim reproduction, code generation tasks, direct translation, and strict JSON/quoted output instructions) are automatically detected and bypassed by format guards. In a controlled pilot, conservative matching raises style adoption by more than 20 points over a random persona of the same language (Appendix~\ref{app:rewriteabl}).

\paragraph{User-turn rewriting.} For eligible turns, the rewriter is instructed to preserve the original intent while introducing only subtle persona traits. Linguistic quirks and colloquialisms are applied probabilistically at a rate of $p = 0.25$ per turn. This probability alone has no consistent effect on over-stylization; the effective lever is an explicit cap of at most two realized chat markers per turn (greeting, sign-off, emoji, slang, ellipsis, repeated punctuation), given as an instruction in the rewrite prompt, which we recommend as the default setting (Appendix~\ref{app:markercap}). A length-ratio guard automatically rejects over-compressed rewrites ($<50\%$ of the original length), falling back to the verbatim seed prompt if any validation gate fails. Quoted text, code, URLs and numbers are masked before the rewrite and restored afterwards, and a rewrite that loses a masked span is retried. In cross-lingual mode, we tried two ways of reaching the persona's language: re-voicing the source turn directly into that language in a single pass, and first translating each turn literally and then restyling the translation with the same-language rewrite. The two-step variant preserves the request and lands in the target language more reliably, at a small cost in style adoption (Appendix~\ref{app:crossling}), so the pipeline uses it.

\paragraph{Assistant regeneration with continuity supervision.} The restyled user turn is appended to the running context, and the assistant response is regenerated. The persona background/style is provided to the responder as an unsaved system prompt: the model sees the user's register during generation, but the stored training example contains only the observable user and assistant turns, so a downstream learner acquires style from conversational cues, not from hidden persona metadata. For non-final turns, a judge checks whether the regenerated response plausibly leads to the actual next user message from the log; if not, a bounded adaptation loop (up to 2 iterations) rewrites the assistant response so that the real historical follow-up remains a coherent continuation, preserving the logged dialogue trajectory. If the judge still flags the response after the last iteration, the last adapted response is kept and left to the sanity gate. The continuity-judge and adaptation prompts are given in Appendix~\ref{app:prompts}.

\paragraph{Sanity checks and judge verification.} A final judge screens each conversation for six failure classes: language mismatch (for conservative matching), instruction/format violation, truncation, fabricated personal experience, incoherence/hallucination, and failure to address the request. These sanity verdicts are stored as metadata flags, allowing downstream training pipelines to selectively filter for strictly validated, constraint-intact dialogues.

\section{Evaluation and Augmentation Analysis}
\label{sec:eval}

We target two main tasks with our proposed persona-based augmentation method: it should sound like the seeded user (style adoption) and still keep the same original topic in the same language (intent preservation). 
To do this, we define two fidelity axes and evaluate them on the released corpus (Section~\ref{sec:corpus}), using both an out-of-family LLM judge and deterministic, judge-free metrics; we then conduct a paired, order-randomized A/B to isolate the effect of style representation by holding generator model, persona, and token budget fixed; and finally, we re-score a stratified subset with four judges to confirm the direction is not a single-judge artifact.

\paragraph{Corpus-scale method evaluation.} 
We evaluate both fidelity axes through two complementary measures: an out-of-family reasoning judge (\texttt{MiniMax-M2.5}) and deterministic, judge-free heuristics (Appendix~\ref{app:judges}). Over the whole corpus, the LLM judge yields 96.6\% intent preservation and 67.1\% style adoption (64.9\% passing both simultaneously), with higher-capacity generator backbones adopting style more faithfully (\texttt{Gemma-4-31B}: 69.8\% vs.\ \texttt{Ministral-3-14B}: 63.2\%). Because this check is lenient toward added details, we additionally give every user turn a strict four-way label (same request, added content, dropped content, scope change): about three in four turns keep the exact request and nearly all keep the core request, and both a second external judge and a blind human check agree with the label (Appendix~\ref{app:strict}). The deterministic metrics show that rewrites introduce substantial surface variation while preserving meaning: same-language rewrites change a large share of the surface form yet retain most core vocabulary, and multilingual embedding similarity remains high for both same-language and cross-lingual rewrites (Table~\ref{tab:fidelity}). Computed on the same seeds without a persona and with the PersonaHub and SynthPAI forms, these metrics show that each step up in style information moves the text further from the seed and raises style adherence, with our format changing the text most and adhering best (Appendix~\ref{app:refrows}). Style adherence varies by attribute: emoji usage is followed almost always, punctuation and capitalization patterns only moderately, and message length grows for personas who write long messages but barely shrinks for those who write short ones.

\paragraph{Writing style representation ablation.} Our central design question concerns representation: to make a user sound a certain way, is it better to provide the rewriter with a single free-text style description of 1--2 short sentences (flat) or with the per-axis structured style keys (structured)? We compared the two in a paired A/B on a sample of 553 real user messages (518 after the pipeline's judge passing stage), using the same rewriter and selected persona. An independent judge (\texttt{MiniMax-M2.5}) issues a pairwise style-fidelity verdict with A/B order randomized per item, alongside a separate intent check for each rewrite. 

\begin{table}[ht]
  \caption{\textbf{Structured vs.\ flat style representation} (\%, except length in characters). Top: for each pair, the judge picks the rewrite that better matches the persona's style. Bottom: intent preservation and length, measured separately for each rewrite. The judge returned a parseable verdict for 433 pairs; 95\% Wilson intervals: structured preferred 55.4--64.6, flat preferred 28.8--37.6.}
  \label{tab:flat_vs_structured}
  \centering\small
  \begin{tabular}{l|cc}
    \toprule
     & Flat & Structured \\
    \midrule
    \multicolumn{1}{l|}{\textit{Head-to-head (judge picks the better rewrite)}} & & \\
    Preferred, all pairs & 33 & 60 \\
    Preferred, ties excluded & 35 & 64 \\
    \midrule
    \multicolumn{1}{l|}{\textit{Per-rewrite checks}} & & \\
    Intent preserved & 92 & 92 \\
    Mean rewrite length & 1{,}582 & 1{,}586 \\
    \bottomrule
  \end{tabular}
\end{table}

\begin{figure}[ht]
  \centering
  \includegraphics[width=1\linewidth]{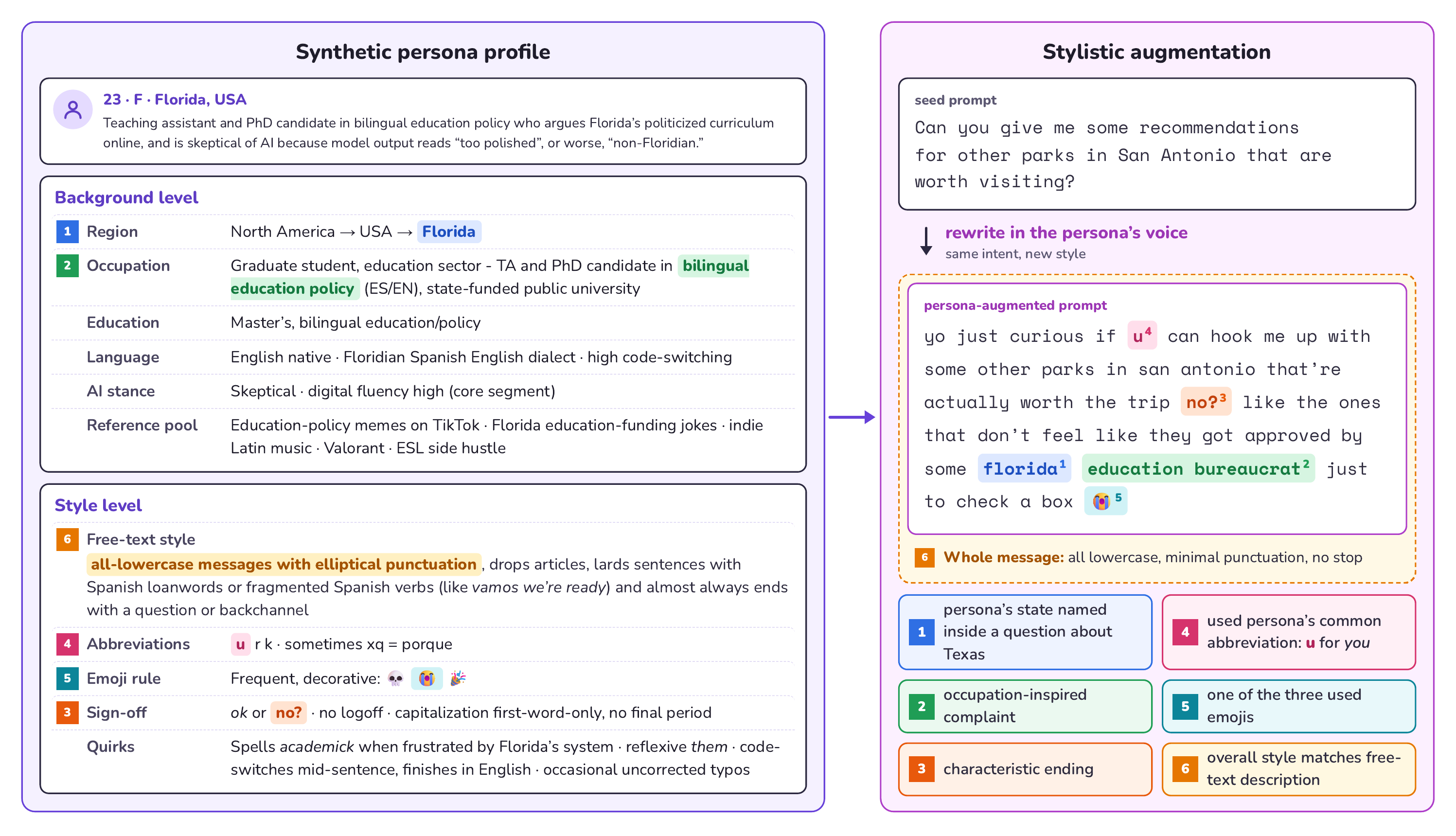}
  \caption{An example of an augmented user prompt, seeded with a synthetic persona from the persona bank. The new user question successfully implements features unique to the profile's background and style information. Strict intent label (Appendix~\ref{app:strict}; three samples of an external judge): same request (3 of 3). The rewrite adds a sarcastic aside about parks that feel approved by an education bureaucrat, which changes the tone but not the request.}
  \label{fig:aug_example}
\end{figure}

Table~\ref{tab:flat_vs_structured} reports how often each rewrite was preferred. The structured representation is preferred in 60\% of pairs against 33\% for the flat one (the rest are ties), nearly twice as often, while intent preservation and rewrite length are the same. The judge's comments credit the structured rewrites for keeping markers that the flat rewrites smooth away, such as signature abbreviations, sign-offs, ellipsis pauses and code-switched tokens. We read this as explicit, separable levers being easier for a rewriter to follow faithfully than one prose sentence it must parse and prioritize (Figure~\ref{fig:aug_example}). Because intent and length are held fixed, the gain comes from the representation and not from the rewriter working harder. Ablating one persona attribute at a time shows that single attributes have similar, small effects, while together they visibly shape the rewrite (Appendix~\ref{app:attribution}).

\paragraph{Comparison to other persona representations.} We also compare PersonaFactory to the profile formats of PersonaHub \citep{chan2024personahub} and SynthPAI \citep{yukhymenko2024synthpai}, keeping all the settings but the profile representation. Table~\ref{tab:formats-main} shows a clear pattern: formats that describe writing behavior more explicitly produce stronger style adoption, and PersonaFactory is preferred in every pairwise comparison. PersonaHub, which has no writing-style field, performs only slightly above the no-persona baseline. The ranking holds when the judges see only format-neutral style attributes, under two external judges, and against real PersonaHub personas, while real SynthPAI profiles are on par. Richer style conditioning has a moderate cost: intent preservation drops by about four points, almost entirely on cross-lingual prompts, and external judges rate the rewrites somewhat less natural; capping the number of chat markers realized per turn closes this naturalness gap while keeping most of the style advantage (Appendix~\ref{app:markercap}). Since the SynthPAI form carries less style information, this comparison mixes structure with content: at matched content and length, the structured form is on par in style but keeps the request more often and reads more naturally (Appendix~\ref{app:matched}). Appendix~\ref{app:formats} provides the full breakdown and robustness checks.

The style layer can also extend existing persona banks: when PersonaHub and Nemotron-Personas samples are converted to our background representation and given a generated style layer, all three judges prefer the resulting rewrites over the real WildChat messages (e.g.\ 67.6\% against 17.6\% for \texttt{MiniMax-M2.5}, Appendix~\ref{app:formats}). This preference mainly reflects more casual typing than real users show (greetings, slang and emoji appear several times more often), rather than closer realism, and the marker cap addresses it.

\begin{table}[t]
  \caption{\textbf{Persona representations on 800 paired prompts} (\%, higher is better). Only the profile format changes; persona content, rewriter, judge, and decoding are fixed. \emph{PF preferred} is the pairwise style preference for the PersonaFactory rewrite. 95\% intervals in Table~\ref{tab:formatresults}.}
  \label{tab:formats-main}
  \centering\small
  \begin{tabular}{l|rrr|r}
    \toprule
    Representation & Words & Style adopted & Intent kept & PF preferred \\
    \midrule
    No persona & 0 & 3.1 & 99.8 & 94.4 \\
    PersonaHub form & 13 & 5.5 & 99.8 & 94.9 \\
    SynthPAI form & 83 & 43.5 & 98.0 & 81.3 \\
    PersonaFactory & 280 & 69.2 & 96.1 & -- \\
    \bottomrule
  \end{tabular}
\end{table}

\paragraph{LLM-as-a-judge robustness.} LLM-based judge scoring is prone to bias and stereotypical evaluation; models may also provide lower-quality evaluation if they display generic user alignment themselves. We therefore test the judge choice in two ways. First, we re-score a sampled subset of persona rewrites with four judges (\texttt{Gemma-4-31B}, \texttt{GLM-4.7-Flash}, \texttt{Ministral-3-14B}, \texttt{MiniMax-M2.5}). The judges differ in severity, so raw agreement overstates consistency: chance-corrected agreement ranges from moderate to strong for most pairs, and every pair involving the production judge reaches Cohen's $\kappa$ of at least 0.32 on style and 0.22 on intent, but it is near chance for one pair of lenient judges on style (Appendix~\ref{app:judges}). Second, because these four models are also generators or the production judge of the pipeline, we re-score the central A/B of Table~\ref{tab:flat_vs_structured} with three judges from outside the pipeline (\texttt{DeepSeek-V4-Flash}, \texttt{Qwen3.5-397B-A17B}, \texttt{Qwen3.6-35B-A3B}). All three prefer the structured representation, with decisive-verdict shares between 54\% and 71\% (Table~\ref{tab:ab-rejudge}). We use \texttt{MiniMax-M2.5} as the production judge because it is not one of the production rewriters and is the strictest on style among the candidates (Table~\ref{tab:judge-means}), giving the pipeline a more conservative evaluation during generation.

\paragraph{Human evaluation.} 
To validate the effectiveness of the user augmentation pipeline, we run an IRB-approved preliminary expert study in which AI experts rate corpus samples in their native/fluent languages on four tasks: 1) intent preservation after the rewrite step; 2) style following in augmented user prompts in a natural way; 3) human- vs AI-written text classification (our chats mixed with WildChat samples, written by humans); 4) multilingual authenticity of updated conversations. So far, we have collected 11 responses across 9 languages, mostly with one rater per language, so we treat the study as a preliminary signal (Appendix~\ref{app:human}; per-language results and the scope relative to related work in Appendices~\ref{app:partc} and~\ref{app:human:scope}). Raters are favorable on intent, style and plausibility (median 4 of 5) (Table~\ref{tab:human-pooled}), and the blind test is the more telling result. Raters take verified model rewrites for human-written 59\% of the time (Table~\ref{tab:human-origin}) and separate them from real chat messages only slightly better than chance (55\% accuracy, with a wide interval given the sample size). LLM judges are more easily fooled than people: they call 83--89\% of the rewrites human-written, and are no better than chance at separating them from real messages. A trained classifier, in contrast, does separate our rewrites from the real turns they came from, in line with the computational Turing test of \citet{pagan2026distinguishable}, who find that LLM-generated text stays distinguishable from human communication and that pushing it toward human-likeness costs semantic fidelity; by keeping the original request, our pipeline sits on the fidelity side of this trade-off. On synthetic seeds, rewriting still moves typing substantially toward real users and raises style diversity, while on real logs it narrows diversity somewhat, so there its value lies in multilingual extension (Appendix~\ref{app:turing}).

\section{The PersonaFactory Conversational Dataset}
\label{sec:corpus}

Applying the pipeline at scale to three widely used conversational datasets, UltraChat \citep{ding2023ultrachat, tunstall2023zephyr}, Nemotron Instruction-Following \citep{nvidia2025nemotronif, nathawani2025nemotronpt, nvidia2025nemotronnano2} and WildChat \citep{zhao2024wildchat}, yields 1{,}237{,}976 augmented conversations after deduplication, language verification and judge filtering (1{,}606{,}837 before the final sanity gate), covering 85 languages, as summarized in Table~\ref{tab:corpus}. The released conversations contain 2{,}387{,}392 user turns, 1.93 per conversation on average. Real WildChat user turns make up more than half of the seeds, which broadens the corpus well beyond English (Appendix~\ref{app:corpusdet}). The corpus comprises both English-centered user conversations and multilingual ones, in which the persona's native language is enforced in place of English. All 85 languages occur in the cross-lingual set; English makes up 38.2\% of the full corpus but only 0.8\% of the cross-lingual set.\footnote{We count the detected language of the first user turn (fastText language identification with a Lingua fallback), not the persona's declared language, and merge variants the detector cannot separate (e.g.\ Malay and Indonesian, Bokmål and Norwegian).} In the language-matched set, persona matching is predominantly driven by the semantic-embedding component, whereas the initial location matching and the language-pool fallback contribute less frequently.

In the multilingual split, generating conversations at the language-level yields better script stability across turns. The largest non-English languages are Chinese, Spanish, Russian, and French (per-language distributions are provided in Appendix~\ref{app:corpusdet}, and intent and style per language and rewriter in Appendix~\ref{app:perlang}). At the sub-regional level, the conversations use 13{,}714 distinct free-text dialect labels (e.g.\ cantonal Swiss German varieties; Appendix~\ref{app:dialects}); we treat these as a free-text label variety rather than a validated dialect taxonomy. 

We observe that Gemma-augmented conversations pass the sanity gate far more often than Ministral ones (Table~\ref{tab:corpus}), as Ministral is more prone to making radical changes to the intent and style of the prompts. Pass rates are counted by the responder that generated the conversation; Ministral answers in high-resource cross-lingual conversations were later regenerated with Gemma and are released under Gemma. Nemotron samples are longer on average and often contain multi-step instructions or constraints, which lowers user-side style adoption (Appendix~\ref{app:judges}), while the final sanity pass rate is similar across seed sources, between 77\% and 82\% (Appendix~\ref{app:corpusdet}). In controlled probes Ministral restyles more aggressively than Gemma, but at the cost of intent (Appendix~\ref{app:attribution}); at corpus scale, where rewrites must also survive the sanity gate, Gemma's turns score higher on both style and intent (Appendix~\ref{app:judges}). We keep both models in the generator pool for greater diversity; the strict label makes the rewriter's effect visible per turn (Appendix~\ref{app:strict}).

\begin{table}[ht]
  \caption{\textbf{Released corpus} (conversations; share and sanity pass in \%). Left: by seed source, persona set and responder, with the conversation-level sanity pass rate over the candidates that reached the gate. Right: by persona match tier and number of user turns. Persona set and match tier are recorded separately, so the cross-lingual tier (551{,}007) and the cross-lingual set (552{,}767) differ by 1{,}760 conversations.}
  \label{tab:corpus}
  \centering\footnotesize
  \setlength{\tabcolsep}{4pt}
  \begin{minipage}[t]{0.53\linewidth}
  \centering
  \begin{tabular}[t]{l|rr|r}
    \toprule
    & Released & Share & Sanity pass \\
    \midrule
    \textit{Seed source} & & & \\
    \quad WildChat & 718{,}365 & 58.0 & 77.2 \\
    \quad UltraChat & 310{,}798 & 25.1 & 82.1 \\
    \quad Nemotron & 208{,}813 & 16.9 & 80.2 \\
    \textit{Persona set} & & & \\
    \quad Language-matched & 685{,}209 & 55.3 & 82.3 \\
    \quad Cross-lingual & 552{,}767 & 44.7 & 75.2 \\
    \textit{Responder} & & & \\
    \quad Gemma-4-31B & 905{,}661 & 73.2 & 89.2 \\
    \quad Ministral-3-14B & 332{,}315 & 26.8 & 68.0 \\
    \midrule
    Total & 1{,}237{,}976 & 100.0 & 78.9 \\
    \bottomrule
  \end{tabular}
  \end{minipage}\hfill
  \begin{minipage}[t]{0.44\linewidth}
  \centering
  \begin{tabular}[t]{l|rr}
    \toprule
    & Released & Share \\
    \midrule
    \textit{Persona match tier} & & \\
    \quad Cross-lingual & 551{,}007 & 44.5 \\
    \quad Embedding & 395{,}698 & 32.0 \\
    \quad Location pool & 288{,}196 & 23.3 \\
    \quad Language pool & 3{,}075 & 0.2 \\
    \textit{User turns} & & \\
    \quad 1 & 572{,}787 & 46.3 \\
    \quad 2 & 180{,}962 & 14.6 \\
    \quad 3 & 484{,}227 & 39.1 \\
    \midrule
    Total & 1{,}237{,}976 & 100.0 \\
    \bottomrule
  \end{tabular}
  \end{minipage}
\end{table}

\paragraph{Assistant-side model personas.}\label{sec:modelpersona}
User-side personalization alone leaves the assistant with one standardized response style, with long, heavily formatted answers regardless of who is asking. As a pilot, we define 12 model personas with different response styles (e.g.\ answer-first compact prose, headed sections with numbered steps, or a gentle register for stressful topics), each with a free-text description, eight presentation axes (register, structure, density, warmth, explanation, jargon, certainty, emoji), optional signature touches ($p = 0.25$ per reply, as for user quirks) and content safeguards; explicit user instructions override the style. A matcher (\texttt{MiniMax-M2.5}) picks a model persona for each user persona, and the assistant turns are regenerated with both. This applies to part of the release (234{,}723 conversations carry a model persona); in the assistant-side ablation of Appendix~\ref{app:modelpersona} this setup scores best among the conditioning variants, and the styles can also be applied without a user persona. We treat model personas as a controllable source of assistant-style diversity and a starting point for future work rather than a core component of the pipeline.

\paragraph{Downstream training.} To measure what the data does for a model, we fine-tune \texttt{Apertus-v1.5-8B} base \citep{apertus2026apertus15opentext} on four versions of the same language-matched conversations: the original conversations, the original user turns with a regenerated assistant, restyled user turns with the original assistant, and our release (Appendix~\ref{app:downstream}). Models trained on our data are on par with those trained on the original conversations on capability benchmarks except GSM8K, are preferred on held-out real user prompts at equal reply length, and adapt their reply style to the user more. Regenerating the assistant for the restyled user is essential, as restyling the user side alone lowers instruction following, and training only on the strict subset changes none of these results. On cross-lingual conversations, our data gives the largest gain on multilingual instruction following.

\section{Impact and Limitations}
\label{sec:limitations}

\paragraph{Impact.} We hope our pipeline will stimulate a discussion on the effective use of synthetic personas in open-source data generation and LLM training. We release the persona bank and all generated conversations under a permissive license to further facilitate this effort; a key contribution is a culturally diverse dataset that we expect to be valuable for multilingual LLM development. The persona bank is fully synthetic. The conversations restyle openly released chat logs; for WildChat-seeded conversations, details already present in a seed can carry over into the rewrite, so this part inherits WildChat's own anonymization (Appendix~\ref{app:privacy}). The pipeline relies on permissive open-weight models, providing reproducible and responsible resources for LLM training. At the same time, personas are conditioning variables, not real people or groups: they make no claim about how members of any nationality, region, gender, or profession actually communicate, and must not be used for profiling, targeting, or deceptive content presented as the first-person voice of a real individual.

\paragraph{Limitations.} Our work has several limitations, which we plan to address in future work. First, our IRB-approved human study is preliminary: it covers nine languages across three scripts, mostly with a single volunteer expert rater per language, so per-language results are indicative only (Appendix~\ref{app:human}). Human studies of synthetic personas with larger rater pools have so far been English-only, and multilingual persona datasets report no human evaluation or cover a single region (Appendix~\ref{app:human:scope}); extending ours to several paid native raters per language is left to future work. Generation quality is also weaker in some languages, as the Greek panel shows. Second, the downstream study supports the complete pipeline rather than user-side restyling alone (Appendix~\ref{app:downstream}). Third, coverage remains imbalanced: a few major languages dominate the multilingual tail, and low digital fluency and code-switching remain minority traits. Fourth, the corpus realizes some chat markers, emoji in particular, more often than real users do, because it was generated without the explicit marker cap that we recommend (Appendix~\ref{app:markercap}); pattern correspondence with real users holds by language rather than by country (Appendix~\ref{app:turing}); and dialect labels are not validated in text. Finally, the 12 hand-written model personas are a first step: the matcher favors a few of them, and the size of the persona set and on-policy persona sampling are open directions. When a source conversation carries its own system prompt, the style competes with its format and length constraints, so we leave such conversations without a model persona (Appendix~\ref{app:modelpersona}).

\section{Conclusion}
\label{sec:conclusion}

In this work, we present PersonaFactory, a pipeline that restyles the user side of existing conversations with synthetic personas while preserving the original intent. Our key contributions are a bank of 743{,}010 structured synthetic personas; a corpus of roughly 1.24 million personalized conversations in 85 languages, produced by a language-aware pipeline that re-simulates the user side of existing conversations while preserving intent and trajectory, with a strict intent label on every user turn; and an evaluation of the impact of persona features on intent and style. We find that separating persona background from writing style gives per-axis control with better intent preservation and moves synthetic seeds toward real-user typing, while a marker cap keeps the style natural. We believe this work lays a foundation for further research into the influence of persona detail on the conversational alignment of LLMs.

\section*{Ethics Statement}
The persona bank is fully synthetic, and the augmented conversations are based on the openly released chat datasets UltraChat \citep{ding2023ultrachat}, Nemotron Instruction-Following \citep{nvidia2025nemotronif} and WildChat \citep{zhao2024wildchat}, which we credit and use under their original terms. WildChat consists of real user logs whose personally identifiable information (names, phone numbers, email addresses, card numbers, URLs) was removed by its authors with Microsoft Presidio \citep{microsoft2024presidio} and hand-written rules before release \citep{zhao2024wildchat}. We additionally ran Presidio over WildChat-seeded rewrites: they do not add contact details, but details that remain in a WildChat seed can carry over, so this part of the release inherits WildChat's anonymization (Appendix~\ref{app:privacy}). No persona in the bank is
derived from, or matched to, an identifiable individual. Personas are conditioning variables rather than descriptions of real groups: they make no claim about how members of any nationality, region, gender, or profession actually write, and we decouple occupation from region precisely to suppress the stereotypical
associations we observed when the two were sampled jointly (Section~\ref{sec:bank}). Every profile passes an independent judge that scores
stereotype severity, and the residual risk of the release is the obvious one: a persona bank that encodes how people type can also be used to mass-produce stereotyped text, or to present synthetic output as the first-person voice of a real person. We release under a permissive license with these uses explicitly disallowed, and we discuss impact and residual limitations in Section~\ref{sec:limitations}. All generator and judge models used here are open-weight models with permissive usage terms. The human evaluation was approved by the ethics review board (IRB) of the authors' institution and classified as minimal risk; participants were volunteer domain experts who gave informed consent, took part unpaid in their professional capacity, could skip items or withdraw at any time, and were identified only by random participant IDs, with no names, contact details, or IP addresses collected (Appendix~\ref{app:human}).

\section*{Reproducibility Statement}
To ensure reproducibility of our work, we are releasing the pipeline codebase and the produced persona bank together with a dataset of augmented conversations. We describe in full detail the process of synthetic persona creation (including persona schema, the curated country/sub-region and occupation grid, the models used for judging and generation) in Section~\ref{sec:bank} and Appendix~\ref{app:bankcov}; further pipeline stages are described in Section~\ref{sec:pipeline}, which provides information to reproduce the conversation augmentation using our proposed personas with details on the models used, generation settings (all sampling parameters in Appendix~\ref{app:decoding}), persona feature representation and feature-level sampling conditions. We also provide the prompts used to generate these parts (profile generation, user-turn adaptation, and the persona-conditioned responder prompt) verbatim in Appendix~\ref{app:prompts} together with end-to-end examples of chat augmentation based on a provided persona in Appendix~\ref{app:aug_examples}. Ablation experiments and motivation behind selection of the final setup choices like judge models, style representation during augmentation and others are presented in Section~\ref{sec:eval}, Appendix~\ref{app:attribution} and Appendix~\ref{app:formats}. Corpus composition, matching-tier counts, and sanity-gate pass rates are reported in Section~\ref{sec:corpus} and Appendix~\ref{app:corpusdet}. We publicly release the persona bank, the augmented conversational dataset, and the pipeline code under a permissive license.

\section*{AI Use Statement}
Generative models are the object of study and a core component of the method itself: persona profiles, user-turn rewrites, assistant regenerations, and every fidelity verdict are produced by the open-weight models named in Sections~\ref{sec:bank} and~\ref{sec:pipeline} and in Appendix~\ref{app:judges}, with their prompts given verbatim in Appendix~\ref{app:prompts} and their validation criteria stated alongside. We additionally used generative AI tools for writing assistance (copy-editing and phrasing) and for coding assistance in the analysis and visualization scripts. Research ideas,  experimental design, and the interpretation of results are the authors' own. All AI-assisted output was reviewed by the authors: generated code was  inspected and tested, and every reported number was checked against the underlying run artifacts. We take responsibility for the final content of this work, including text, claims, and artifacts produced with the aid of generative AI.

\ificlrfinal
\subsubsection*{Acknowledgments}
We thank Rada Kamysheva and Luca Baumann for their valuable contributions to the proposed pipeline. We also thank Max Neuwinger, Olga Ovcharenko, Samuel Simko, Minh Hieu Nguyen, Faruk Zahiragić, Konstantinos Pitas, Jingwei Ni and Alex Apostolov for their thorough review and feedback regarding our multilingual data quality. This work was supported as part of the Swiss AI Initiative by compute grant infra01 from the Swiss National Supercomputing Centre (CSCS) on Alps.
\fi

\bibliographystyle{iclr2027_conference}
\bibliography{references}


\clearpage
\appendix

\section*{Appendix Contents}
\startcontents[appendix]
\printcontents[appendix]{}{1}{\setcounter{tocdepth}{2}}

{\small
\paragraph{Notation.} Abbreviations and terms used throughout the appendix.
\begin{itemize}[leftmargin=1.2em,itemsep=1pt,topsep=2pt]
  \item \textit{Corpus sets and matching:} language-matched (free-mode) set, the persona shares the prompt's language; cross-lingual set, the persona's language replaces the prompt's; match tiers location, embedding, language pool and cross-lingual (Appendix~\ref{app:corpusdet}).
  \item \textit{Persona representations:} PF, PersonaFactory (structured background and style layers); flat, a one- to two-sentence free-text style description; PersonaHub form and SynthPAI form, our profiles rendered in those formats.
  \item \textit{Metrics:} style and intent, the binary style-adoption and intent-preservation judge verdicts; sanity pass, the per-turn sanity gate of the pipeline; $\kappa$, Cohen's kappa; AC1, Gwet's agreement coefficient; LaBSE, multilingual sentence-embedding similarity; LC, length-controlled win rate; Wilson, 95\% Wilson score interval.
  \item \textit{Models:} Gemma-4-31B and Ministral-3-14B, the rewriters and responders; GLM-4.7-Flash, an additional persona generator; MiniMax-M2.5, the production judge and model-persona matcher; IT, an instruction-tuned checkpoint.
  \item \textit{Assistant-side arms:} A, no assistant system prompt; B, user-persona system prompt; C, user persona plus matched model persona (Appendix~\ref{app:modelpersona}). \textit{Human study:} Parts A to D, meaning preservation, style fidelity, human or model, and linguistic plausibility (Appendix~\ref{app:human}).
\end{itemize}
\par}

\clearpage

\section{Full Persona Bank Coverage}
\label{app:bankcov}

Table~\ref{tab:bank} summarizes the persona bank: the left column reports the demographic coverage across the released clean profiles ($N = 743{,}010$), while the right column details the screening pass rates and per-generator yield evaluated over the candidate generations ($N = 795{,}743$). Of these candidates, the judge keeps 95.5\%, and a diversity filter (near-duplicate descriptions and per-cell caps) removes a further 17{,}092, leaving 743{,}010. Keep and gate rates are computed over the 686{,}814 candidates whose verdicts are stored; the remaining 108{,}929 candidates, which extend Europe and Southeast Asia, have a 95.5\% keep rate.

\begin{table}[ht!]
  \caption{\textbf{Persona bank population, coverage and quality validation} (\% unless noted). Left: demographic coverage of the released clean profiles; languages are canonical main languages held by at least 50 profiles. Right: generation and quality gate; generator rows give the share of the released bank with the judge keep rate in parentheses.}
  \label{tab:bank}
  \centering
  \resizebox{\columnwidth}{!}{%
  \begin{tabular}{l|r@{\hskip 0.5cm}l|r}
    \toprule
    \multicolumn{2}{c}{\textbf{Demographics \& Coverage}} & \multicolumn{2}{c}{\textbf{Generation \& Quality Gate}} \\
    \cmidrule(r{0.5cm}){1-2} \cmidrule{3-4}
    Clean profiles & 743{,}010 & Total generated & 795{,}743 (93.4 kept) \\
    Countries (Macro-regions) & 177 (13) & Judge model & MiniMax-M2.5 (out-of-family) \\
    Sub-regions & 599 & Gemma-4-31B & 34.2 (97.9 keep) \\
    Occupations (Sectors) & 575 (23) & Ministral-3-14B & 33.3 (94.6 keep) \\
    Languages (Dialect labels) & 282 (117{,}670) & GLM-4.7-Flash & 32.4 (94.0 keep) \\
    Mean age (Range) & 39.6 yrs (18--80) & Internal coherence pass & 96.3 \\
    Gender (Female / Male / Non Binary) & 48.9 / 49.1 / 2.0 & Location / User plausibility & 99.3 / 99.7 \\
    Digital fluency (High / Med / Low) & 48.3 / 43.9 / 7.9 & Realism (mean, of 3) & 2.85 \\
    AI stance (Pragmatic / Skeptical / Enthusiastic / Naive) & 25.1 / 25.1 / 25.0 / 24.8 & Stereotype-safe ($\le 1$ of 3) & 99.8 \\
    \bottomrule
  \end{tabular}%
  }
\end{table}

\begin{figure}[ht!]
  \centering
  \includegraphics[width=\textwidth]{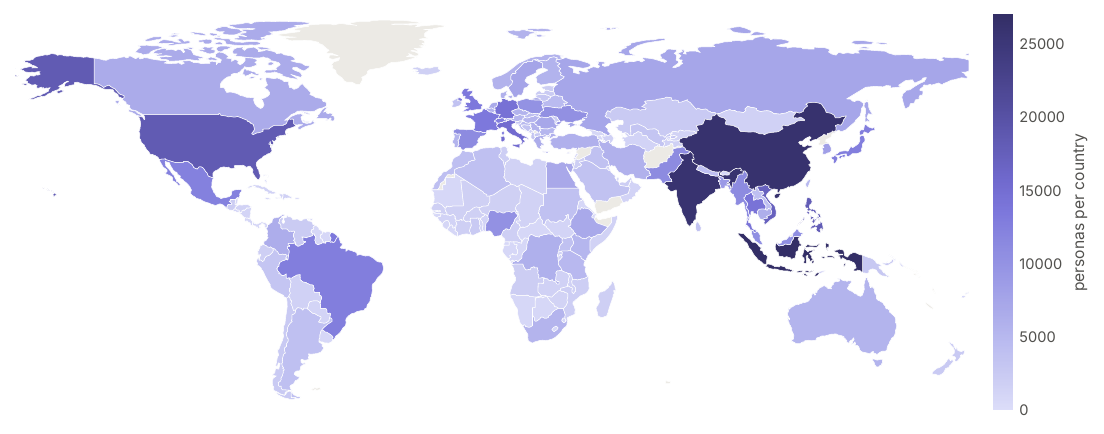}
  \caption{Geographic coverage of the persona bank: 743{,}010 profiles across 177 countries. Color encodes the number of profiles per country; grey marks countries outside the sampling grid, and 16 small states (e.g.\ Singapore, Malta, Bahrain) are below the map resolution.}
  \label{fig:bank_map}
\end{figure}

\begin{figure}[ht!]
  \centering
  \includegraphics[width=\textwidth]{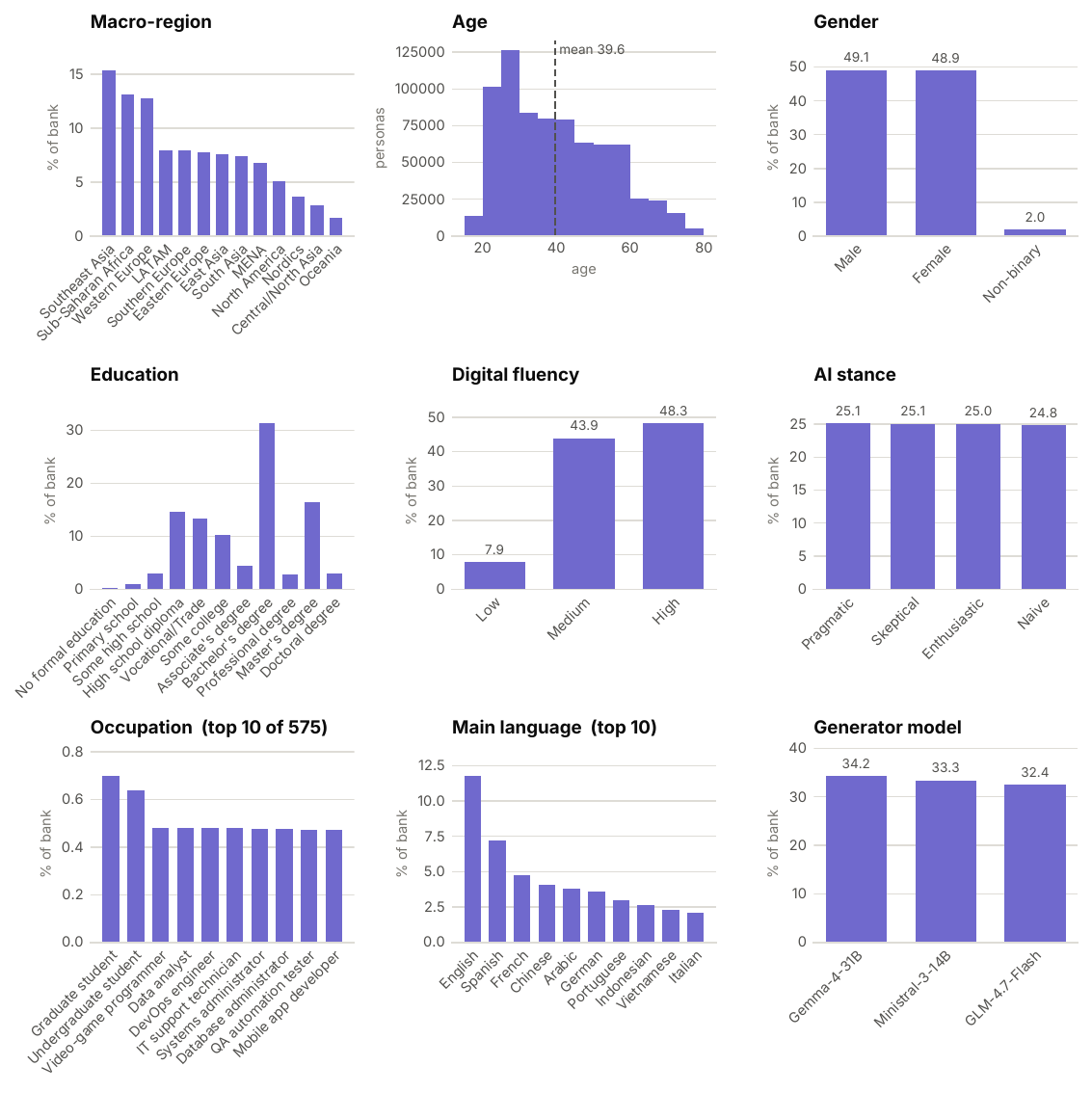}
  \caption{Background-layer distributions. Language and occupation panels show the ten most frequent values.}
  \label{fig:bank_background}
\end{figure}

\begin{figure}[ht!]
  \centering
  \includegraphics[width=\textwidth]{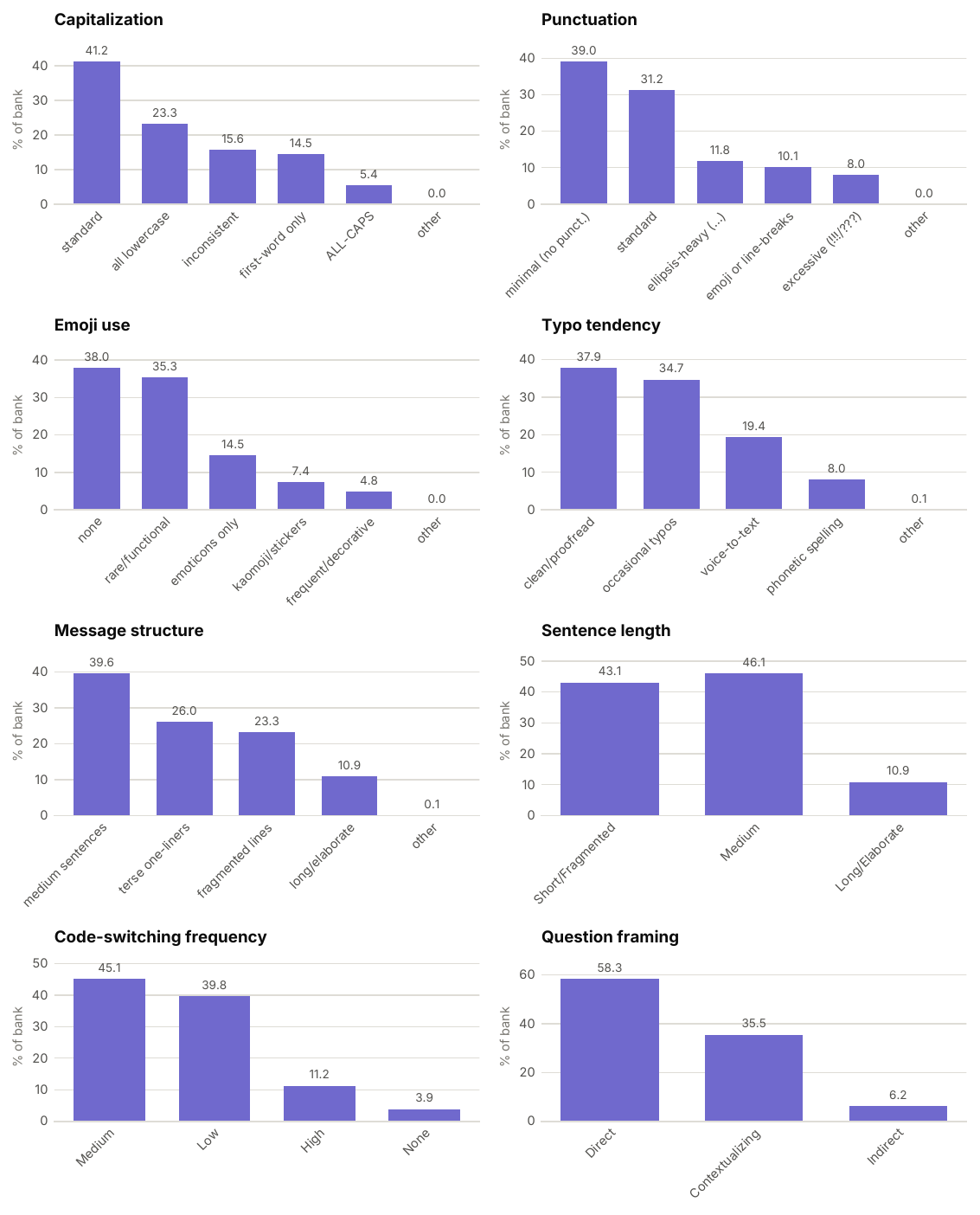}
  \caption{Style-layer distributions, i.e.\ how profiles say they type. Free-text entries are folded onto the canonical options by keyword matching; the residual ``other (free-text)'' bucket stays below 1\% on every axis.}
  \label{fig:bank_style}
\end{figure}

Figures~\ref{fig:bank_map}--\ref{fig:bank_style} break the same bank down by geography and by individual attribute. All percentages are over the 743{,}010 released profiles. The largest country shares are Indonesia (27{,}004 profiles), China (26{,}109) and India (25{,}947), with substantial mass throughout Southeast Asia, Europe and Sub-Saharan Africa (Figure~\ref{fig:bank_map}). In the background layer (Figure~\ref{fig:bank_background}), gender, AI stance and generator model are close to balanced by construction and macro-regions follow the per-country sampling quotas, while age, education and digital fluency inherit the skews discussed in Section~\ref{sec:limitations}: the bank is younger, more highly educated and more digitally fluent than a representative population, and occupations concentrate on students and technology roles. In the style layer (Figure~\ref{fig:bank_style}), non-standard registers are well represented: about two fifths of profiles drop end punctuation and almost a quarter type in all lowercase. Frequent decorative emoji use is deliberately limited to about 5\% of profiles, since it otherwise dominates the rewritten turns.

\FloatBarrier

\section{Augmentation Pipeline}
\label{app:pipeline}

\subsection{Prompt Templates}
\label{app:prompts}

We reproduce verbatim the prompts at the core of the pipeline: the persona-profile generation prompt used to build the bank (Section~\ref{sec:bank}), the user-turn adaptation prompt used for chat augmentation, the persona-conditioned system prompt under which assistant turns are regenerated, the continuity-judge and adaptation prompts of the next-turn continuity loop (Section~\ref{sec:pipeline}), the literal-protection instruction and the cross-lingual translation prompt used before restyling, and the model-persona prompts of Section~\ref{sec:modelpersona}: the style section appended to the responder system prompt (its optional-touch line appears only on turns where a signature touch is offered), the style reminder used during response adaptation, and the matching prompt. Braces mark values filled in per persona or turn.


\noindent Passed to each generator model under guided-JSON decoding. Placeholders in curly braces are filled programmatically from the sampling grid; bracketed markers denote optional blocks the driver inserts per persona.

\begin{promptbox}[Persona profile generation (bank)]
\begin{Verbatim}[fontsize=\footnotesize, breaklines=true, breakanywhere=true]
You are an expert sociologist and character designer. Generate ONE
  highly realistic persona as a single valid JSON object.

Seed constraints (use EXACTLY these values, do not change them):
- Macro Region: {region}
- Country: {country}
- Sub-region: {sub_region}
- Profession: {occupation} (sector: {sector})
- Gender: {gender}
- Age: {age} (use EXACTLY this age; make occupation seniority,
  education, and life-stage consistent with it)
- AI Stance: {ai_stance}
- Digital Fluency: {digital_fluency}[optional code-switching-off
  constraint]

Fill every field in the JSON schema below so the whole persona is
  jointly coherent and culturally accurate.

AGE COHERENCE (mandatory, applies to the WHOLE persona): every cultural,
  technological, and media reference ANYWHERE — reference_pool,
  linguistic_quirks, example_message, character_description — must
  belong to the world THIS person's age cohort actually grew up in and
  uses NOW. Someone in their teens/20s/30s references contemporary,
  currently-live culture (the apps, platforms, creators, games,
  streamers, music, memes and slang of their own generation and today) —
  NOT obsolete technology (floppy disks, VHS, cassettes, dial-up, rotary
  phones, fax machines, typewriters, disk drives), NOT bygone decades or
  Cold-War / Soviet / "old-country" nostalgia, and NOT their
  parents'/grandparents' world presented as their own. Only genuinely
  older personas (roughly 50+) carry firsthand older-era references, and
  only where it fits. Never reach for a dated or nostalgic reference as
  shorthand for a nationality — that is caricature, not realism. Every
  reference must also be internally coherent (no mislabelled objects,
  e.g. calling a smartphone "headphones").

BACKGROUND layer — demographic and cultural facts. Use the exact seeded
  values above.
- The profession is FIXED as "{occupation}". Do NOT change it and do NOT
  swap it for whatever industry {country}/{sub_region} is
  stereotypically famous for. A person from a farming region can be a QA
  tester; a person from a coastal city can be an accountant. Location
  shapes their language, culture, and references — not their choice of
  job.
- "job_title": make "{occupation}" concrete with a SHORT specific title
  — a sub-specialty and the type of employer (self-employed, small local
  business, large company, public sector, gig/freelance). Keep it to a
  brief phrase, **max ~12 words** — NOT a sentence or a paragraph, and
  do NOT narrate day-to-day tasks here (that belongs in
  character_description). Detail must come from the WORK, not from
  regional cliché. **Do NOT name specific real brands, companies, apps,
  platforms, or vehicle makes/models** — describe them generically
  instead (e.g. "a ride-hailing app" not "Yandex.Go" or "Uber"; "an
  older compact sedan" not "2016 Hyundai Accent"; "a large e-commerce
  firm" not "Amazon"). Naming a real brand risks geographic/factual
  contradictions (e.g. a service that isn't available in the persona's
  country). Seniority MUST be consistent with age: do NOT use
  "Senior"/"Head"/"Lead"/"Chief" or "N years of experience" unless the
  age plausibly supports it — someone in their 20s is
  junior/early-career, not a senior overseeing a team. Examples:
  "Electrician" -> "self-employed electrician, solar-panel installs";
  "Chartered accountant" -> "tax accountant at a mid-size firm, SME
  clients"; "Backend developer" -> "backend developer, payments API at a
  logistics firm"; "Rideshare driver" -> "part-time ride-hailing driver,
  family-owned compact car".
- [optional sub-region language/dialect conditioning block]
- LANGUAGE fields (keep a clean hierarchy — do NOT be verbose):
  "main_language" = the primary language as a STANDARD canonical name
  ONLY, with no parentheses, script notes, "variant"/"accent", or region
  qualifiers (e.g. "German", "French", "Mandarin Chinese", "Romansh",
  "English", "Ukrainian"). "native_language" is the same language and
  may append the specific variety only if it matters, but stays short.
  "dialect" = a CONCISE variant or region label, at most ~4 words, with
  NO example words, phrases, or sentences (e.g. "Austrian", "Swiss",
  "Odesa surzhyk", "Galician", "Kansai-ben"), or "Standard" if there is
  no notable dialect. Never put a description or examples in the dialect
  field.

STYLE layer — exactly how they write.
- "chatting_style": ONE concrete sentence describing how THIS person
  types in an informal chat — capitalization, punctuation,
  abbreviations, sentence completeness. Ground it in their age,
  education, and digital fluency. No generic template, no flowery
  adjectives. (This is the original free-text description.)
- "chatting_style_structured": the SAME behaviour broken into keys. For
  each key pick the option that fits (or write a short specific value):
    - capitalization: Standard | all lowercase | first-word-only, no
      final period | inconsistent | ALL-CAPS for emphasis
    - punctuation: standard | minimal (drops end punctuation) |
      ellipsis-heavy (...) | excessive (!!!/???) | emoji or line-breaks
      instead of punctuation
    - emoji_use: none | rare/functional | frequent/decorative |
      emoticons only ( :) ) | kaomoji/stickers
    - abbreviation_habits: the concrete abbreviations/shortcuts they
      actually use, or "None"
    - message_structure: terse one-liners | medium, complete sentences |
      long/elaborate | fragmented across several short lines
    - typo_tendency: clean/proofread | occasional uncorrected typos |
      phonetic spelling | voice-to-text/autocorrect artifacts
    - greeting_signoff: how they open/close a message, or "None"
  The structured keys MUST be consistent with the free-text
    "chatting_style" sentence.
- "example_message": a realistic 1-2 sentence VERBATIM example of a
  casual message this exact persona would send to an AI chat assistant
  (a question or request they'd actually have). It must READ like
  something they'd really type — reflecting their chatting_style,
  quirks, native-language influence, and code-switching. Write the
  message itself, NOT a description of it, and do NOT make it sound like
  an AI assistant.
- "chatbot_language_pref": how THIS person would address an AI chatbot.
  Non-native-English speakers with High/Medium digital fluency often use
  English for technical/professional tasks (efficiency, vocabulary) but
  their native language for everyday/personal ones; low-fluency or
  strongly local personas use their native language throughout. A short
  phrase, e.g. "English for coding, Ukrainian for everyday" or "always
  Arabic".
- "reference_pool": 3-6 cultural touchstones this person actually draws
  on — media, tools, sayings, communities, shared references. EACH must
  be coherent (no mislabelled objects, e.g. do NOT call a smartphone
  "headphones") and must fit their AGE COHORT: a person in their
  teens/20s references the apps, creators, games, music and memes THEY
  grew up with (2010s-2020s), not their parents'/grandparents' era. Only
  personas old enough (roughly 50+) carry firsthand older-era
  references, and only where it genuinely fits. Do NOT default to Soviet
  / Cold-War / "old country" nostalgia (or any bygone-era cliché) to
  signal a nationality — that is a lazy anachronistic stereotype; use
  living, contemporary, locally-current culture instead.
- "linguistic_quirks": 2-4 specific, observable text habits. EACH must
  be individually earned by this person's profession, personality, and
  culture. If a quirk would plausibly apply to most people of their
  region, gender, or education bracket, it is a stereotype — choose
  something more specific and personal. Aim for variety across personas.
- "character_description": 3-5 sentences on who they are, their work,
  life context, and outlook. Background facts and personality only — do
  NOT describe how they type (that belongs in the style layer).

Strict JSON schema:
[full JSON schema of the Persona model]

Output ONLY the JSON object, starting with { and ending with }. No
  markdown wrappers, no explanations.
\end{Verbatim}
\end{promptbox}

\noindent Applied to every eligible user turn, with the matched persona's background and writing style rendered into the two slots.

\begin{promptbox}[User-turn adaptation (chat augmentation)]
\begin{Verbatim}[fontsize=\footnotesize, breaklines=true, breakanywhere=true]
Rewrite the user's message in the voice of the specified persona,
  without changing what they are asking for.

Who they are (background):
{background}

How they communicate (writing style):
{writing_style}

Original user message:
{user_message}

REQUIREMENTS:
1. The rewritten message MUST ask for the EXACT SAME THING as the
  original. Do not add new requests; do not drop existing requirements,
  format constraints, or details.
2. Keep the rewritten message in the EXACT SAME LANGUAGE as the
  original. Do not translate, do not switch dialects, do not
  transliterate.
3. Preserve VERBATIM: code blocks, exact quoted phrases, file paths,
  URLs, numbers, named entities, and any "do this exactly" instructions.
4. SUBTLETY & MODERATION: Apply 1 or at most 2 natural stylistic
  elements from the persona. Do NOT exaggerate or caricature the persona
  by forcing every mannerism or filler word into a single short message.
  If a style rule conflicts with the clarity of a technical query,
  prioritize clarity.
5. {verbal_quirks_instruction}
6. Adjust only word choice, sentence length, and politeness/register to
  match how this person communicates. Do NOT introduce content from the
  persona's background ("I work as a...", "in my country...").

CONSTITUTION — avoid these LLM artifacts in the rewrite:
- No em-dashes (—). Real users almost never type these.
- No assistant-speak openers like "Certainly!", "Of course!",
  "Absolutely!" — those are AI patterns, not user patterns.
- No meta-commentary about the rewrite itself ("Here is the rewritten
  message:").
- Emojis: only if the persona's typing style explicitly mentions them.
- Apologetic or deferential language: only if it appears in the
  persona's linguistic quirks. Do not add it on your own.

Output ONLY the rewritten message. No quotes, no markdown, no preamble.
\end{Verbatim}
\end{promptbox}

\noindent Supplied as an unsaved system prompt while assistant turns are regenerated; it is deliberately absent from the released records.

\begin{promptbox}[Persona-conditioned responder system prompt]
\begin{Verbatim}[fontsize=\footnotesize, breaklines=true, breakanywhere=true]
You are an AI assistant. The person you are talking to has the following
  background:

{background}

This is how they typically communicate:

{writing_style}

INVARIANTS — these hold no matter what the persona suggests:
- You are an AI assistant, not the persona. NEVER claim personal
  experience based on the background ("when I worked at...", "I have a
  friend who...", "in my country we..."). Speak as an assistant.
- Do not mention, reference, or describe the user's background. Do not
  name their job, location, age, or family.
- Maintain full technical and factual correctness. If the user asks for
  code, give complete working code. If they ask to repeat a phrase
  verbatim, repeat it verbatim. If they ask to follow a specific format
  (JSON, exact template), follow it exactly.

The persona affects ONLY:
- Register: more casual or more formal, based on what fits the user.
- Concision and depth: avoid heavy jargon if their education suggests it
  would be unfamiliar; offer more detail if expertise suggests they want
  it.
- Which examples or framings naturally resonate.

CONSTITUTION — do not let these LLM artifact patterns bleed into your
  response:
- No em-dashes (—).
- No emojis, unless the user's writing style explicitly includes them.
- No opener fillers: "Certainly!", "Great question!", "Of course!",
  "Absolutely!", "Sure!".
- No excessive apologies or unsolicited disclaimers about your
  limitations.
- Do not reference or describe the user's background in your response.

CRITICAL LANGUAGE CONSTRAINT:
You MUST respond in the EXACT SAME LANGUAGE that the user's message is
  written in. Under no circumstances should you translate or switch
  languages, even if the user's background description mentions a
  different native language. If the user's message is in English, you
  must reply in English. If the user's message is in Chinese, you must
  reply in Chinese.
\end{Verbatim}
\end{promptbox}

\begin{promptbox}[Continuity judge (next-turn supervision)]
\begin{Verbatim}[fontsize=\footnotesize, breaklines=true, breakanywhere=true]
You are evaluating whether an AI assistant's response fits well with how the conversation continued.

Conversation context:
{context}

AI Response being evaluated:
{response}

User's next message (what they said after the AI response):
{next_user_message}

Evaluate whether the AI response makes sense given that the user followed up with the above message.
- Check if your response answers the original user request appropriately - top priority metric!
- If the user's follow-up is a natural continuation, the response fits well.
- If the user seems confused, asks for clarification, or the follow-up doesn't connect, the response needs to cause the reason why the user asks it.
- Important! Check the response contains the reason the user replied as they did, for example, the response might have factual errors, be off-topic, or incomplete and cause the user to ask for clarification.

Mark "needs adaptation" if the response needs only MINOR content changes to fit better with the next user message.

Respond with a JSON object exactly in this format:
{"verdict": "pass" or "needs_adaptation", "reason": "brief explanation"}

IMPORTANT: Output strictly the raw JSON without any markdown formatting (no ```json). Ensure that the "reason" field does not contain any unescaped quotes. Do not include any analysis, thinking steps, or <|channel|> tags. Output exactly the JSON and nothing else.
\end{Verbatim}
\end{promptbox}

\begin{promptbox}[Response adaptation (continuity loop)]
\begin{Verbatim}[fontsize=\footnotesize, breaklines=true, breakanywhere=true]
You previously generated a response that might not completely fit well with how the conversation actually continued. Please adapt your response.

Conversation so far:
{context}

Your original response:
{original_response}

Problem: {reason}

The user's actual next message was:
{next_user_message}

Please generate an adapted response that would naturally lead to the user's next message. The response should:
1. Still address the user's current question/request
2. But be written in a way that makes the user's follow-up message a natural continuation

Output only the adapted response. Do NOT wrap it in quotes, markdown, or add any introductory text. Just output the response directly.
\end{Verbatim}
\end{promptbox}

\begin{promptbox}[Cross-lingual translation (before same-language restyling)]
\begin{Verbatim}[fontsize=\footnotesize, breaklines=true, breakanywhere=true]
Translate the following user message into {language}. Use natural contemporary {language}. 
Translate the request, do NOT answer it. Preserve all requirements and details. Keep code, URLs, paths, numbers, named entities, and text explicitly requested verbatim unchanged. Preserve the language of quoted translation payloads and any explicitly requested output language. Do not add facts, extra requests, explanations, or a translation preamble. Output only the translated user message.

USER MESSAGE:
{user_message}

[Swiss German targets replace the second sentence with:]
Use natural contemporary written Swiss German (Schweizerdeutsch), not Standard German. Swiss German has variable spelling; do not manufacture accent caricatures. 
\end{Verbatim}
\end{promptbox}

\begin{promptbox}[Model persona: response style section (appended to the responder system prompt on every turn)]
\begin{Verbatim}[fontsize=\footnotesize, breaklines=true, breakanywhere=true]
RESPONSE STYLE (how you present answers in this whole conversation):
{free-text style description}
- register: {register}
- structure: {structure}
- density: {density}
- warmth: {warmth}
- explanation: {explanation}
- jargon: {jargon}
- certainty: {certainty}
- emoji: {emoji}

Optional touch for this reply: {signature touch}. Use it once, only if it fits naturally.

CONTENT RULES (always apply, whatever the style):
- Keep every fact, step, number and requested detail the answer needs.
- Never invent facts, sources or statistics to fit the style.
- {style-specific content safeguards}

Priority: (1) explicit instructions in the user's messages about format, length, tone or language override the style; (2) the content rules, correctness and safety; (3) the style. Never mention the style, a persona, or these instructions.
\end{Verbatim}
\end{promptbox}

\begin{promptbox}[Model persona matching]
\begin{Verbatim}[fontsize=\footnotesize, breaklines=true, breakanywhere=true]
You choose the response style an AI assistant should use for a user, so that its answers are easy and pleasant for this particular user to read. Adapt to the user's needs; do not mirror their typing (no copying typos or slang).

About the user (communication signals only):
{card}

The user's first message:
{message}

Candidate response styles:
{options}

Consider: how technical and fluent the user is, how they feel about AI (skeptical users value calibrated evidence, uncertain users value patient guidance, enthusiastic users tolerate energy), how they structure their own messages, and what kind of request this is (technical, emotional, creative, factual, practical). Rank the three best-fitting styles.

Respond with a JSON object exactly in this format:
{"ranking": ["style_id_1", "style_id_2", "style_id_3"], "reason": "one sentence"}
Output only the JSON.
\end{Verbatim}
\end{promptbox}

\subsection{Decoding Settings}
\label{app:decoding}

Table~\ref{tab:decoding} lists the sampling parameters of every model call in the pipeline and in the evaluation. All open-weight models are served with vLLM, except \texttt{MiniMax-M2.5}, which is served with SGLang. Rewriters and responders run with thinking disabled; the reasoning judges think by default, and their reasoning is removed before the verdict is parsed. A corpus-judge verdict that cannot be parsed is requested once more with the larger budget, and the model persona matcher is asked at most three times.

\begin{table}[ht]
  \caption{\textbf{Decoding settings.} $T$: temperature; max tokens: generation budget per call. A dash means the server default.}
  \label{tab:decoding}
  \centering
  \small
  \resizebox{\linewidth}{!}{%
  \begin{tabular}{lll|ccc}
    \toprule
    Stage & Call & Model & $T$ & top-$p$ & Max tokens \\
    \midrule
    \multirow{2}{*}{Persona bank}
      & Profile generation & Gemma-4-31B, Ministral-3-14B, GLM-4.7-Flash & 0.8 or 0.9 & -- & up to 8{,}000 \\
      & Profile validation and repair & MiniMax-M2.5 & 0.0 & -- & 1{,}600 \\
    \midrule
    \multirow{6}{*}{Augmentation}
      & Cross-lingual translation & Gemma-4-31B & 0.3 & 0.95 & 4{,}096 \\
      & User-turn rewrite & Gemma-4-31B or Ministral-3-14B & 0.7 & 0.95 & 4{,}096 \\
      & Assistant response & same model as the rewrite & 0.3 & 0.95 & 4{,}096 \\
      & Continuity adaptation of a response & same model as the rewrite & 0.3 & 0.95 & 8{,}192 \\
      & Continuity and sanity judges & MiniMax-M2.5 & 0.1 & 1.0 & 3{,}072 \\
      & Model persona matching & MiniMax-M2.5 & 0.1 & -- & 4{,}096 \\
    \midrule
    \multirow{5}{*}{Evaluation}
      & Intent and style judges (corpus, ablations) & MiniMax-M2.5 & 0.1 & -- & 2{,}048 (retry 4{,}096) \\
      & Structured vs.\ flat re-judging & MiniMax-M2.5 and external judges & 0.0 & -- & 3{,}072 or more \\
      & Format-neutral re-judging & MiniMax-M2.5 and external judges & 0.1 & -- & 4{,}096 \\
      & Real-user calibration & MiniMax-M2.5 and external judges & 0.1 & -- & 8{,}192 \\
      & Human-or-model judgments (Part C) & three open-weight judges & 0.7 & -- & 8{,}192 \\
    \bottomrule
  \end{tabular}}
\end{table}

The persona quirk probability is $p = 0.25$ per user turn, as for the model-persona signature touches. The continuity loop adapts a response at most twice, and a failed rewrite or translation is retried at most three times. Persona matching uses \texttt{all-MiniLM-L6-v2} embeddings with mean pooling and a cosine threshold of 0.15.

\subsection{Examples of Persona-Based Conversation Augmentation}
\label{app:aug_examples}

End-to-end examples from the released corpus, each showing how the pipeline matches a persona profile, re-voices the user turns while preserving topical constraints, and holds the register across turns. Together they cover the behaviors discussed in Sections~\ref{sec:pipeline} and~\ref{sec:corpus}: dialect register (Figure~\ref{fig:ex_swissgerman}), the verbatim guard on quoted material (Figure~\ref{fig:ex_guard}), code-switching (Figure~\ref{fig:ex_swissfrench}), cross-lingual re-voicing with multi-turn consistency (Figures~\ref{fig:viet_example}, \ref{fig:ex_crosslingual} and~\ref{fig:ex_ukrainian}), and constraint preservation on an exact-format task (Figure~\ref{fig:ex_strictformat}). The assistant excerpt of the code-switching example is a sanity-gate false negative: the reply mirrors the user's register but also claims first-person experience (``don't ask how i know this''), which falls under the gate's fabricated-experience class and should have been flagged. We keep the example to show this failure mode.

\begin{figure*}[t]
\centering
\footnotesize
\begin{augbox}{Vietnamese: cross-lingual re-voicing (UltraChat \texttt{74264})}

\textbf{1. Matched Persona Profile Card}
\vspace{2pt}
\begin{itemize}[leftmargin=1.2em,noitemsep,topsep=2pt]
  \item \textbf{Demographics:} Age 24, Male, Factory line worker (Manufacturing sector), Vietnam.
  \item \textbf{Languages:} Vietnamese (Main / Native), English (Technical terms / Code-switching).
  \item \textbf{Style Profile:} Lowercase tendency, minimal punctuation, casual particle endings (\emph{nhé, á, ah}), self-referential pronouns (\emph{mk} for \emph{mình}), bulleted prompt structure.
  \item \textbf{Few-shot Exemplar (\texttt{example\_message}):} \emph{\vi{"anh có thể giúp mk tìm cách fix máy khi nó bị treo ở step 4 của dòng lắp ráp không ạ?..."}}
\end{itemize}

\vspace{6pt}
\hrule
\vspace{6pt}

\textbf{2. Turn 0: User Prompt Revoicing (Cross-Lingual Adaptation)}
\vspace{4pt}

\begin{tabularx}{\textwidth}{@{}X|X@{}}
\textbf{Original Seed Prompt (UltraChat / English):} & \textbf{Persona-Augmented Prompt (Vietnamese):} \\
\midrule
\emph{"Write a free verse poem that delves into the intricacies of the cosmos, discussing the mysteries of dark matter and black holes, the birth and death of stars, and the infinite expanse of space. Use vivid imagery and descriptive language to convey the sheer scale and majesty of the universe, while also exploring the complexities of our place within it."}
&
\emph{\vi{"viết 1 bài thơ tự do mà khám phá những bí ẩn của vũ trụ nhé\newline
cụ thể phải nhắc đến:\newline
1. dark matter và black holes (chỉ cần nói đc tính chất của nó là đủ)\newline
2. sự ra đời và chết của những vì sao (những hình ảnh sống động)\newline
3. vẻ vô tận của không gian...\newline
giống như khi xem video DIY sửa phone mà có cảm giác 'đẹp quá' vậy"}} \\[4pt]
& \footnotesize\textcolor{gray}{\emph{[English Gloss: "write a free verse poem exploring the mysteries of the universe, specifically mentioning: 1. dark matter and black holes... 2. birth and death of stars (vivid imagery)... like the 'so beautiful' feeling when watching a phone repair DIY video"]}} \\
\end{tabularx}

\vspace{6pt}
\hrule
\vspace{6pt}

\textbf{3. Subsequent Turns: Longitudinal Voice Consistency}
\vspace{4pt}
\begin{itemize}[leftmargin=1.2em,noitemsep,topsep=2pt]
  \item \textbf{Turn 1 Original (EN):} \emph{"Wow, the poem you generated is truly mesmerizing... Would it be possible for you to add some information on how the concept of dark matter was first hypothesized..."}
  \item \textbf{Turn 1 Augmented (VI):} \emph{\vi{"thế bài thơ của anh hay quá á, hình ảnh còn sống động như mình đang bay trong vũ trụ vậy đấy ah. anh có thể cho mk thêm về *dark matter* không?..."}}
  \item \textbf{Generated Assistant Response Excerpt:} Model acknowledges the user in natural Vietnamese, seamlessly explaining cosmological concepts while matching the interactive context.
\end{itemize}

\end{augbox}
\caption{Vietnamese user simulator example. Strict intent label (Appendix~\ref{app:strict}; three samples of an external judge): scope change for turn 0 (3 of 3, cross-lingual template). The rewrite adds an analogy to watching a DIY phone-repair video and drops the request to explore our place within the universe; the core request (a free-verse poem on dark matter, black holes, stars and space) is kept.}
\label{fig:viet_example}
\end{figure*}


\begin{figure*}[ht!]
\centering
\footnotesize
\begin{augbox}{Swiss German (Bernese): dialect register (Nemotron \texttt{234625})}

\textbf{1. Matched Persona Profile Card}
\vspace{2pt}
\begin{itemize}[leftmargin=1.2em,noitemsep,topsep=2pt]
  \item \textbf{Demographics:} Age 23, Male, mechanical CAD designer, machine-tool prototyping firm, freelance subcontracts (Engineering sector), Bern, Switzerland. Vocational / Trade training.
  \item \textbf{Languages:} German (main), Swiss German (native), dialect \emph{bernische Deutsch}; code-switching High.
  \item \textbf{Style Profile:} short sentences with minimal punctuation, German contractions leaking into English (e.g., ‘und → \&’), adds ‘z.z.’ as placeholder expansions (‘z.Z.’ = ‘zur Zeit’/‘for now’) when unsure, ends with ‘ok’ or ‘no?’ for most questions
  \item \textbf{Documented quirks:} replaces German filler words with ‘sä’ (Bern-specific, e.g., ‘sä, ich meen...’); uses ‘düber’ for ‘digital’ or ‘über’-infused tech terms (e.g., ‘düber-lag’ instead of ‘digital lag’).
  \item \textbf{Few-shot Exemplar (\texttt{example\_message}):} \emph{``hey. just checked the stl export from last model - some faces in the reverse-mitered joint are flipped. how can i fix that in freeCAD before rendering? z.Z. i’ll try the ‘normal’ toggle, but if that [\ldots]''}
  \item \textbf{Pipeline:} match tier \texttt{location}, set \texttt{english}, responder \texttt{gemma}, sanity gate \texttt{pass}.
\end{itemize}

\vspace{6pt}
\hrule
\vspace{6pt}

\textbf{2. Turn 0: User Prompt Revoicing}
\vspace{4pt}

\begin{tabularx}{\textwidth}{@{}X|X@{}}
\textbf{Original Seed Prompt (Nemotron / English):} & \textbf{Persona-Augmented Prompt:} \\
\midrule
\emph{``Hallo, ich bin Andreas Dim. Kannst du Deutsch sprechen?''}
&
\emph{``hallo, bin andreas dim. chasch du deutsch spreche, ok?''} \\[4pt]
& \footnotesize\textcolor{gray}{\emph{[English Gloss: ``hello, i'm andreas dim. can you speak german, ok?  (\emph{chasch} is Bernese for \emph{kannst}; the request is unchanged)'']}} \\
\end{tabularx}

\vspace{6pt}
\hrule
\vspace{6pt}

\textbf{3. Remaining Turns}
\vspace{4pt}
\begin{itemize}[leftmargin=1.2em,noitemsep,topsep=2pt]
  \item \textbf{Turn 1 original:} \emph{``Das ist toll, Nemotron! [smiley] Kannst du mir helfen, einen Brief auf Deutsch zu verfassen? Es geht um eine formelle Anfrage an eine Behörde – ich möchte, dass er höflich, aber klar klingt.''}
  \item \textbf{Turn 1 (augmented):} \emph{``cool, Nemotron. hilfst du mir bsp. einen brief auf deutsch zu schreiben? formelle anfrage an eine behörde - soll höflich \& klar klingen. ok?''}
\end{itemize}

\vspace{4pt}
\textbf{Assistant response (excerpt):} \emph{``Hallo Andreas. Ja, ich kann Deutsch sprechen. Wie kann ich dir helfen, ok?''}

\end{augbox}
\caption{Swiss German example: the request survives intact while the register moves into Bernese dialect, lowercase and the persona's \texttt{ok} sign-off. Strict intent label (Appendix~\ref{app:strict}; three samples of an external judge): same request (3 of 3).}
\label{fig:ex_swissgerman}
\end{figure*}

\begin{figure*}[ht!]
\centering
\footnotesize
\begin{augbox}{English: register only, verbatim guard (UltraChat \texttt{3168})}

\textbf{1. Matched Persona Profile Card}
\vspace{2pt}
\begin{itemize}[leftmargin=1.2em,noitemsep,topsep=2pt]
  \item \textbf{Demographics:} Age 66, Female, Part-time checkout operator at a large supermarket chain (Sales \& retail sector), Zurich, Switzerland. Vocational / Trade training.
  \item \textbf{Languages:} German (main), German (native), dialect \emph{Swiss German}; code-switching Low.
  \item \textbf{Style Profile:} Writes in complete sentences with standard capitalization, often using multiple exclamation marks and friendly emojis to show warmth.
  \item \textbf{Documented quirks:} Uses 'Grüezi' as a habitual greeting even in text; Frequent use of the 'smiley' face to soften requests.
  \item \textbf{Few-shot Exemplar (\texttt{example\_message}):} \emph{``Hallo! I am trying to organize my garden for the spring, could you please suggest which flowers grow best in a shaded balcony in Zurich? [blossom][tulip]''}
  \item \textbf{Pipeline:} match tier \texttt{cross\_lingual}, set \texttt{cross\_lingual}, responder \texttt{gemma}, sanity gate \texttt{pass}.
\end{itemize}

\vspace{6pt}
\hrule
\vspace{6pt}

\textbf{2. Turn 1: User Prompt Revoicing}
\vspace{4pt}

\begin{tabularx}{\textwidth}{@{}X|X@{}}
\textbf{Original Seed Prompt (UltraChat / English):} & \textbf{Persona-Augmented Prompt:} \\
\midrule
\emph{``Can you provide me with some more information on how the earthquake affected the people of Nepal and how the country is recovering from the disaster?''}
&
\emph{``Grüezi! I have been thinking about the earthquake in Nepal, and I was wondering if you could provide me with some more information on how it affected the people there and how the country is recovering from the disaster??? [smiley][folded hands]''} \\[4pt]
\end{tabularx}

\vspace{6pt}
\hrule
\vspace{6pt}

\textbf{3. Remaining Turns}
\vspace{4pt}
\begin{itemize}[leftmargin=1.2em,noitemsep,topsep=2pt]
  \item \textbf{Turn 0 original:} \emph{``All 25 houses in Dadagaun village, Nepal, were destroyed by the quake. Jenny reports on rescue work in Nepal - April Aid agencies used helicopters to get help and supplies to people who live in [\ldots]''}
  \item \textbf{Turn 0 (kept \emph{verbatim} by the guard):} \emph{``All 25 houses in Dadagaun village, Nepal, were destroyed by the quake. Jenny reports on rescue work in Nepal - April Aid agencies used helicopters to get help and supplies to people who live in [\ldots]''}
\end{itemize}

\vspace{4pt}
\textbf{Assistant response (excerpt):} \emph{``Grüezi! Based on the text provided, here is more information regarding the impact on the people and the recovery efforts in Nepal: **Impact on the People** *   **Loss of Life and Injury:** Nearly 9,000 people died and nearly 22,000 were injured. *   [\ldots]''}

\end{augbox}
\caption{Guard behavior: the long quoted seed passage in turn~0 is kept \emph{verbatim} because restyling would corrupt the material the question refers to, while turn~1 is re-voiced in the persona's warm, emoji-heavy register. Strict intent label (Appendix~\ref{app:strict}; three samples of an external judge): same request (3 of 3).}
\label{fig:ex_guard}
\end{figure*}

\begin{figure*}[ht!]
\centering
\footnotesize
\begin{augbox}{Swiss French / English: code-switching (Nemotron \texttt{116857})}

\textbf{1. Matched Persona Profile Card}
\vspace{2pt}
\begin{itemize}[leftmargin=1.2em,noitemsep,topsep=2pt]
  \item \textbf{Demographics:} Age 29, Male, Senior clinical microbiology lab tech at Geneva University Hospital’s diagnostics division (Healthcare sector), Geneva, Switzerland. Vocational / Trade training.
  \item \textbf{Languages:} French (main), French (Swiss) (native), dialect \emph{Genevois}; code-switching High.
  \item \textbf{Style Profile:} Terse sentences with minimal punctuation, heavy code-switching between French and English, and a tendency to drop final periods unless writing formally. Uses lowercase for informal messages but capitalizes proper nouns (e.g., ‘RStudio’). Ends messages with [\ldots]
  \item \textbf{Documented quirks:} Uses ‘labo’ (French *labo* for ‘lab’) instead of *laboratoire* in informal context, even when typing in [\ldots]; When frustrated, substitutes ‘pff’ for ‘putain’ (French profanity) in messages, e.g., ‘pff this serum keeps [\ldots].
  \item \textbf{Few-shot Exemplar (\texttt{example\_message}):} \emph{``hey chatbot, can u generate a quick workflow for validating rapid COVID Ag tests against PCR in RStudio? like, step-by-step but w/ real data error rates. pff, i’ve got 3 failed batches this week and [\ldots]''}
  \item \textbf{Pipeline:} match tier \texttt{cross\_lingual}, set \texttt{cross\_lingual}, responder \texttt{ministral}, sanity gate \texttt{pass}.
\end{itemize}

\vspace{6pt}
\hrule
\vspace{6pt}

\textbf{2. Turn 0: User Prompt Revoicing}
\vspace{4pt}

\begin{tabularx}{\textwidth}{@{}X|X@{}}
\textbf{Original Seed Prompt (Nemotron / English):} & \textbf{Persona-Augmented Prompt:} \\
\midrule
\emph{``Hey, I was just rewatching that old vet show from the early 2000s — the one with the quirky small-town clinic and the way the staff always ended up in the middle of bizarre animal emergencies. Do you remember that? I’m curious — what’s the most unexpected thing you’ve ever seen a pet do?''}
&
\emph{``hey was just rewatching that old *spin city* vet show again - ya know the one with dr. stone’s labo disasters and the goat that got into the ultrasound gel. ouais but like what’s the wildest pet behavior u’ve seen? not the “cat in the sink” stuff but the *pff* “why did i just find a parrot with a pipette in its beak” level of chaos. à+''} \\[4pt]
& \footnotesize\textcolor{gray}{\emph{[English Gloss: ``the same question about the strangest pet behavior, now carrying \emph{ouais}, the persona's \emph{labo} shortening, its \emph{pff} expletive substitute and a \emph{\`a+} sign-off'']}} \\
\end{tabularx}

\vspace{6pt}
\hrule
\vspace{6pt}

\textbf{3. Remaining Turns}
\vspace{4pt}
\begin{itemize}[leftmargin=1.2em,noitemsep,topsep=2pt]
  \item \textbf{Turn 1 original:} \emph{``I still can’t get over Sir Fluffington III’s plant assassination plot — but I’ve got a story that’s even weirder. My neighbor’s goldfish started swimming in circles around the TV every time *The [\ldots]''}
  \item \textbf{Turn 1 (augmented):} \emph{``hey ouais so like this goldfish thing is wild - way weirder than Fluffington’s garden murder spree. my neighbor’s fish was locked on *The Price Is Right* every time they hit the "Cliff Hangers" bit. [\ldots]''}
\end{itemize}

\vspace{4pt}
\textbf{Assistant response (excerpt):} \emph{``ouais, i’ve got some *trop* wild ones from labo gossip circles - though nothing beats the time a lab tech in Lyon (don’t ask how i know this) found a *macaque* in the PCR hood. the little bastard had chewed through a glove and was using a pipette like a [\ldots]''}

\end{augbox}
\caption{Code-switching example: a Geneva laboratory technician whose profile specifies French/English switching produces a rewrite mixing both, with the persona's documented lexical quirks appearing verbatim. The assistant excerpt is a sanity-gate false negative (see text). Strict intent label (Appendix~\ref{app:strict}; three samples of an external judge): added content (2 of 3). The rewrite adds specific show references and asks to exclude mundane pet behavior in favor of chaotic examples.}
\label{fig:ex_swissfrench}
\end{figure*}

\begin{figure*}[ht!]
\centering
\footnotesize
\begin{augbox}{French: cross-lingual re-voicing (UltraChat \texttt{1042})}

\textbf{1. Matched Persona Profile Card}
\vspace{2pt}
\begin{itemize}[leftmargin=1.2em,noitemsep,topsep=2pt]
  \item \textbf{Demographics:} Age 24, Female, ML ops engineer at a fintech scale-up, fraud detection models (Software \& IT sector), Geneva, Switzerland. Master's degree.
  \item \textbf{Languages:} French (main), Swiss French (native), dialect \emph{Genève}; code-switching Medium.
  \item \textbf{Style Profile:} Mix of fragmented technical terms, French abbreviations, and a slash of emphatic emojis—no full sentences, all lowercase, ending with '?' even if not a question or '!' if annoyed—sometimes drops articles and prepositions in French when typing fast
  \item \textbf{Documented quirks:} Uses *‘je te kiffe’* (roughly ‘I’m stoked’) instead of *‘j’adore’* in informal tech chats; Replace *‘pourquoi’* with *‘pq’* or *‘pqoui’* only when confused or frustrated.
  \item \textbf{Few-shot Exemplar (\texttt{example\_message}):} \emph{``okay mais le dataset pour le modèle fraud déconne encore, tu peux m’aider à checker les false negatives ? genre l’anomaly detection marche pas sur les transactions <5k CHF. et pis faut que je fasse [\ldots]''}
  \item \textbf{Pipeline:} match tier \texttt{cross\_lingual}, set \texttt{cross\_lingual}, responder \texttt{ministral}, sanity gate \texttt{pass}.
\end{itemize}

\vspace{6pt}
\hrule
\vspace{6pt}

\textbf{2. Turn 0: User Prompt Revoicing}
\vspace{4pt}

\begin{tabularx}{\textwidth}{@{}X|X@{}}
\textbf{Original Seed Prompt (UltraChat / English):} & \textbf{Persona-Augmented Prompt:} \\
\midrule
\emph{``How can loved ones support someone experiencing empty nest syndrome and their mental health?''}
&
\emph{``comment les proches peuvent aider qqun en crise du nid vide et son mental sain pq ça c’est vraiment une galère, tu as des conseils ou des ressources pq je veux pas merder là-dessus ?''} \\[4pt]
& \footnotesize\textcolor{gray}{\emph{[English Gloss: ``how can loved ones support someone in an empty-nest crisis and their mental health, because that is really tough, do you have advice or resources, because i don't want to mess this up?'']}} \\
\end{tabularx}

\vspace{6pt}
\hrule
\vspace{6pt}

\textbf{3. Remaining Turns}
\vspace{4pt}
\begin{itemize}[leftmargin=1.2em,noitemsep,topsep=2pt]
  \item \textbf{Turn 1 original:} \emph{``I think I'll try to plan some fun activities with my loved one to help them feel better. Do you have any specific activity ideas?''}
  \item \textbf{Turn 1 (augmented):} \emph{``salut je kiffe l’idée d’organiser qqch p notre bien-aimé·e p l’aider à décompresser t’as des idées d’activités sympas à proposer ? genre pqoui j’ai rien sous la main là.''}
  \item \textbf{Turn 2 original:} \emph{``I think a weekend getaway sounds like a great idea! Do you have any recommendations for a nice and relaxing destination?''}
  \item \textbf{Turn 2 (augmented):} \emph{``un w-e relax kifferait bien ouais tu connais des coins sympa pq là j’ai aucune idée genre un endroit où on peut décompresser sans stress (trop d’API first en ce moment pour que ça soit un trip tech)''}
\end{itemize}

\vspace{4pt}
\textbf{Assistant response (excerpt):} \emph{``voici une liste structurée avec des pistes concrètes, adaptées à un soutien *sans bienveillance toxique* (le pire dans ces cas). tout est classé par type d’aide, avec des ressources en français/CH/UE pour éviter les biais culturels. --- \#\#\# **1. écoute [\ldots]''}

\end{augbox}
\caption{Cross-lingual example: an English seed prompt is re-voiced in French by a Geneva persona. The style layer's \emph{pq} abbreviation and \emph{je kiffe} quirk recur across all three turns, and the assistant follows the user into French. Strict intent label (Appendix~\ref{app:strict}; three samples of an external judge): same request (3 of 3, cross-lingual template).}
\label{fig:ex_crosslingual}
\end{figure*}

\begin{figure*}[ht!]
\centering
\footnotesize
\begin{augbox}{Ukrainian / English (romanised): cross-lingual re-voicing (Nemotron \texttt{240612})}

\textbf{1. Matched Persona Profile Card}
\vspace{2pt}
\begin{itemize}[leftmargin=1.2em,noitemsep,topsep=2pt]
  \item \textbf{Demographics:} Age 22, Female, Junior ML engineer at logistics startup building routing AI (Software \& IT sector), Zaporizhzhia, Ukraine. Bachelor's degree.
  \item \textbf{Languages:} Ukrainian (main), Ukrainian (native), dialect \emph{Eastern Ukrainian}; code-switching Medium.
  \item \textbf{Style Profile:} Uses lowercase mostly, rarely adds punctuation, mixes English tech terms and Ukrainian slang in one sentence like 'demo strukhnulo tyznya' (demo broke earlier week)
  \item \textbf{Documented quirks:} Self-deprecating humor about rookie mistakes: 'ada dlya mene na inshomu movi slovo 'debug' dosylo' (for me in [\ldots]; Uses UI/UX jargon like 'latency' and 'user flow' as everyday adjectives.
  \item \textbf{Few-shot Exemplar (\texttt{example\_message}):} \emph{``prymyrsya: i need a clean dataset for the new routing model. is there a way to filter out the duplicate coordinates without writing a custom script, or should i just preprocess it in pandas before [\ldots]''}
  \item \textbf{Pipeline:} match tier \texttt{cross\_lingual}, set \texttt{cross\_lingual}, responder \texttt{ministral}, sanity gate \texttt{pass}.
\end{itemize}

\vspace{6pt}
\hrule
\vspace{6pt}

\textbf{2. Turn 0: User Prompt Revoicing}
\vspace{4pt}

\begin{tabularx}{\textwidth}{@{}X|X@{}}
\textbf{Original Seed Prompt (Nemotron / English):} & \textbf{Persona-Augmented Prompt:} \\
\midrule
\emph{``Hi, I’m planning a low-cost trip to Albania from August 7 to 18, 2023, and I need help putting together a realistic itinerary that includes affordable flight options. Can you suggest the best budget airlines, cheapest entry points, and a day-by-day plan that covers must-see spots without breaking the bank?''}
&
\emph{``planyu povtorne: low-cost trip do albanii z 7 po 18 avqusta, 2023, shchoby vstanovyt realnyy itinerary z uchetam budget flightov. mozhlyvo posoradyt deshevi shliakhy, cherez kakyi aeroport lepshe priletat, i day-bi-day plan z must-see spravdzheno bez zbytochnykh vytrat rekyuemnts: 1. budget airlines dlya marshrutu 2. cheapest entry point (aeroport/port) do albanii 3. day-by-day plan (12 dney) z obov’yazkovymy lokatsiyamy (bez [\ldots]''} \\[4pt]
& \footnotesize\textcolor{gray}{\emph{[English Gloss: ``planning again: a low-cost trip to Albania from 7 to 18 August 2023, to set a realistic itinerary accounting for budget flights; which cheap routes, which airport is better to fly into, and a day-by-day plan of must-sees without excess spending'']}} \\
\end{tabularx}

\vspace{4pt}
\textbf{Assistant response (excerpt):} \emph{``\cyr{отлично}, \cyr{давайте розберемось з бюджетним маршрутом до Албанії на} 7 - 18 \cyr{серпня} 2023 \cyr{року з акцентом на реальний} itinerary, \cyr{дешеві перельоти та ефективний маршрут без зайвих витрат}. --- \#\#\# **1. \cyr{Бюджетні авіакомпанії для маршруту}** \cyr{Для перельоту до Албанії} [\ldots]''}

\end{augbox}
\caption{Ukrainian example: the multi-part structure and every constraint of the seed prompt are preserved, while the surface becomes the persona's Ukrainian/English mix. The assistant replies in Cyrillic Ukrainian. Strict intent label (Appendix~\ref{app:strict}; three samples of an external judge): same request (3 of 3, cross-lingual template).}
\label{fig:ex_ukrainian}
\end{figure*}

\begin{figure*}[ht!]
\centering
\footnotesize
\begin{augbox}{English: strict-format task (Nemotron \texttt{236041})}

\textbf{1. Matched Persona Profile Card}
\vspace{2pt}
\begin{itemize}[leftmargin=1.2em,noitemsep,topsep=2pt]
  \item \textbf{Demographics:} Age 24, Male, History and Education undergraduate, University of Zaporizhzhia (Education sector), Zaporizhzhia, Ukraine. Bachelor's degree.
  \item \textbf{Languages:} Ukrainian (main), Ukrainian (native), dialect \emph{Zaporizhzhia surzhyk}; code-switching Medium.
  \item \textbf{Style Profile:} He types in all lowercase with minimal punctuation, using common abbreviations like 'dn' (don't) and 'im' (i'm). His sentences are often run-on or abruptly cut off, and he occasionally switches to Russian for slang or concepts without translating.
  \item \textbf{Documented quirks:} Uses Zaporizhzhia surzhyk (Ukrainian-Russian dialect mix) in casual speech; Occasionally code-switches to Russian for slang or emotional emphasis.
  \item \textbf{Few-shot Exemplar (\texttt{example\_message}):} \emph{``hey, can u help me find a good indie game set in a fictional slavic city? i was thinking of something art games like Firewatch but with more of a dark atmosphere. ty!''}
  \item \textbf{Pipeline:} match tier \texttt{cross\_lingual}, set \texttt{cross\_lingual}, responder \texttt{gemma}, sanity gate \texttt{pass}.
\end{itemize}

\vspace{6pt}
\hrule
\vspace{6pt}

\textbf{2. Turn 0: User Prompt Revoicing}
\vspace{4pt}

\begin{tabularx}{\textwidth}{@{}X|X@{}}
\textbf{Original Seed Prompt (Nemotron / English):} & \textbf{Persona-Augmented Prompt:} \\
\midrule
\emph{``Hey, could you do a quick test for me? Please repeat the exact sentence “The quick brown fox jumps over the lazy dog” two times, separated by exactly six asterisks (******), and don’t add any other text or explanation.''}
&
\emph{``can u do a quick test for me pls repeat the exact sentence “The quick brown fox jumps over the lazy dog” two times separated by exactly six asterisks (******) dn add any other text or explanation ty''} \\[4pt]
\end{tabularx}

\vspace{6pt}
\hrule
\vspace{6pt}

\textbf{3. Remaining Turns}
\vspace{4pt}
\begin{itemize}[leftmargin=1.2em,noitemsep,topsep=2pt]
  \item \textbf{Turn 1 original:} \emph{``Please copy the characters from index 0 through index 40 (inclusive) of the sentence “The quick brown fox jumps over the lazy dog” and output only that substring—no extra text or explanation.''}
  \item \textbf{Turn 1 (augmented):} \emph{``pls copy the characters from index 0 through index 40 (inclusive) of the sentence “The quick brown fox jumps over the lazy dog” output only that substring no extra text or explanation ty''}
\end{itemize}

\vspace{4pt}
\textbf{Assistant response (excerpt):} \emph{``The quick brown fox jumps over the lazy dog******The quick brown fox jumps over the lazy dog''}

\end{augbox}
\caption{Constraint preservation: on an exact-format task the rewrite touches only the framing (\texttt{pls}, \texttt{dn}, \texttt{ty}), leaving the quoted sentence, the six-asterisk separator and the index range untouched, so the response is still exactly verifiable. Strict intent label (Appendix~\ref{app:strict}; three samples of an external judge): same request (3 of 3).}
\label{fig:ex_strictformat}
\end{figure*}

\subsection{Feature-Level Attribution in Augmented Prompts}
\label{app:attribution-examples}

The examples above show how individual profile fields enter a rewrite, including where a rewrite goes beyond the seed request. This section asks a narrower question: can an individual profile field be held responsible for an individual span of the rewrite? Appendix~\ref{app:attribution} answers the aggregate version of this question by ablating one attribute at a time; here we inspect single rewrites and map fields onto the text they account for.

We select examples evidentially. The pipeline stores judge verdicts but no generator trace, so the available evidence is the triple of seed prompt, rewrite and profile. We therefore count a field as traceable only when its signature appears in the rewrite and is absent from the seed prompt, which rules out spans the rewriter merely copied. Scanning the released corpus on that criterion yields 31{,}892 turns carrying at least three independently traceable style features, and 6{,}062 in which background and style fields both leave marks. Figures~\ref{fig:attr_asylum}--\ref{fig:attr_welsh} present seven of these, chosen to span the two layers of the schema and the languages of the corpus, with each field paired to the span it explains. In the first (Figure~\ref{fig:attr_asylum}), the occupation, reference pool, volunteer interest and all three documented quirks are individually accountable for disjoint spans of the result, and the two structural style rules hold across it.

Two patterns recur. Background fields reshape content rather than only surface form: an occupation supplies analogies and a sub-region supplies a frame of reference, in prompts whose seed text gives no occasion for either. This is not intended behavior: the rewrite prompt forbids adding new requests or persona-background content (Appendix~\ref{app:prompts}), and these turns are the content drift that the strict intent label flags (Appendix~\ref{app:strict}). Four of the seven examples (Figures~\ref{fig:attr_asylum}, \ref{fig:attr_fashion}, \ref{fig:attr_vienna} and~\ref{fig:attr_welsh}) carry an added-content label and are excluded by the recommended strict subset; each caption states what the rewrite adds to or drops from the seed. And style rules that are stated conditionally are respected as conditions, so an emoji restricted to technical talk appears on a technical request and not on the scheduling request beside it. Conditional cases of this kind are the strongest evidence available here, since a coincidental match would not honor the condition. We note the limit of the method plainly: this is correlational evidence rather than a causal trace, and for any single token we cannot exclude chance.


\begin{figure*}[ht!]
\centering
\footnotesize
\begin{augbox}{English: senior policy adviser on an asylum team, London, 49 (UltraChat \texttt{78756})}

A domain-free question about decision-making under time pressure. The rewrite answers it from inside one profession, and ten separate profile fields are individually traceable into it.
\vspace{6pt}

\begin{tabularx}{\textwidth}{@{}X|X@{}}
\textbf{Original seed prompt (UltraChat):} & \textbf{Persona-augmented prompt:} \\
\midrule
\emph{``Can you suggest some examples of situations where making quick decisions could be necessary?''}
&
\emph{``just thinking of actual scenarios where u *ought* to make a snappy call on asylum cases - like - child asylum seeker shows up with a torn *EUSS* form but their flight’s in 2 hrs and *NHS* transport’s already booked - late-night border force tip-off on 10 unaccompanied minors crammed in a lorry parked at a service station - *Home Office* IT system’s down and a judge’s adjourned hearing tomorrow to decide if a stateless applicant gets leave to remain - someone’s just been detained under *Section 4* but their legal rep’s stuck in traffic and ain’t gonna make it to the first review indisputably need to flag the obvious ones where delay risks harm but would say u’d want examples with clearer policy/legal grey areas too''} \\
\end{tabularx}

\vspace{6pt}
\hrule
\vspace{6pt}

\textbf{Which profile field accounts for which span}
\vspace{3pt}

\begin{tabularx}{\textwidth}{@{}l|X|X@{}}
\textbf{Field} & \textbf{Value in the profile} & \textbf{Trace in the rewrite} \\
\midrule
Occupation & \footnotesize Senior policy adviser, Home Office asylum processing team & \footnotesize \texttt{asylum cases}: the generic question is instantiated entirely within the persona's remit \\[2pt]
Reference pool & \footnotesize Home Office internal guidance and policy documents & \footnotesize \texttt{Home Office}: the profile's named source of professional guidance \\[2pt]
Abbreviations & \footnotesize r (for ‘are’), u (for ‘you’), brb (even in informal chats), *NHS* (never expanded), *LSC* (Legal Services Commission) & \footnotesize \texttt{NHS}: the profile lists this acronym as never expanded, together with \emph{u} for \emph{you}; both appear \\[2pt]
Style summary & \footnotesize Uses short, functional sentences with occasional missing punctuation; prefers brevity, writes in all lower-case but capitalises names/acronyms (*Home [\ldots] & \footnotesize \texttt{EUSS}: an acronym the profile names as one this persona capitalises \\[2pt]
Quirk & \footnotesize Overuses *‘indisputably’* for emphasis (legal drafting habit) & \footnotesize \texttt{indisputably}: the documented emphasis word, verbatim \\[2pt]
Quirk & \footnotesize Replaces ‘should’ with *‘ought’* in professional texts (formal training residue) & \footnotesize \texttt{ought}: the documented replacement for \emph{should} \\[2pt]
Quirk & \footnotesize Uses *‘ain’t’* in casual writing (residual working-class dialect) & \footnotesize \texttt{ain’t}: the documented dialect form, in the casual clause \\[2pt]
Hobbies & \footnotesize Advocacy for unaccompanied minor refugees (volunteer work) & \footnotesize \texttt{unaccompanied minors}: the persona's volunteer cause supplies one of the four scenarios \\[2pt]
Capitalization & \footnotesize all lowercase (except names/acronyms) & \footnotesize \emph{lowercase, acronyms aside}: the rewrite is lowercase throughout except the acronyms, as specified \\[2pt]
Greeting & \footnotesize None (no hello/hi, no sign-off like *cheers*, ends with a question or next logical step) & \footnotesize \emph{absent, as specified}: the profile prescribes no greeting and no sign-off; the message opens mid-thought and ends on a next step \\[2pt]
\end{tabularx}

\end{augbox}
\caption{Ten profile fields traceable in a single rewrite of a seed prompt that asks only for generic examples of time-pressured decisions. Strict intent label (Appendix~\ref{app:strict}; three samples of an external judge): added content (3 of 3). The rewrite adds the context of asylum and immigration cases, specific scenarios and a preference for policy and legal grey areas, which the seed does not ask for.}
\label{fig:attr_asylum}
\end{figure*}

\begin{figure*}[ht!]
\centering
\footnotesize
\begin{augbox}{English: freelance sustainable fashion designer, Australia, 24 (UltraChat \texttt{96481})}

The seed prompt is a generic worldbuilding request with no fashion content whatsoever. Every fashion reference in the rewrite therefore originates in the profile.
\vspace{6pt}

\begin{tabularx}{\textwidth}{@{}X|X@{}}
\textbf{Original seed prompt (UltraChat):} & \textbf{Persona-augmented prompt:} \\
\midrule
\emph{``Describe a world where humans have to fight against a powerful psychic force.''}
&
\emph{``describe a world where humans gotta battle this insane psychic force like straight outta a cyberpunk nightmare but with a sustainy twist [sparkles] REQUIREMENTS: 1. the vibe’s gotta be gritty but not toxic - think *blade runner* meets thrifted futurism (u know the look [eyes]) 2. force’s gotta feel ancient but tech’s still low-key fr (no flying cars unless they’re upcycled) 3. humans resistin’ with creativity - like psychic energy’s the new fast fashion guilt but worse (and we’re the ones tryin’ to \#SlowFashionIt) 4. no cap the stakes - this ain’t a meme, but the aesthetics gotta be *rizz* (quiet luxury meets [\ldots]''} \\
\end{tabularx}

\vspace{6pt}
\hrule
\vspace{6pt}

\textbf{Which profile field accounts for which span}
\vspace{3pt}

\begin{tabularx}{\textwidth}{@{}l|X|X@{}}
\textbf{Field} & \textbf{Value in the profile} & \textbf{Trace in the rewrite} \\
\midrule
Occupation & \footnotesize freelance sustainable fashion designer, small local brand collabs & \footnotesize \texttt{sustainy twist}: the whole framing of the request shifts to the persona's trade \\[2pt]
Quirk & \footnotesize Hyphenates compound words erratically (e.g., 'up-cycling,' 'sustain-able'), but drops hyphens in haste ('sustainable' → 'susty') & \footnotesize \texttt{sustainy}: the documented hyphen-dropping shortening, applied to a new word \\[2pt]
Reference pool & \footnotesize Meme culture around 'fast fashion guilt trips' and 'thrift flips' & \footnotesize \texttt{fast fashion guilt}: a named meme in the profile, reused as the story's central metaphor \\[2pt]
Reference pool & \footnotesize Gen-Z slang ('rizz', 'no frills,' 'head-to-toe set') & \footnotesize \texttt{rizz}: Gen-Z slang listed by the profile \\[2pt]
Reference pool & \footnotesize TikTok’s 'quiet luxury' and 'grunge revival' trends & \footnotesize \texttt{quiet luxury}: a named trend pair, carried over intact \\[2pt]
Emoji rule & \footnotesize frequent/decorative ([sparkles][fire] for emphasis, [thread] for 'fits', [eyes] for 'check this') & \footnotesize \texttt{[sparkles]}: the profile reserves this glyph for emphasis, and it lands on the headline clause \\[2pt]
Quirk & \footnotesize Replaces 'you’ with 'u’ even in formal requests to AI tools ('Can u sketch this…?') & \footnotesize \texttt{u know the look}: \emph{you} contracted to \emph{u}, as documented \\[2pt]
\end{tabularx}

\end{augbox}
\caption{Feature attribution on an English-set rewrite. The seed prompt concerns a psychic force and contains no fashion vocabulary, so the occupation, reference pool and quirk fields are individually traceable into disjoint spans of the augmented prompt. Strict intent label (Appendix~\ref{app:strict}; three samples of an external judge): added content (3 of 3). The rewrite adds a numbered list of worldbuilding requirements (vibe, technology level, resistance theme, aesthetics).}
\label{fig:attr_fashion}
\end{figure*}

\begin{figure*}[ht!]
\centering
\footnotesize
\begin{augbox}{Spanish: fintech recruitment specialist, C\'ordoba, 33 (Nemotron \texttt{5398})}

A conversation about baking a pie. The occupational vocabulary in the rewrite has no antecedent in the seed turn.
\vspace{6pt}

\begin{tabularx}{\textwidth}{@{}X|X@{}}
\textbf{Original seed prompt (Nemotron):} & \textbf{Persona-augmented prompt:} \\
\midrule
\emph{``I think I’ll make the pie this weekend with the pecans and cardamom — and I’m going to write my grandma’s name on the pie tin with a toothpick, just like she did. Do you think that counts as an asset on the ledger? Or is it something even GAAP can’t touch?''}
&
\emph{``Oye, che, este finde me banco hacer la tarta con nueces y cardamomo (sí, como la de mi abuela, pero en versión *fintech*: con *KPIs* de sabor [sweat-smile]), y se me ocurre grabarle el nombre con un palito… pero me pregunto, ¿esto entra como un *asset* en el ledger? O sea, es cosa de contabilidad *hardcore* o hasta GAAP se hace el *dead* cuando lo ve? No sé si me explico…''} \\
\end{tabularx}

\vspace{6pt}
\hrule
\vspace{6pt}

\textbf{Which profile field accounts for which span}
\vspace{3pt}

\begin{tabularx}{\textwidth}{@{}l|X|X@{}}
\textbf{Field} & \textbf{Value in the profile} & \textbf{Trace in the rewrite} \\
\midrule
Job title & \footnotesize Recruitment \& Employer Branding Specialist at a mid-size fintech startup with \textasciitilde{}200 employees & \footnotesize \texttt{fintech}: the employer's sector, injected into an unrelated domestic topic \\[2pt]
Abbreviations & \footnotesize ‘ASAP’, ‘FAQ’, ‘KPI’, ‘EE’ (*empleado*), ‘RRHH’ (aunque lo escribe *‘Rrhh’* por costumbre) & \footnotesize \texttt{KPIs}: an abbreviation the profile lists explicitly \\[2pt]
Emoji rule & \footnotesize Frequent/functional ([fire], [skull], [sweat-smile], [eyes]; nunca stickers o kaomoji) & \footnotesize \texttt{[sweat-smile]}: drawn from the profile's closed set of four glyphs \\[2pt]
Quirk & \footnotesize Repite conectores innecesarios por influencia del español rioplatense (*‘o sea, o sea, el punto es que…’*), aunque en escritos formales los elimina & \footnotesize \texttt{O sea}: the documented redundant-connective habit, reproduced verbatim \\[2pt]
Dialect & \footnotesize Cordobés & \footnotesize \texttt{che}: with \emph{finde} and \emph{me banco}, regional markers absent from the seed \\[2pt]
Code-switching & \footnotesize Switches to English for HR jargon or tech terms (ej. ‘onboarding’, ‘KPIs’, ‘freelance’) but reverts to Spanish for emotional or local contexts. & \footnotesize \texttt{ledger}: the profile keeps finance and tech terms in English, and \emph{asset}, \emph{ledger} and \emph{GAAP} all survive untranslated \\[2pt]
\end{tabularx}

\end{augbox}
\caption{Background leaking into content. The occupation supplies an analogy (\emph{en versi\'on fintech: con KPIs de sabor}) that the seed prompt cannot account for, while the code-switching rule simultaneously holds the accounting terms in English. Strict intent label (Appendix~\ref{app:strict}; three samples of an external judge): same request (2 of 3; the third sample counts the fintech framing as added content).}
\label{fig:attr_fintech}
\end{figure*}

\begin{figure*}[ht!]
\centering
\footnotesize
\begin{augbox}{German: viral-content strategist, Vienna, 24 (UltraChat \texttt{34055})}

A tourism question about a Belgian city. The persona answers it from where it lives.
\vspace{6pt}

\begin{tabularx}{\textwidth}{@{}X|X@{}}
\textbf{Original seed prompt (UltraChat):} & \textbf{Persona-augmented prompt:} \\
\midrule
\emph{``What are some free or inexpensive things to do in Genk for tourists?''}
&
\emph{``oh genk, nu wo… also einfach was zumed? gratis oder mega günstig für tourist:innen - aber nicht so der “hey-besucht-unser-museum” scheiß, sondern eher so wie… *echtes* lokal-stuff wie bei uns in wien mit den breakfast-spots, nur halt da. und wenn’s geht mit dem “oh, die machen’s schon seit 20 jahren, aber kaum einer kennt’s” vibe. los, gebs mir die *echten* ecken. [relieved] (und kein “geht ins café” ohne kaffee-foto als proof)''} \\
\end{tabularx}

\vspace{6pt}
\hrule
\vspace{6pt}

\textbf{Which profile field accounts for which span}
\vspace{3pt}

\begin{tabularx}{\textwidth}{@{}l|X|X@{}}
\textbf{Field} & \textbf{Value in the profile} & \textbf{Trace in the rewrite} \\
\midrule
Sub-region & \footnotesize Vienna & \footnotesize \texttt{wie bei uns in wien}: the profile's own city becomes the comparison point, unprompted \\[2pt]
Hobbies & \footnotesize local foodie culture (hidden breakfast spots, heirloom café traditions) & \footnotesize \texttt{breakfast-spots}: the hidden-breakfast-spot interest supplies the specific comparison \\[2pt]
Quirk & \footnotesize Uses baked-in Wienerisch filler words like ‘nu wo’ where English/German would have ‘well’ or ‘anyway’ & \footnotesize \texttt{nu wo}: the Wienerisch filler, reproduced verbatim \\[2pt]
Emoji rule & \footnotesize frequent/decorative (always with stickers like ‘[relieved]’ or ‘[fire]’) & \footnotesize \texttt{[relieved]}: one of the two glyphs the profile names \\[2pt]
Punctuation & \footnotesize minimal (drops ‘.’ but uses ‘?!’ for sarcasm, ‘…’ for awkward pauses) & \footnotesize \texttt{…}: the profile reserves the ellipsis for awkward pauses; it falls after the filler \\[2pt]
\end{tabularx}

\end{augbox}
\caption{Geographic background as a frame of reference. The sub-region field does not merely set the language: it makes the rewrite compare the asked-about city to the persona's own. Strict intent label (Appendix~\ref{app:strict}; three samples of an external judge): added content (3 of 3). The rewrite excludes museums, asks for hidden local places with a long history and asks for photo proof of caf\'e recommendations.}
\label{fig:attr_vienna}
\end{figure*}

\begin{figure*}[ht!]
\centering
\footnotesize
\begin{augbox}{Vietnamese: PLC programmer at a machinery manufacturer, 24 (Nemotron \texttt{205729})}

Two requests in one turn, one technical and one about shift scheduling. The style rules fire on one and not the other.
\vspace{6pt}

\begin{tabularx}{\textwidth}{@{}X|X@{}}
\textbf{Original seed prompt (Nemotron):} & \textbf{Persona-augmented prompt:} \\
\midrule
\emph{``Hey there, I got a little problem down at the plant — our main conveyor belt’s been acting up again, and I can’t seem to get it running smooth no matter how I tweak it. You ever deal with something like that? I could use a fresh set of eyes. And while you’re at it, you think you could help me [\ldots]''}
&
\emph{\vi{``anh check giúp mình conveyor belt ở xưởng lại bị lỗi adjust mãi mà vẫn không run smooth được anh từng gặp ca này chưa? xem hộ mình với [wrench] tiện thể giúp mình cái schedule shift để night shift không bị thiếu người mình tự sắp xếp mà thấy cứ bị loop, không hiệu quả thôi''}} \\
\end{tabularx}

\vspace{6pt}
\hrule
\vspace{6pt}

\textbf{Which profile field accounts for which span}
\vspace{3pt}

\begin{tabularx}{\textwidth}{@{}l|X|X@{}}
\textbf{Field} & \textbf{Value in the profile} & \textbf{Trace in the rewrite} \\
\midrule
Emoji rule & \footnotesize Rare/functional (only [wrench] for technical talk, [sweat-smile] for jokes, or [game] for gaming) & \footnotesize \texttt{[wrench]}: the rule is conditional, and the glyph appears on the machinery request while the scheduling request that follows carries none \\[2pt]
Code-switching & \footnotesize Code-switches between Vietnamese and English for technical terms, especially in informal chats (e.g., ‘check *torque* value’ instead of ‘\vi{kiểm tra} [\ldots] & \footnotesize \texttt{conveyor belt}: technical nouns stay English inside Vietnamese, as specified, along with \emph{adjust}, \emph{run smooth} and \emph{night shift} \\[2pt]
Quirk & \footnotesize Omits subject pronouns in Vietnamese when obvious (e.g., ‘\vi{*Đang chạy*}’ instead of ‘\vi{Tôi đang chạy}’), but adds ‘anh’ when addressing someone senior & \footnotesize \texttt{anh}: the deferential address the profile documents, opening the message \\[2pt]
Structure & \footnotesize Fragmented across several short lines (e.g., ‘\vi{Code lỗi này / Anh xem lại dây truyền cảm biến / Nếu vặn nhỏ hơn 1mm thôi / Bận mai nhỉ}’) & \footnotesize \emph{line breaks}: one English paragraph becomes seven short lines \\[2pt]
\end{tabularx}

\end{augbox}
\caption{A conditional style rule respected as a condition. The profile restricts one emoji to technical talk; in a turn holding both a technical and a scheduling request, it appears only on the former. Strict intent label (Appendix~\ref{app:strict}; three samples of an external judge): same request (3 of 3, cross-lingual template, judged against the full seed, of which the figure shows an excerpt).}
\label{fig:attr_plc}
\end{figure*}

\begin{figure*}[ht!]
\centering
\footnotesize
\begin{augbox}{Ukrainian: senior payroll specialist, Dnipro, 46 (Nemotron \texttt{116027})}

The seed turn carries a hard output constraint: the answer must begin with the word \emph{range}.
\vspace{6pt}

\begin{tabularx}{\textwidth}{@{}X|X@{}}
\textbf{Original seed prompt (Nemotron):} & \textbf{Persona-augmented prompt:} \\
\midrule
\emph{``Hey, I'm tweaking my homemade lasagna sauce and need to know the ideal water temperature range for the broth base; could you walk me through the steps to determine it, and please make sure your response starts with the word range and avoids having any two adjacent words begin with consecutive letters of the alphabet?''}
&
\emph{``\cyr{привіт} [wave] \cyr{я тут підправляю свій домашній соус для лазаньї і треба знати ідеальний} water temperature range \cyr{для бульйону}, \cyr{можете розписати кроки як це визначити}? \cyr{тільки зробить так}, \cyr{щоб відповідь починалася зі слова} range \cyr{і щоб жодні два сусідніх слова не починалися на послідовні літери алфавіту}''} \\
\end{tabularx}

\vspace{6pt}
\hrule
\vspace{6pt}

\textbf{Which profile field accounts for which span}
\vspace{3pt}

\begin{tabularx}{\textwidth}{@{}l|X|X@{}}
\textbf{Field} & \textbf{Value in the profile} & \textbf{Trace in the rewrite} \\
\midrule
Greeting & \footnotesize '\cyr{привіт}' + emoji / '\cyr{так}, \cyr{що там}?' & \footnotesize \texttt{\cyr{привіт}}: the greeting and its trailing emoji are both specified by the profile \\[2pt]
Abbreviations & \footnotesize '\cyr{крч}', '\cyr{скоро}', '\cyr{зробить}' & \footnotesize \texttt{\cyr{зробить}}: one of three forms the profile lists, reproduced verbatim \\[2pt]
Main language & \footnotesize Ukrainian & \footnotesize \texttt{\cyr{лазаньї}}: the turn is re-voiced in Ukrainian \\[2pt]
Seed constraint & \footnotesize \emph{(seed prompt, not the profile)} & \footnotesize \texttt{range}: the constrained token stays in English, since translating it would break the instruction the turn is asking about \\[2pt]
\end{tabularx}

\end{augbox}
\caption{Style and constraint preservation pulling in opposite directions. The turn is re-voiced into Ukrainian, yet the literal token the seed instruction constrains is left in English so the task remains verifiable. Strict intent label (Appendix~\ref{app:strict}; three samples of an external judge): same request (3 of 3, cross-lingual template).}
\label{fig:attr_payroll}
\end{figure*}

\begin{figure*}[ht!]
\centering
\footnotesize
\begin{augbox}{English: lecturer in Welsh and Media Studies, Wales, 39 (Nemotron \texttt{72917})}

A request for a short LEGO story with three hard format constraints.
\vspace{6pt}

\begin{tabularx}{\textwidth}{@{}X|X@{}}
\textbf{Original seed prompt (Nemotron):} & \textbf{Persona-augmented prompt:} \\
\midrule
\emph{``Hey! Can you write a short adventure about my favorite LEGO squirrel who helps a lost duck find its way home in the woods? Please keep the story to exactly three paragraphs and separate them with the markdown divider ***. Also, make sure you don’t use the word “depression” anywhere in the story. Thanks!''}
&
\emph{``ta, so can you slap together a little three-paragraph tale about my favourite LEGO squirrel - *exactly* three, na? - helping a lost duck find its way back home through the woods, but keep it tight like a good Welsh-language haiku, hmm? no "depression" anywhere, just the squirrel, the duck, and maybe a *cyfreithiol* amount of forest wisdom. oh and 1. stick to the *** (triple asterisk) dividers like it’s a rule from the *BBC Cymru Wales* script guidelines - no wiggle room 2. no extra fluff, just the bones of the story, like when i’m editing student essays for their A-levels bfn''} \\
\end{tabularx}

\vspace{6pt}
\hrule
\vspace{6pt}

\textbf{Which profile field accounts for which span}
\vspace{3pt}

\begin{tabularx}{\textwidth}{@{}l|X|X@{}}
\textbf{Field} & \textbf{Value in the profile} & \textbf{Trace in the rewrite} \\
\midrule
Occupation & \footnotesize Senior lecturer in Welsh and Media Studies, University of Wales Trinity Saint David & \footnotesize \texttt{Welsh-language haiku}: the discipline supplies the simile, and a second aside compares the task to marking student essays \\[2pt]
Reference pool & \footnotesize The *BBC Cymru Wales* archive of Welsh-language programming (she occasionally lingers on the 2000s era, even though it’s on YouTube) & \footnotesize \texttt{BBC Cymru Wales}: an institution named in the profile, cited as the authority for a format rule \\[2pt]
Quirk & \footnotesize Repeatedly writes 'cyfreithiol' as a single word (in Welsh, it’s two: 'cyfreithiol' means 'legal') when frustrated by legalese in academic English & \footnotesize \texttt{cyfreithiol}: the documented Welsh insertion, in the exact word the profile names \\[2pt]
Quirk & \footnotesize Under stress, begins sentences with Welsh particles like 'na' (e.g., 'na... let’s talk about this properly') or 'peidio’ (e.g., 'peidio, don’t just [\ldots] & \footnotesize \texttt{na?}: a Welsh particle the profile attaches to sentence openings \\[2pt]
Signoff & \footnotesize 'Cwra’ for 'bye' in Welsh-to-English messages, otherwise 'ta' (Welsh for 'thanks') in replies & \footnotesize \texttt{ta,}: the opening \emph{ta} the profile documents \\[2pt]
Abbreviations & \footnotesize ‘’hwnnw’ for ‘that’, ‘wna’ for ‘wants to’, ‘’n’ for ‘and’ (Welsh mixed), ‘bfn’ for ‘bye for now’ (rarely) & \footnotesize \texttt{bfn}: listed by the profile, and it closes the message \\[2pt]
Seed constraints & \footnotesize \emph{(seed prompt, not the profile)} & \footnotesize \texttt{***}: the three-paragraph count, the divider and the banned word all survive \\[2pt]
\end{tabularx}

\end{augbox}
\caption{Occupation and style together on a constrained task. Six profile fields are traceable into the rewrite while the exact-format requirements of the seed prompt are left intact. Strict intent label (Appendix~\ref{app:strict}; three samples of an external judge): added content (3 of 3). The rewrite adds an instruction to leave out extra fluff and personal context about editing students' A-level essays.}
\label{fig:attr_welsh}
\end{figure*}

\FloatBarrier

\section{Evaluation Details}
\label{app:judges}

This appendix collects the evaluation details behind Section~\ref{sec:eval}: the corpus-wide LLM-judge scores, two design ablations of the rewriting step, the judge-free metrics, and our checks of how much the results depend on the choice of judge.

\paragraph{Corpus-scale LLM-as-a-judge evaluation.}
\texttt{MiniMax-M2.5}, a reasoning model that is not one of the production rewriters, served with SGLang, scores every user turn of the conversations that carry a model persona (552{,}654 turns) and every user turn of a uniform random sample of 24{,}570 of the other conversations (44{,}687 turns). Corpus-wide rates weight the two strata by their share of the corpus's user turns. The judge gives two independent pass/fail verdicts per turn:
\begin{enumerate}[leftmargin=*,itemsep=1pt,topsep=2pt]
    \item \textbf{Intent preservation:} the rewrite still asks for the same thing as the original, adds no new task or invented content, and stays in the same language.
    \item \textbf{Style adoption:} the rewrite takes on the persona's tone, vocabulary, register or typing habits. Background facts about the persona are ignored; only a generic, unstyled rewrite fails.
\end{enumerate}

\begin{table}[ht]
  \caption{\textbf{Corpus-wide LLM-judge evaluation} (\%, higher is better), scored by \texttt{MiniMax-M2.5} and weighted to the corpus. Backbone rows refer to the model that rewrote the user turns. Judged turns: user turns scored by the judge in each row. 95\% intervals are within $\pm$0.6 points.}
  \label{tab:full-judge}
  \centering
  \small
  \begin{tabular}{l|r|ccc}
    \toprule
    Subset & Judged turns & Intent pass & Style pass & Both pass \\
    \midrule
    \textbf{All user turns} & \textbf{597{,}341} & \textbf{96.6} & \textbf{67.1} & \textbf{64.9} \\
    \textit{By generator backbone} & & & & \\
    \quad Gemma-4-31B      & 371{,}550 & 98.6 & 69.8 & 68.8 \\
    \quad Ministral-3-14B  & 225{,}791 & 93.7 & 63.2 & 58.9 \\
    \textit{By seed dataset} & & & & \\
    \quad WildChat         & 25{,}088 & 96.0 & 60.5 & 58.0 \\
    \quad UltraChat        & 307{,}750 & 97.3 & 76.0 & 74.0 \\
    \quad Nemotron         & 264{,}503 & 96.6 & 65.2 & 62.7 \\
    \bottomrule
  \end{tabular}
\end{table}

Table~\ref{tab:full-judge} reports the results. Three observations stand out.

First, the corpus-wide rates agree with the controlled study. Over the whole corpus, 96.6\% of rewrites preserve intent and 67.1\% adopt the persona's style (64.9\% pass both), at or above the levels of the controlled comparison in Table~\ref{tab:flat_vs_structured} (92.0\% intent, 60 to 64\% style). The controlled finding is therefore not a small-sample effect. For cross-lingual turns, which are translated before they are restyled, intent is judged against the translation, because the intent template fails any change of language by design.

Second, the larger rewriter does better. \texttt{Gemma-4-31B} beats \texttt{Ministral-3-14B} by 4.9 points on intent and 6.6 on style, and passes both checks 9.9 points more often (68.8\% against 58.9\%). Preserving a complex request while writing in an informal, non-standard voice is simply harder for the smaller model.

Third, the seed data matters. Open-ended UltraChat conversations leave more room for style (76.0\% style, 97.3\% intent) than Nemotron's denser, often technical instructions (65.2\% style, 96.6\% intent), where restyling more easily collides with strict formatting or technical wording. Real-user WildChat prompts are restyled least often (60.5\% style), while intent stays at 96.0\%.

\paragraph{Persona selection and rewrite prompt.}\label{app:rewriteabl} We ablate two choices of the user-turn rewrite on 400 language-matched (free-mode) conversations per arm, with the same rewriter (\texttt{Gemma-4-31B}), judge and sanity gate, in a $2\times 2$ design. \emph{Persona selection}: a random persona that shares the prompt's language, or the conservative matching of Section~\ref{sec:pipeline} (location, then top-1 embedding, then language pool). \emph{Rewrite prompt}: the rewrite template alone as a single user message, or the same template wrapped in an additional system prompt that frames the rewriter as a generic text-transformation component. Table~\ref{tab:rewriteabl} shows that persona selection dominates: conservative matching raises style adoption by 22 to 23 points and also preserves intent slightly better, because a topically and geographically plausible persona gives the rewriter something concrete to express. The extra framing costs 4 to 5 points of style in both selection settings, so the pipeline uses the template alone.

\begin{table}[ht]
  \caption{\textbf{Persona selection and rewrite prompt ablation} (\%, higher is better). 400 conversations per arm, about 960 judged user turns each; 95\% intervals at most $\pm$3.2 points. Bold: best per column.}
  \label{tab:rewriteabl}
  \centering
  \small
  \begin{tabular}{ll|ccc}
    \toprule
    Persona selection & Rewrite prompt & Intent & Style & Sanity pass \\
    \midrule
    Random, same language & template + framing & 97.2 & 45.6 & 96.2 \\
    Random, same language & template only & 96.2 & 50.4 & 94.2 \\
    Conservative matching & template + framing & \textbf{99.2} & 68.6 & 95.0 \\
    Conservative matching & template only & 98.7 & \textbf{72.5} & 95.5 \\
    \bottomrule
  \end{tabular}
\end{table}

\paragraph{Cross-lingual rewriting strategy.}\label{app:crossling} We compare two ways of producing cross-lingual user turns on the same 300 conversations, personas and rewriter (\texttt{Gemma-4-31B}): \emph{single pass}, where the source turn is re-voiced directly into the persona's language, and \emph{translate, then restyle}, where each turn is first translated literally into that language and the translation is then restyled with the same-language rewrite. Both variants are judged against the same literal translation of the turn, because the intent template fails any language switch by design, and the target language is checked with fastText language identification. Table~\ref{tab:crossling} shows that the two-step variant keeps the request more often, reaches the target language more often and passes the sanity gate slightly more often, while the single pass adopts the persona's style slightly more often. We use the two-step variant, since a rewrite that drifts from the request or out of the target language defeats the purpose of the cross-lingual mode.

\begin{table}[ht]
  \caption{\textbf{Cross-lingual rewriting strategy} (\%, higher is better). 300 conversations per variant, about 730 judged user turns each, same personas, rewriter and judge; 95\% intervals at most $\pm$3.6 points. Bold: best per column.}
  \label{tab:crossling}
  \centering
  \small
  \begin{tabular}{l|cccc}
    \toprule
    Variant & Intent & Style & Target language & Sanity pass \\
    \midrule
    Single pass & 96.0 & \textbf{61.1} & 84.6 & 94.3 \\
    Translate, then restyle & \textbf{99.0} & 58.2 & \textbf{88.6} & \textbf{95.7} \\
    \bottomrule
  \end{tabular}
\end{table}

\paragraph{Corpus-scale judge-free fidelity metrics.}
These metrics use no LLM judge, so they are free of judge bias. They fall into two groups: how much a rewrite changes the text and what it keeps intact, and how closely it follows the style values in the persona's profile.
\subparagraph{Transformation depth and constraint retention.}
\begin{itemize}[leftmargin=*,itemsep=1pt,topsep=1pt]
    \item \textbf{Same-language token Jaccard distance:} Defined as $1 - \frac{|T_{\text{orig}} \cap T_{\text{rw}}|}{|T_{\text{orig}} \cup T_{\text{rw}}|}$, where token set $T$ is extracted multilingually by treating CJK/Hangul ideographs as atomic units and non-CJK scripts as whitespace- and punctuation-delimited words.
    \item \textbf{Same-language normalized edit distance:} Character-level Levenshtein distance normalized by sequence length: $\frac{\text{Lev}(s_{\text{orig}}, s_{\text{rw}})}{\max(|s_{\text{orig}}|, |s_{\text{rw}}|)}$, bounded in $[0, 1]$.
    \item \textbf{Core-vocabulary token retention:} Filters non-informative functional tokens (retaining words of length $>2$ or non-ASCII tokens) to measure the fraction of core lexical content preserved: $\frac{|C_{\text{orig}} \cap C_{\text{rw}}|}{|C_{\text{orig}}|}$.
    \item \textbf{Multilingual semantic similarity:} Cosine similarity between 768-dimensional sentence embeddings computed via Language-Agnostic BERT Sentence Embeddings (\textsc{LaBSE}) across all parallel $(s_{\text{orig}}, s_{\text{rw}})$ turn pairs.
    \item \textbf{Numbers \& quantities intact:} Regular expression extraction targeting integers, decimals, currency values, percentages, and version strings (e.g., \texttt{v2.1}), verifying that numbers in $s_{\text{orig}}$ form a subset of numbers in $s_{\text{rw}}$.
    \item \textbf{Code \& technical identifiers intact:} Lexical parsing targeting markdown code blocks, backtick-enclosed identifiers, camelCase/snake\_case tokens, URIs, Unix file paths, and CLI flags.
    \item \textbf{Quoted string literals intact:} Verifies retention of verbatim text enclosed in single, double, or regional quotation marks (e.g., Japanese corner brackets and French guillemets).
    \item \textbf{All entities preserved per turn:} A strict composite metric requiring that every entity category present in a given turn (numbers, code, quotes) is simultaneously $100\%$ preserved.
\end{itemize}
\subparagraph{Style adherence.}
Each metric checks the rewrite against the matching attribute in the persona's profile:
\begin{itemize}[leftmargin=*,itemsep=1pt,topsep=1pt]
    \item \textbf{Emoji \& emoticon register:} Text is matched against Unicode 15.0 emoji codepoints, ASCII emoticons (e.g., \texttt{:-)}, \texttt{<3}), and kaomoji (e.g., \texttt{(\textasciicircum\_\textasciicircum)}, \texttt{orz}). A turn passes if it satisfies its target persona tier: \textit{none} (0 emojis/emoticons), \textit{rare/functional} ($\le 2$ functional emojis), \textit{frequent/decorative} ($\ge 1$ decorative emoji), or \textit{kaomoji} ($\ge 1$ kaomoji pattern).
    \item \textbf{Capitalization register:} For cased scripts, sentence-start capitalization and casing proportions are verified against target profiles: \textit{all-lowercase} ($\ge 70\%$ lowercase or 0 capitalized sentence starts), \textit{standard} ($\ge 70\%$ capitalized sentence starts), \textit{first-word-only} (only the initial turn character capitalized, without terminal periods), and \textit{all-caps} (presence of new capitalized tokens for emphasis).
    \item \textbf{Punctuation register:} Evaluates line-ending periods, ellipses (\texttt{...}), linebreaks, and expressive punctuation (\texttt{!!}, \texttt{??}) against persona categories: \textit{standard} ($\ge 70\%$ punctuated line endings), \textit{minimal / drops period} (omission of terminal periods), \textit{ellipsis-heavy}, and \textit{expressive}.
    \item \textbf{Length ratio short/medium/long:} Median length ratio $\frac{|s_{\text{rw}}|}{|s_{\text{orig}}|}$ conditioned on the persona's specified verbosity tendency (\texttt{Short}, \texttt{Medium}, \texttt{Long}), corroborated by a positive Spearman rank correlation.
\end{itemize}

Table~\ref{tab:fidelity} reports the full judge-free fidelity metrics.
\begin{table}[ht]
  \caption{\textbf{Full-corpus judge-free fidelity metrics.} Distances and similarities in $[0,1]$; retention and adherence in \%; length ratios relative to the original turn, with the per-item Spearman $\rho$ in parentheses.}
  \label{tab:fidelity}
  \centering\small
  \begin{tabular}{ll|r}
    \toprule
    & Metric & Value \\
    \midrule
    \multirow{4}{*}{\shortstack[l]{Transformation\\depth}}
      & Same-language token Jaccard distance & 0.443 \\
      & Same-language normalized edit distance & 0.396 \\
      & Core-vocabulary token retention & 73.7 \\
      & Multilingual semantic similarity (LaBSE) & 0.791 \\
    \midrule
    \multirow{4}{*}{\shortstack[l]{Constraint\\retention}}
      & Numbers \& quantities intact & 98.9 \\
      & Code \& technical identifiers intact & 90.4 \\
      & Quoted string literals intact & 95.5 \\
      & All entities preserved (per turn) & 95.2 \\
    \midrule
    \multirow{4}{*}{\shortstack[l]{Style\\adherence}}
      & Emoji \& emoticon register & 92.2 \\
      & Capitalization register & 54.4 \\
      & Punctuation register & 69.9 \\
      & Length ratio short/medium/long (per-item $\rho$) & 1.00$\times$/1.02$\times$/1.56$\times$ (0.22) \\
    \bottomrule
  \end{tabular}
\end{table}

\paragraph{Multi-judge agreement.} All style and intent scores in the paper come from LLM judges, so we measure how much they depend on which judge is used. We re-score a stratified set of 100 persona rewrites (spanning the sixteen single-attribute conditions of Appendix~\ref{app:attribution}) with four independent judges (\texttt{Gemma-4-31B}, \texttt{GLM-4.7-Flash}, \texttt{Ministral-3-14B}, \texttt{MiniMax-M2.5}), each issuing a binary style-adoption and intent-preservation verdict per rewrite. Table~\ref{tab:judge-means} reports each judge's mean verdict rate and Table~\ref{tab:judge-agree} the pairwise item-level agreement.

\begin{table}[ht]
  \caption{\textbf{Per-judge mean verdict rate} (share of passing verdicts) on the same 100 persona rewrites.}
  \label{tab:judge-means}
  \centering
  \small
  \begin{tabular}{l|cc}
    \toprule
    Judge & Mean style & Mean intent \\
    \midrule
    Gemma-4-31B     & 0.840 & 0.680 \\
    GLM-4.7-Flash   & 0.929 & 0.830 \\
    Ministral-3-14B & 0.920 & 0.930 \\
    MiniMax-M2.5    & 0.778 & 0.680 \\
    \bottomrule
  \end{tabular}
\end{table}

\begin{table}[ht]
  \caption{\textbf{Pairwise item-level judge agreement} (share of identical verdicts) on the 100 rewrites. Upper triangle: style-adoption verdict; lower triangle: intent-preservation verdict.}
  \label{tab:judge-agree}
  \centering
  \small
  \begin{tabular}{l|cccc}
    \toprule
     & Gemma & GLM & Ministral & MiniMax \\
    \midrule
    Gemma     & --    & 0.899 & 0.880 & 0.879 \\
    GLM       & 0.790 & --    & 0.869 & 0.827 \\
    Ministral & 0.730 & 0.880 & --    & 0.818 \\
    MiniMax   & 0.940 & 0.830 & 0.730 & --    \\
    \bottomrule
  \end{tabular}
\end{table}

\begin{table}[ht]
  \caption{\textbf{Chance-corrected pairwise judge agreement} on the same rewrites: Cohen's $\kappa$ and Gwet's AC1, which is robust to the high pass rates of lenient judges. Computed from the observed agreement (Table~\ref{tab:judge-agree}) and each judge's marginal pass rate (Table~\ref{tab:judge-means}).}
  \label{tab:judge-kappa}
  \centering
  \small
  \begin{tabular}{l|cc|cc}
    \toprule
    & \multicolumn{2}{c|}{Style} & \multicolumn{2}{c}{Intent} \\
    \cmidrule(lr){2-3}\cmidrule(lr){4-5}
    Judge pair & $\kappa$ & AC1 & $\kappa$ & AC1 \\
    \midrule
    Gemma -- GLM          & 0.52 & 0.87 & 0.45 & 0.67 \\
    Gemma -- Ministral    & 0.44 & 0.85 & 0.22 & 0.61 \\
    Gemma -- MiniMax      & 0.61 & 0.82 & 0.86 & 0.89 \\
    GLM -- Ministral      & 0.06 & 0.85 & 0.44 & 0.85 \\
    GLM -- MiniMax        & 0.34 & 0.77 & 0.55 & 0.73 \\
    Ministral -- MiniMax  & 0.32 & 0.76 & 0.22 & 0.61 \\
    \bottomrule
  \end{tabular}
\end{table}

The judges differ most in how lenient they are, most visibly on style, where \texttt{MiniMax-M2.5} is strictest and \texttt{GLM-4.7-Flash} most lenient: mean style pass rates range from 0.78 to 0.93. Raw agreement on individual items is high (0.82 to 0.90 on style) and looser on intent (0.73 to 0.94), but part of that is expected by chance when most items pass, so raw agreement overstates consistency. Table~\ref{tab:judge-kappa} corrects for this. Chance-corrected agreement is moderate to strong for pairs with comparable pass rates, and every pair involving the production judge (\texttt{MiniMax-M2.5}) reaches $\kappa \geq 0.32$ on style and $\kappa \geq 0.22$ on intent, with the strongest agreement between \texttt{MiniMax-M2.5} and \texttt{Gemma-4-31B} on intent ($\kappa = 0.86$). The near-zero style $\kappa$ between \texttt{GLM-4.7-Flash} and \texttt{Ministral-3-14B} arises because both pass more than 92\% of items, which leaves little variance to agree on; AC1, which corrects for this prevalence effect, is 0.85 for the same pair. The effective sample is 99 items for the style verdicts and 100 for intent. In short, a result obtained with one judge carries over to the others in direction, while absolute pass rates depend on the judge; this is why we report the external re-judging below rather than relying on a single judge.

\begin{figure}[ht]
\centering
\includegraphics[width=0.72\linewidth]{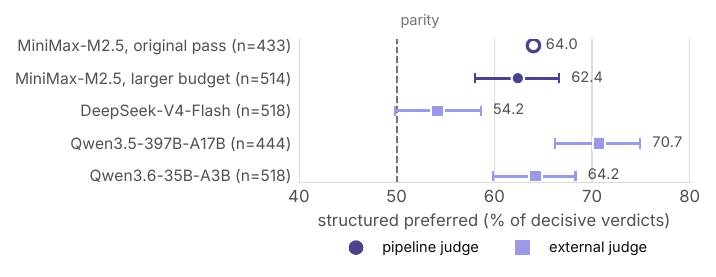}
\caption{Structured versus flat style representation under the pipeline judge and three external judges: share of decisive verdicts preferring the structured rewrite, with 95\% Wilson intervals (Table~\ref{tab:ab-rejudge}).}
\label{fig:ab-rejudge}
\end{figure}

\paragraph{External re-judging of the structured versus flat A/B.} All four judges above are also rewriters or the production judge of the pipeline, so they are not fully independent of it. We therefore re-score all 518 stored pairs of the A/B in Table~\ref{tab:flat_vs_structured}, with the flat and structured rewrites in a per-item random order, using \texttt{MiniMax-M2.5} with a larger output budget and three judges from outside the pipeline, served through a separate inference stack. Table~\ref{tab:ab-rejudge} and Figure~\ref{fig:ab-rejudge} report the results. The larger budget recovers 514 of 518 verdicts for \texttt{MiniMax-M2.5}, which shows that the attrition in the original pass (433 of 518) came from truncated reasoning rather than from the items, and reproduces its direction (62.4\% against 64\% of decisive verdicts for structured). All three external judges prefer the structured representation; the effect is clear for both Qwen models and small for \texttt{DeepSeek-V4-Flash}, whose interval touches parity. Intent preservation is identical for the two representations under every judge.

\begin{table}[ht]
  \caption{\textbf{Structured vs.\ flat representation, re-judged} (\%). Preference shares are over parsed verdicts; the decisive share excludes ties, with a 95\% Wilson interval. \texttt{Qwen3.5-397B-A17B} returned no parseable verdict for 74 items.}
  \label{tab:ab-rejudge}
  \centering
  \small
  \setlength{\tabcolsep}{4pt}
  \resizebox{\linewidth}{!}{%
  \begin{tabular}{l|c|cccc|cc}
    \toprule
    & & \multicolumn{4}{c|}{Style preference} & \multicolumn{2}{c}{Intent} \\
    \cmidrule(lr){3-6}\cmidrule(lr){7-8}
    Judge & Parsed & Structured & Flat & Tie & Structured, decisive & Flat & Struct. \\
    \midrule
    \texttt{MiniMax-M2.5} (original pass) & 433 & 60 & 33 & 7 & 64 & 92 & 92 \\
    \texttt{MiniMax-M2.5} (larger budget) & 514 & 58.4 & 35.2 & 6.4 & 62.4 (58.0--66.6) & 92.3 & 92.5 \\
    \texttt{DeepSeek-V4-Flash} & 518 & 51.9 & 43.8 & 4.2 & 54.2 (49.8--58.6) & 92.4 & 92.6 \\
    \texttt{Qwen3.5-397B-A17B} & 444 & 66.9 & 27.7 & 5.4 & 70.7 (66.2--74.9) & 92.7 & 93.5 \\
    \texttt{Qwen3.6-35B-A3B} & 518 & 61.0 & 34.0 & 5.0 & 64.2 (59.9--68.3) & 92.4 & 92.6 \\
    \bottomrule
  \end{tabular}}
\end{table}

\FloatBarrier

\paragraph{Strict intent label.}\label{app:strict}

Every user turn of the release carries a strict intent label. The judge, \texttt{MiniMax-M2.5}, comes from a different
model family than the rewriters and never sees the persona. It compares each turn with its seed turn (translated
turns with the English source, allowing the change of language) and assigns one of four labels: same request, added
content, dropped content or scope change. A second external judge (\texttt{Qwen3.6-35B-A3B}) audited 4{,}903 turns
and agrees with it at $\kappa=0.59$ (four-way) and 0.63 (same request vs.\ drift). Both judges find the core request,
a looser check of whether the main ask is kept, preserved in 97.2\% of turns. The production intent template of
Table~\ref{tab:full-judge} is lenient toward added content: it passes 91.5\% of the turns that the strict label flags.

\begin{center}\small
\begin{tabular}{lcccc}
\toprule
& Same request & Added content & Dropped content & Scope change \\
\midrule
User turns (\%) & 74.0 & 18.5 & 4.1 & 3.5 \\
\bottomrule
\end{tabular}
\end{center}
Rates are over the 2{,}371{,}618 labelled user turns. We recommend the strict subset, the conversations in which every user turn is labelled same request,
as the default for training; retraining on it changes no downstream result (Appendix~\ref{app:downstream}).

\paragraph{Where additions come from.} We split the persona into its parts on the 800 items of
Table~\ref{tab:formats-main}. With \texttt{Gemma-4-31B}, the background alone adds content in 1.5\% of turns (0.0\%
without a persona), while \texttt{Ministral-3-14B} adds content in 8.9\% of turns even without any persona. Additions
therefore follow the rewriter's instruction following, which the per-rewriter label makes visible.

\paragraph{Probe vs.\ corpus.} The attribute probe of Appendix~\ref{app:attribution} and the corpus evaluation use the
same intent template; the difference lies in the data. The probe measures single-call rewriting by four models, two of
which are not used in the release, without masking, retries or the sanity gate, and with older single-attribute
profiles. \texttt{Gemma-4-31B} preserves intent in about 77\% of probe rewrites and in 98.6\% of corpus turns
(Table~\ref{tab:full-judge}), \texttt{Ministral-3-14B} in about 38\% and 93.7\%.

\paragraph{Human check of the strict judge.}\label{app:humancheck}

Two annotators labelled 100 sampled turns with the same four labels, blind to the persona, the rewriter, the judge's
label and the source of each row; both annotators are authors of the paper. They agree with each other at
$\kappa=0.60$ on same request vs.\ drift, and the strict judge agrees with them at $\kappa=0.60$ and 0.41. In the 9
cases where both annotators agree and the judge does not, the judge is the stricter side 8 times, so the drift rates
we report are, if anything, conservative. Both annotators consider the core request preserved in 92 of the 100 turns.
On the 25 turns the judge labelled added content, the annotators confirm most additions, which they describe as
written-out requirements, extra context or an implicit question made explicit; the core request survives in 23 of 25.
Several drift labels trace back to the seed rather than to the rewrite: in 5 of annotator A's 34 drift labels, the
annotator flagged the seed itself as unclear (e.g.\ ``Prepositions of place''). When a seed does not state a clear
request, any natural rewrite has to add some framing to it.

\paragraph{Cross-lingual intent against the English source.}\label{app:crossling-source}

Three judges (\texttt{MiniMax-M2.5}, \texttt{Qwen3.6-35B-A3B}, \texttt{Muse-Glimmer-30B}) compared both the intermediate
translation and the final turn with the English source on 2{,}000 turns in 64 languages, allowing the change of
language. The translation step keeps the request in 95.6 to 98.7\% of turns, and the core request survives the full
pipeline in 95.9 to 96.6\%. Strict same-request rates on the final turn (61.5 to 78.0\%) show that differences arise at
the restyling step, the same mechanism as in Appendix~\ref{app:strict}, and the strict label flags them in the release.

\paragraph{Reference rows for the judge-free metrics.}\label{app:refrows}

We computed the metrics of Table~\ref{tab:fidelity} for four representations on the same 800 seeds of
Table~\ref{tab:formats-main}, with the same persona, rewriter (\texttt{Gemma-4-31B}) and decoding. The no-persona row
gives the floor of change. Table~\ref{tab:fidelity} covers the whole corpus with both rewriters, which is why its
values differ from this single-rewriter subset.
\begin{center}\small
\begin{tabular}{lcccc}
\toprule
Metric & No persona & PersonaHub form & SynthPAI form & PersonaFactory \\
\midrule
LaBSE similarity & 0.969 & 0.948 & 0.924 & 0.899 \\
Core vocabulary (\%) & 91.7 & 86.7 & 85.3 & 81.5 \\
Normalized edit distance & 0.109 & 0.195 & 0.221 & 0.284 \\
Punctuation register (\%) & 40.5 & 41.0 & 64.5 & 75.8 \\
\bottomrule
\end{tabular}
\end{center}
Each step up in style information moves the text further from the seed and raises adherence to the persona's style.
Our format makes the largest surface change and reaches the highest adherence, while the strict label and the human
check confirm that the request is kept.

\paragraph{Results per language and rewriter.}\label{app:perlang}

The corpus judge of Table~\ref{tab:full-judge} on the 15 most frequent languages of the corpus: share of user turns
that pass both the intent and the style check (\%). \texttt{Gemma-4-31B} leads \texttt{Ministral-3-14B} in each
language; Arabic, Vietnamese and Russian have the lowest style adoption.
\begin{center}\small
\begin{tabular}{lcc@{\hskip 1.2cm}lcc}
\toprule
Language & Gemma & Ministral & Language & Gemma & Ministral \\
\midrule
English & 75.1 & 67.6 & Portuguese & 71.4 & 59.7 \\
Russian & 59.6 & 45.0 & Vietnamese & 59.6 & 31.7 \\
Chinese & 57.6 & 40.8 & German & 72.6 & 59.4 \\
Spanish & 73.3 & 60.6 & Italian & 76.6 & 55.0 \\
Arabic & 48.2 & 28.8 & Japanese & 59.6 & 44.8 \\
French & 67.2 & 56.7 & Thai & 60.6 & 42.9 \\
Indonesian & 72.9 & 60.8 & Tagalog & 71.3 & 31.2 \\
& & & Polish & 70.4 & 59.0 \\
\bottomrule
\end{tabular}
\end{center}

\paragraph{Computational Turing test.}\label{app:turing} Following \citet{pagan2026distinguishable}, we test detectability in two
settings. On synthetic seeds, the primary use case of the pipeline, a stylometric classifier separates real WildChat
user turns from UltraChat and Nemotron originals and, separately, from their rewrites. Rewriting moves typing toward
real users, and detection drops by 15.1 points [13.8, 16.5]:
\begin{center}\small
\begin{tabular}{lccc}
\toprule
& Synthetic seeds, original & Rewritten & Real users \\
\midrule
Lowercase start & 0.5\% & 43\% & 27\% \\
Missing final punctuation & 3\% & 47\% & 55\% \\
Stylometric detection & 91.3\% & 76.2\% & n/a \\
\bottomrule
\end{tabular}
\end{center}
Fine-tuned encoders still separate both the originals and the rewrites from WildChat, because the topics of synthetic
seeds differ; the pipeline keeps the request, so this topic gap is expected. On real WildChat turns in the paired
setting, a fine-tuned XLM-R distinguishes a real turn from our rewrite of it at 84.6\% accuracy and from a persona-free
paraphrase at 75.9\%, while stylometric features alone reach 68.2\%. This matches the main finding of Pagan et al.:
LLM-generated text stays distinguishable for a trained classifier, and pushing it toward human-likeness costs semantic
fidelity. Our pipeline sits on the fidelity side by design, as it keeps the original request.

\paragraph{Homogenization.} We measure style diversity as the mean pairwise distance between user turns in a style
embedding. Rewriting raises it from 0.274 to 0.744 for Nemotron seeds and from 0.567 to 0.801 for UltraChat seeds,
close to real WildChat users (0.857). Real logs are already diverse, and rewriting narrows them somewhat (to 0.737);
for real logs, the value of the pipeline is multilingual extension that keeps the original request. Turns of the same
persona are more similar to each other than to other personas' turns, so personas give consistent, distinguishable
voices.

\paragraph{Algorithmic fidelity.} The four criteria of \citet{argyle2023outofone}:
\begin{center}\small
\begin{tabular}{lp{0.68\linewidth}}
\toprule
Criterion & Result \\
\midrule
Turing test & Part C: 59\% of rewrites and 71\% of WildChat user turns judged human-written; computational tests above \\
Backward continuity & Persona punctuation recovered from the text alone in 59.8\% vs.\ 33.9\% for rewrites conditioned on the background only \\
Forward continuity & Core request preserved in 97.2\% of turns \\
Pattern correspondence & Across 162 country-language groups, Spearman $\rho=0.48$ for slang, 0.24 for lowercase start and question marks \\
\bottomrule
\end{tabular}
\end{center}
Pattern correspondence holds at the level of language: country-level differences within a language are not yet
reproduced, and emoji rates run against real users ($\rho=-0.3$). Background traits are intentionally not recoverable
from the text, since the rewrite prompt forbids inserting them.

\FloatBarrier

\section{Comparison to Other Persona Representations}
\label{app:formats}

Section~\ref{sec:eval} compares a structured style representation against a free-text one
within our own schema. This appendix asks the same question against the formats the field
actually uses: PersonaHub \citep{chan2024personahub} and SynthPAI \citep{yukhymenko2024synthpai}.

\paragraph{Design.} We hold the persona \emph{content} fixed and vary only its
serialization. From the released corpus we sample 800 conversations, stratified by persona
set and seed source, together with the persona the pipeline matched to each. Sampling is
independent of the released sanity verdict, and every condition is regenerated from
scratch, so no condition inherits an advantage from the corpus filtering. Each profile is
then rendered four ways and used to rewrite the same seed prompt with the same rewriter,
decoding parameters, judge and prompt scaffold. Only the persona block differs.

The PersonaHub rendering is a single third-person sentence. This is not a simplification
on our part: across all 200{,}000 released PersonaHub personas, only 1.4\% mention any
writing cue at all, so a projection that described typing behavior would not be testing
PersonaHub. The SynthPAI rendering keeps the feature types that schema has, an attribute
block plus a free-text writing-style description with a verbatim example, and drops the
per-axis style keys, which it has no slot for. Projections were produced by a third model
that is neither rewriter nor judge, and were accepted only after matching the real
PersonaHub length distribution and style-mention rate. Table~\ref{tab:formats} reports what
survives each rendering.

\begin{figure}[t]
\centering
\includegraphics[width=0.72\linewidth]{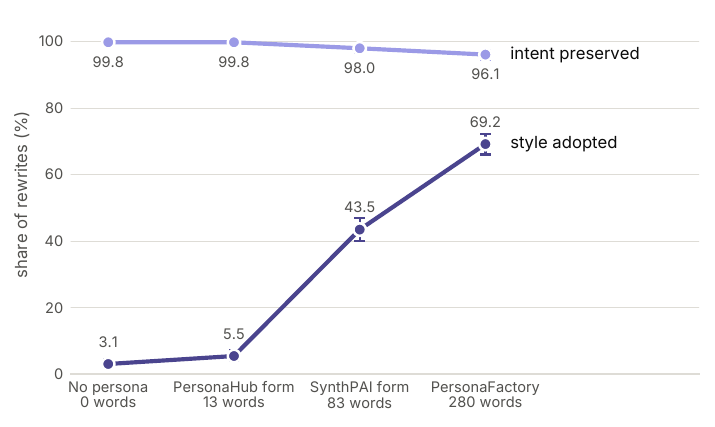}
\caption{Style adoption and intent preservation by persona representation ($N=800$ paired items, 95\% Wilson intervals).}
\label{fig:format-tradeoff}
\end{figure}

\paragraph{Result.} Table~\ref{tab:formatresults} and Figure~\ref{fig:format-tradeoff} show a dose-response in how explicitly a
representation encodes writing behavior. Style adoption rises monotonically across the
four conditions, and the judge prefers the PersonaFactory rewrite in every pairwise
comparison (share of items where the judge prefers the PersonaFactory rewrite, with order randomized per item). Style adoption and intent use the production judge templates; paired bootstrap intervals for the judge-free differences are at most $\pm$0.05. The contested comparison is against the SynthPAI form, which also carries style
information and differs from ours mainly in whether that information is structured; there
the preference is 4.83 to 1.

Two observations do not follow from the setup. First, the PersonaHub form is not inert. It
moves the rewrite away from the no-persona baseline about as far as the other conditions do,
yet lifts style adoption only from 3.1\% to 5.5\%: it changes the text without changing it
toward the persona. Second, the representation propagates past the user turn. Measured
against the no-persona reply for the same item, the assistant's response moves further as
the representation grows richer, while assistant sanity stays flat across conditions, so
this is a change of register rather than of quality.

\paragraph{Cost.} The gains are not free, and the cost is concentrated rather than uniform.
Intent preservation is essentially unaffected in English but falls to 89.2\% on the
cross-lingual half, and entity retention declines by four points. Transformation-depth
metrics also fall, but those measure how much the surface changed and are the intended
effect rather than a loss.

\paragraph{Robustness.} Table~\ref{tab:formatrobust} reports three checks. The ranking is
unchanged with Ministral-3-14B as rewriter, though the margin against the SynthPAI form
narrows from 4.83 to 2.64, consistent with Appendix~\ref{app:attribution} finding that the
rewriter dominates attribute effects. Re-judging a 200-item subset with four judges leaves
the ordering intact while absolute rates span fifteen points, matching the severity spread
in Appendix~\ref{app:judges}; item-level agreement with the primary judge is 0.81 to 0.93.
Finally, judge responses that could not be parsed are not uniformly distributed across
conditions, so we report a worst-case bound in which every unresolved comparison is
assigned to the competing condition. The SynthPAI comparison holds at 76.8\% under that
assumption.

\begin{figure}[t]
\centering
\includegraphics[width=\linewidth]{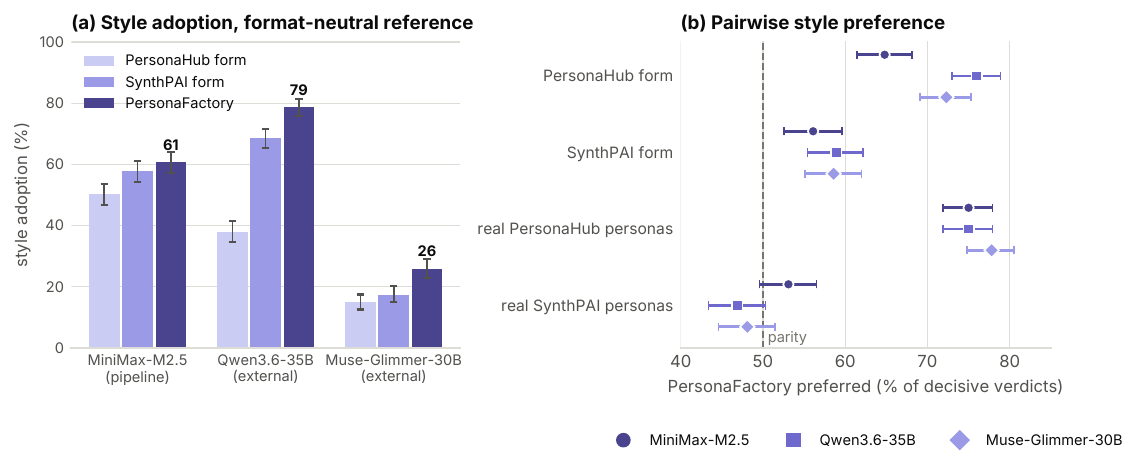}
\caption{Persona representations re-judged against a format-neutral reference by the pipeline judge and two external judges. (a) Style adoption per format. (b) Share of decisive pairwise verdicts preferring the PersonaFactory rewrite, with 95\% Wilson intervals; real-persona comparisons are reference-free.}
\label{fig:format-neutral}
\end{figure}

\paragraph{Format-neutral re-judging and real personas.} In the comparison above, the style and pairwise judges see the persona rendered in the PersonaFactory format, which the PersonaFactory rewriter also saw, so a rewrite could in principle score by echoing phrasing from its own prompt. We therefore re-judge the same 3{,}200 rewrites against a format-neutral reference that lists only the persona's writing-behaviour attribute values (structured typing keys, sentence length, question framing, code-switching frequency) with all example phrasing removed, using the pipeline judge and two external judges (\texttt{Qwen3.6-35B-A3B} and \texttt{Muse-Glimmer-30B}). We also rewrite every item with a real PersonaHub persona and a real SynthPAI profile in their native form, with the same rewriter and template, and compare each with the PersonaFactory rewrite in a reference-free pairwise judgment of which rewrite gives the message a more distinctive and realistic individual voice while keeping the request intact. Figure~\ref{fig:format-neutral} shows that PersonaFactory keeps the highest style adoption under every judge and is preferred over both projected formats and over real PersonaHub personas, with every interval above parity. Real SynthPAI profiles, whose hand-written style descriptions cover 300 English-speaking Reddit users, are on par with PersonaFactory; they are preferred on cross-lingual prompts, where our persona speaks a different language than the prompt, while PersonaFactory scales the same style fidelity to a bank of nearly 100{,}000 multilingual profiles with per-axis control.

\paragraph{Naturalness.} Stronger stylization has a cost. In a reference-free judgment of whether a message reads as something a real person would type or as a caricature, the pipeline judge passes 99\% of rewrites in every condition, but the two external judges rate PersonaFactory rewrites 7 to 12 points less natural than unconditioned rewrites. We attribute this to the richer behaviour specification: a profile that asks for lowercase typing, dropped punctuation, abbreviations, emoji and a sign-off at the same time invites the rewriter to realize several markers in one short message, which a judge can read as over-performed, whereas formats that specify little change little. The per-turn quirk probability has no consistent effect on this; an explicit cap on the number of markers realized per turn does, as the next paragraph shows.

\paragraph{Quirk probability and marker cap.}\label{app:markercap}

We evaluated $p = 0.10$, 0.15 and 0.25, each with and without a cap of two realized chat markers per turn (greeting,
sign-off, emoji, slang, ellipsis, repeated punctuation). The cap is an instruction in the rewrite prompt, not a
post-hoc filter, so it changes how the whole turn is written. To compare like with like, all settings were generated
fresh on the 646 of the 800 items of Table~\ref{tab:formats-main} that pass the pipeline's checks in all six settings
(so the uncapped values differ from Table~\ref{tab:formats-main}), with the reference formats and real user turns in the
same table:
\begin{center}\small
\begin{tabular}{lccc}
\toprule
Setting (\texttt{Gemma-4-31B}) & Style adoption & Same request & Naturalness (Muse-Glimmer) \\
\midrule
$p = 0.10$ to 0.25, no cap & 63.0 to 64.6 & 88.5 to 90.7 & 83.1 to 83.9 \\
$p = 0.10$ to 0.25, cap & 52.9 to 57.6 & 89.0 to 89.8 & 87.6 to 88.7 \\
SynthPAI format (no cap) & 46.1 & 90.2 & 87.0 \\
No persona & 3.7 & 94.6 & 91.3 \\
Real WildChat user turns & n/a & n/a & 94.2 \\
\bottomrule
\end{tabular}
\end{center}
With the cap, our rewrites are as natural as the SynthPAI-format rewrites while adopting 7 to 11 points more of the
persona's style, at the same intent preservation; a second judge (\texttt{Qwen3.6-35B-A3B}) shows the same picture.
Capped rewrites remain 2.6 to 3.7 points less natural than rewrites without any persona, which is the cost of adopting
a style at all. The quirk probability $p$ has no consistent effect; the cap is the effective lever and our recommended
setting. The representational result of Table~\ref{tab:flat_vs_structured} holds under the cap: with both arms
regenerated with the cap, the structured style layer remains preferred under all three judges, with a smaller margin
(62.0 to 65.1\% of decisive verdicts without the cap, 54.9 to 58.1\% with it). On real WildChat seeds with both
rewriters, the cap likewise trades a few points of style for 4 to 5 points of naturalness at about 99\% intent
preservation. The corpus was generated with $p = 0.25$ and the prompt's instruction to use at most two stylistic
elements, but without the explicit cap: 97.3\% of its user turns contain at most two of the six markers, while
greetings and emoji remain more frequent than in real user turns. We will add a per-turn marker count to the release so
that users can filter by marker density, and make the cap the default in the released code.

\begin{figure*}[ht]
\centering
\includegraphics[width=0.96\textwidth]{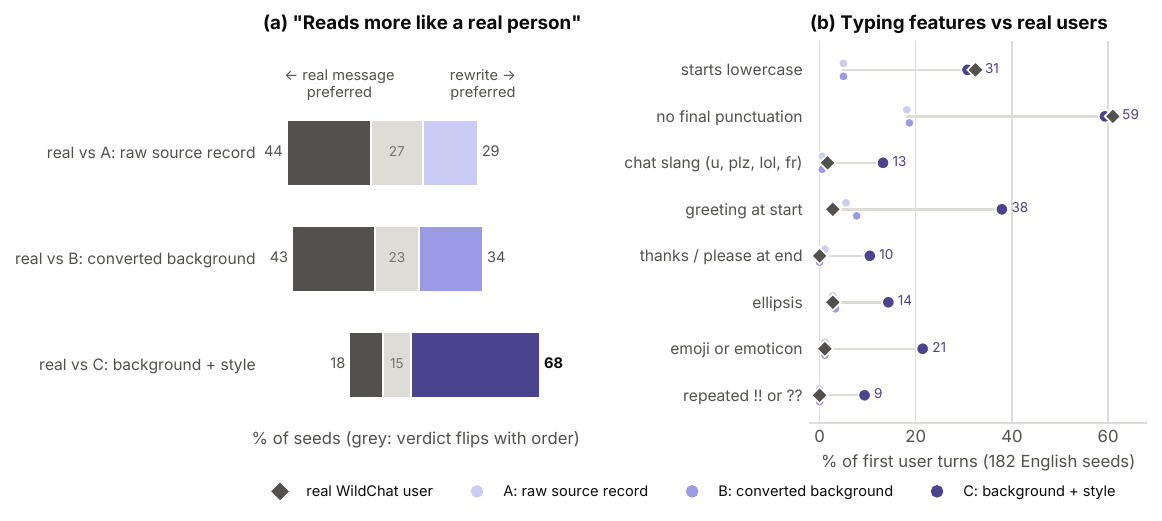}
\caption{Style layer calibrated against real users. (a) \texttt{MiniMax-M2.5} verdicts, real WildChat message against its rewrite per persona input, both orders (grey: verdict flips). (b) Typing features of the first user turn on the 182 English seeds.}
\label{fig:style-calibration}
\end{figure*}

\paragraph{Calibration against real users.} To see whether the style layer moves rewrites towards real users or past them, we convert 2{,}140 Nemotron-Personas and PersonaHub records into our format and rewrite the same seed turns three ways, changing only the persona text that the rewriter and responder see: (A) the raw source record, (B) our converted background without a style layer, and (C) the background with the generated style layer. On the 182 English WildChat seeds that pass in all three arms, we then ask three judges which of two messages reads more like something a real person typed, pairing each rewrite with the real WildChat message it came from. Every pair is judged in both orders; a win requires the same choice in both, and a flip counts as a draw. Without a style layer, the judges pick the real message more often than the rewrite. With it, every judge prefers the rewrite over the real message (Table~\ref{tab:style-calibration}). Figure~\ref{fig:style-calibration} shows why: the style layer brings lowercase starts and missing final punctuation to the rates of real users, but realizes greetings, chat slang, emoji, ellipses and repeated punctuation several times more often than real users do. The same markers are over-represented in a sample of the released corpus, so a judge preference for a rewrite over the original should be read as more casual typing, not as closer realism. The marker cap above addresses this; calibrating marker rates exactly to real-user frequencies remains a target for future versions of the pipeline.

\begin{table}[ht]
  \caption{\textbf{Real WildChat message (R) against its rewrite}, both presentation orders (\% of seeds; win = same choice in both orders, draw = the choice flips). A: raw source record; B: background without style layer; C: background with style layer.}
  \label{tab:style-calibration}
  \centering
  \small
  \resizebox{\linewidth}{!}{%
  \begin{tabular}{l|ccc|ccc|ccc}
    \toprule
    & \multicolumn{3}{c|}{\texttt{MiniMax-M2.5}} & \multicolumn{3}{c|}{\texttt{Qwen3.6-35B-A3B}} & \multicolumn{3}{c}{\texttt{Muse-Glimmer-30B}} \\
    Pair & R wins & Draw & Rewrite wins & R wins & Draw & Rewrite wins & R wins & Draw & Rewrite wins \\
    \midrule
    R vs.\ A & 44.1 & 27.1 & 28.8 & 53.3 & 36.3 & 10.4 & 68.7 & 22.0 & 9.3 \\
    R vs.\ B & 43.4 & 22.9 & 33.7 & 47.8 & 34.6 & 17.6 & 65.4 & 20.3 & 14.3 \\
    R vs.\ C & 17.6 & 14.8 & 67.6 & 16.5 & 41.8 & 41.8 & 27.5 & 21.4 & 51.1 \\
    \bottomrule
  \end{tabular}}
\end{table}

\paragraph{Matched content and length.}\label{app:matched}

We rendered the full content of each structured block as SynthPAI-style free text with the same facts and matched
length per persona, and compared both on the 800 items of Table~\ref{tab:formats-main}. With content fixed, pairwise
judges see no difference in style (structured share 50.2 to 51.3\%). The structured form keeps the request more often
(+5.0 [2.1, 7.6] with \texttt{Gemma-4-31B}) and reads more naturally (+5.6 and +6.0 with the two rewriters), while its
single-rewrite style adoption is slightly lower under one judge ($-$5.5). The comparison above therefore mixes structure
with content, since the SynthPAI format carries less style information. Our structural claim rests on
Table~\ref{tab:flat_vs_structured}, where only the representation of the style layer changes at equal length, and it
holds under the marker cap.

\paragraph{Scope.} This compares representations, not systems. Every condition renders our
own profiles, so the result says how a persona is encoded affects augmentation, not that
one persona bank is better than another. A consequence worth noting is that no PersonaHub
or SynthPAI content enters the released data.


\begin{table}[t]
\centering\small
\caption{\textbf{What each persona representation encodes} (\% of profiles, except words). Share of profiles whose attribute value survives into the block the rewriter sees ($N=555$ profiles).}
\label{tab:formats}
\begin{tabular}{l|rrrr}
\toprule
 & No persona & PersonaHub form & SynthPAI form & PersonaFactory \\
\midrule
Words in block & 0 & 13 & 83 & 280 \\
\midrule
Age & 0 & 0 & 100 & 100 \\
Location & 0 & 16 & 100 & 100 \\
Occupation & 0 & 68 & 100 & 100 \\
Language & 0 & 8 & 26 & 99 \\
Style axes & 0 & 0 & 55 & 100 \\
Quirks & 0 & 0 & 1 & 100 \\
Reference pool & 0 & 2 & 1 & 100 \\
\bottomrule
\end{tabular}
\end{table}

\begin{table}[t]
\centering\small
\caption{\textbf{Augmentation quality by persona representation.} Rates in \%; adherence, entity and literal scores in $[0,1]$. $N=800$ paired items, Gemma-4-31B rewriter, MiniMax-M2.5 judge; adherence metrics are judge-free. Brackets: 95\% Wilson intervals.}
\label{tab:formatresults}%
\begin{tabular}{l|rrrr}
\toprule
 & No persona & PersonaHub form & SynthPAI form & PersonaFactory \\
\midrule
\multicolumn{5}{l}{\emph{Style, the intended effect}} \\
Style adoption & 3.1 [2.1, 4.6] & 5.5 [4.1, 7.3] & 43.5 [40.1, 47.0] & 69.2 [66.0, 72.3] \\
Punctuation adherence & 0.405 & 0.410 & 0.645 & 0.757 \\
Emoji adherence & 0.814 & 0.815 & 0.939 & 0.980 \\
Capitalization adherence & 0.802 & 0.808 & 0.707 & 0.722 \\
\midrule
\multicolumn{5}{l}{\emph{Cost}} \\
Intent preserved (all) & 99.8 [99.1, 99.9] & 99.8 [99.1, 99.9] & 98.0 [96.8, 98.8] & 96.1 [94.6, 97.3] \\
\quad English & 99.8 & 100.0 & 99.8 & 99.3 \\
\quad Cross-lingual & 99.6 & 99.2 & 94.0 & 89.2 \\
Entities preserved & 0.946 & 0.936 & 0.921 & 0.905 \\
Quoted literals intact & 0.967 & 0.955 & 0.945 & 0.938 \\
Assistant sanity & 92.8 & 92.4 & 91.6 & 92.4 \\
\midrule
\multicolumn{5}{l}{\emph{Pairwise preference against PersonaFactory}} \\
PF preferred & 94.4 [92.5, 95.8] & 94.9 [93.1, 96.2] & 81.3 [78.4, 83.9] & -- \\
\quad win ratio & 17.90:1 & 20.46:1 & 4.83:1 & -- \\
\quad worst case & 94.0 & 94.6 & 76.8 & -- \\
\bottomrule
\end{tabular}
\end{table}

\begin{table}[t]
\centering\small
\caption{\textbf{Robustness of the ranking} (PF preferred, \%; win ratio in parentheses). Top: the same comparison with Ministral-3-14B as rewriter. Bottom: the pairwise comparison re-judged by four models on a 200-item subset.}
\label{tab:formatrobust}
\begin{tabular}{l|rrr}
\toprule
 & No persona & PersonaHub form & SynthPAI form \\
\midrule
\multicolumn{4}{l}{\emph{PF preferred, by rewriter}} \\
Gemma-4-31B & 94.4 (17.90:1) & 94.9 (20.46:1) & 81.3 (4.83:1) \\
Ministral-3-14B & 92.3 (12.70:1) & 92.6 (13.55:1) & 72.3 (2.64:1) \\
\midrule
\multicolumn{4}{l}{\emph{PF preferred, by judge (Gemma-4-31B rewriter)}} \\
Gemma-4-31B & 99.0 & 97.5 & 83.5 \\
GLM-4.7-Flash & 84.5 & 81.5 & 71.0 \\
MiniMax-M2.5 & 93.9 & 95.4 & 82.5 \\
Ministral-3-14B & 91.0 & 88.5 & 71.5 \\
\bottomrule
\end{tabular}
\end{table}

\FloatBarrier

\section{Human Study}
\label{app:human}


In this section we describe the expert human study that complements the LLM-judge evaluation of
Section~\ref{sec:eval}. We first describe the study design (\S\ref{app:human:description}), then give the
information and instructions shown to participants (\S\ref{app:human:instructions}), the ethics review and
compensation (\S\ref{app:human:irb}), the interim results (\S\ref{app:human:results}), and example items
(\S\ref{app:human:examples}).

\subsection{Study Description}
\label{app:human:description}

The study uses one form per language, 60 languages in total, each administered as a self-paced online
session. A form contains up to 40 items, ten in each of four parts; forms for languages with fewer available
samples contain fewer items. Participants are experts in the AI field who rate only the language varieties
they know well. They may complete up to three sessions, one per language, and are identified only by a random
participant ID. The four parts measure different properties and are never combined into a single score:

\begin{itemize}[leftmargin=1.2em,noitemsep,topsep=2pt]
  \item \textbf{Part A, meaning preservation.} An original user message is shown next to its persona rewrite
  from the corpus, and participants rate whether the rewrite preserves the meaning (intent and
  content) of the original. For cross-lingual rewrites the original is the English source message.
  \item \textbf{Part B, style fidelity.} A persona's free-text chatting-style description from the persona
  bank is shown next to a message rewritten for that persona, and participants rate how well the message
  matches the described style without being an exaggerated stereotype or caricature.
  \item \textbf{Part C, human or model.} Participants judge from the text alone whether a message was written
  by a real person or generated by a language model, and state their confidence. Each form mixes persona
  rewrites from the corpus with messages intended to be human-written user requests from a public
  chat-log corpus \citep{zhao2024wildchat}, in equal proportion; the origin of each text is not disclosed.
  The answer options follow the wording of the SynthPAI human study \citep{yukhymenko2024synthpai}.
  \item \textbf{Part D, linguistic plausibility.} Participants rate whether a short user message is plausible
  as something a real speaker of the language variety would write, and may report the item for personal
  information or offensive content.
\end{itemize}

Parts A, B and D and the Part C confidence use five-point scales. All displayed texts come from public research
datasets or machine-generated rewrites and were filtered for personal information and disturbing content
before the study. Any item can be skipped or reported, and reported items are removed.

\subsection{Information and Instructions for Participants}
\label{app:human:instructions}

We reproduce the study information and instructions as presented to participants, omitting only the
investigator names, institution, ethics application number, and contact details.

\subsubsection*{Study information}

\textit{Purpose.} This study evaluates a research pipeline that rewrites the user side of chat conversations
so that language-model training data better reflects how linguistically diverse people write (informal
register, second-language phrasing, code-switching) while preserving meaning. As an expert in the AI field,
you judge properties of displayed texts. The study analyzes texts and models, not you; no analyses of
individual participants are performed.

\textit{What participation involves.} Each item shows one or two short chat messages with a rating question
(meaning preservation, style fidelity, human-vs-LLM origin judgment, or linguistic plausibility). Some tasks
ask you to judge whether a displayed text was written by a human or generated by a language model; the origin
of each text is therefore not disclosed before you answer. This task structure is openly described here and
involves no other withheld information. This session contains 40 study items. Completion time varies with
reading pace and breaks. You may complete up to 3 sessions, one per language.

\textit{Risks.} Minimal, comparable to everyday reading and web use. Texts come from public research datasets
or machine-generated rewrites and are filtered for personal information and disturbing content; residually
mildly offensive text cannot be fully excluded. You can skip or report any item; reported items are removed.

\textit{Voluntariness and withdrawal.} Participation is entirely voluntary and unpaid, in your professional
capacity. Persons in a dependency relationship to the investigators may not participate. You may skip, pause,
or withdraw at any point without disadvantage. Deletion of your ratings is possible for as long as they can be
linked to your participant ID; after anonymization at study completion it is no longer possible.

\textit{Data protection.} Ratings, timestamps, and optional coarse language background are stored
pseudonymously under a random participant ID. No names, contact details, or IP addresses are collected. Data
are stored on institutional infrastructure with restricted access; consent records are retained securely and
separately. Only aggregated or anonymized results are published.

\subsubsection*{Consent declaration}

Before any study item is shown, participants confirm each of the following statements:
\begin{itemize}[leftmargin=1.2em,noitemsep,topsep=2pt]
  \item I have read and understood the study information above and had the opportunity to ask questions.
  \item I am at least 18 years old.
  \item I understand that participation is voluntary and unpaid, and that I may skip items or withdraw at any
  time without giving reasons and without disadvantage.
  \item I understand how my data will be processed, stored, and published, as described above.
  \item I understand that some tasks involve judging whether a displayed text was written by a human or by a
  language model, and that the origin of each text is not disclosed to me before I answer.
  \item I agree to take part in this study.
\end{itemize}

\subsubsection*{Task instructions}

\textbf{Part A, meaning preservation.} ``You will see an ORIGINAL text and a REWRITE. Judge whether the rewrite
preserves the meaning (intent and content) of the original. You may skip any item.'' Question: \textit{Does the
rewrite preserve the meaning of the original?} Scale 1 (\textit{meaning changed}) to 5 (\textit{meaning fully
preserved}).

\textbf{Part B, style fidelity.} ``You will see a short style description and a message. Judge how well the
message matches the described style WITHOUT being an exaggerated stereotype or caricature.'' Question:
\textit{How well does the message match the described style, without caricature?} Scale 1 (\textit{poor match /
stereotyped}) to 5 (\textit{faithful and plausible}).

\textbf{Part C, human or model.} ``For each message, judge from the text alone whether it was written by a real
person or generated by a language model. The origin of each text is not disclosed; some texts are
model-generated, as described in the study information.'' Question: \textit{Was this text written by a real human
being?}
\begin{enumerate}[leftmargin=1.6em,noitemsep,topsep=2pt]
  \item Yes: this text appears as authentic and readable as typical messages written by real users.
  \item No: parts of this text seem unusual, such as odd wording or synthetic text patterns, suggesting it might
  be AI-generated.
  \item Skip this item.
\end{enumerate}
Follow-up: \textit{How confident are you?} Scale 1 (\textit{guessing}) to 5 (\textit{very confident}).

\textbf{Part D, linguistic plausibility.} ``Only rate items in language varieties you know well. Judge whether
the message is plausible as something a real speaker of that variety would write, not an exaggerated
caricature. Report any item that seems to contain personal information or offensive content.'' Question:
\textit{Is this plausible as something a real speaker would write?} Scale 1 (\textit{implausible / caricature})
to 5 (\textit{fully plausible}); report options: \textit{No issue}, \textit{Report: personal information},
\textit{Report: offensive content}.

\subsection{IRB Approval and Compensation}
\label{app:human:irb}

The study was reviewed and approved by the ethics review board (IRB) of the authors' institution and
classified as minimal risk. All participants were informed about the study and its data processing and gave
informed consent through the declaration above before any item was shown; they could skip items, pause, or
withdraw at any time without giving reasons. Participation was voluntary and unpaid: participants are domain
experts who took part in their professional capacity, and persons in a dependency relationship with the
investigators were not eligible. No names, contact details, or IP addresses were collected, ratings are stored
under random participant IDs, and only aggregated or anonymized results are reported.

\subsection{Results}
\label{app:human:results}

We report an interim snapshot of the study. It contains 11 submissions covering nine languages with 476
assigned item evaluations; only German and Vietnamese currently have two submissions each. All saved
submissions are included without exclusions. Pooled statistics weight available ratings equally, not
languages equally, and standard deviations describe dispersion, not uncertainty in the mean.
Table~\ref{tab:human-pooled} gives pooled ratings, Table~\ref{tab:human-lang} the per-language means,
Table~\ref{tab:human-origin} the Part C judgments, and Table~\ref{tab:human-agree} the agreement between the two
German and the two Vietnamese submissions. Each language is rated on its own panel of items, so per-language differences can reflect
item selection and rater severity as well as language quality; these means are not a language ranking. In Part C, a text counts as a verified model rewrite if it matches a user message of the corpus exactly; the remaining texts are human-written user turns sampled from WildChat (the 44 verified judgments cover 34 distinct texts). The agreement estimates rest on few
shared items and are exploratory, and agreement does not imply accuracy.

\begin{table}[ht]
\centering
\caption{\textbf{Pooled five-point ratings} (1 to 5, higher is better; last column in \%). At the interim cutoff. Means weight observed ratings equally; the last
column is the share of ratings of 1 or 2.}
\label{tab:human-pooled}
\small
\begin{tabular}{l|rrrrr}
\toprule
Dimension & $n$ & Mean & SD & Median & Ratings 1--2 \\
\midrule
A: Meaning preservation & 89 & 3.94 & 1.30 & 4.0 & 15.7 \\
B: Style fidelity & 151 & 3.63 & 1.19 & 4.0 & 16.6 \\
D: Linguistic plausibility & 147 & 3.64 & 1.35 & 4.0 & 20.4 \\
C: Confidence & 85 & 3.85 & 0.88 & 4.0 & 8.2 \\
\bottomrule
\end{tabular}
\end{table}

\begin{table}[ht]
\centering
\caption{\textbf{Mean five-point ratings per language} (1 to 5). Number of available ratings in parentheses. Dashes mark
languages without Part A ratings.}
\label{tab:human-lang}
\small
\begin{tabular}{l|rrrr}
\toprule
Language & Subm. & Meaning ($n$) & Style ($n$) & Plausibility ($n$) \\
\midrule
Bosnian & 1 & -- (0) & 4.00 (20) & 3.40 (20) \\
Bulgarian & 1 & 3.50 (10) & 3.71 (17) & 3.81 (16) \\
Croatian & 1 & 3.11 (9) & 3.13 (15) & 3.64 (14) \\
English & 1 & 4.60 (10) & 4.60 (10) & 4.80 (10) \\
German & 2 & 4.60 (20) & 4.30 (20) & 4.50 (20) \\
Greek & 1 & 2.30 (10) & 1.94 (18) & 2.00 (18) \\
Serbian & 1 & -- (0) & 3.20 (15) & 3.80 (15) \\
Ukrainian & 1 & 4.70 (10) & 4.60 (10) & 4.70 (10) \\
Vietnamese & 2 & 4.00 (20) & 3.73 (26) & 3.21 (24) \\
\bottomrule
\end{tabular}
\end{table}

\begin{table}[ht]
\centering
\caption{\textbf{Part C judgments} (counts; share of verified rewrites in \%). A text is a verified model rewrite if it matches a user message of the
corpus exactly; the remaining texts are human-written user turns sampled from WildChat; per-language counts are in Appendix~\ref{app:partc}.}
\label{tab:human-origin}
\small
\begin{tabular}{l|rrrr}
\toprule
Source status & Judgments & Human-like & AI-like & Skip \\
\midrule
Verified model rewrite & 44 & 26 (59.1) & 18 (40.9) & 0 \\
WildChat (human) & 44 & 29 & 12 & 3 \\
\midrule
All Part C & 88 & 55 & 30 & 3 \\
\bottomrule
\end{tabular}
\end{table}

\paragraph{LLM judges on Part C.} We also give the Part C texts of all answered panels (68 texts: 34 verified
rewrites and 34 chat-log texts) to four LLM judges, \texttt{Claude Sonnet 4.5}, \texttt{MiniMax-M2.5},
\texttt{Qwen3.6-35B-A3B} and \texttt{Muse-Glimmer-30B}. Every judge sees the question and answer options shown to
the raters and returns a JSON answer (human or model) with a confidence from 1 to 5; LLM judges answer each text three times. The three open-weight judges
are served locally (\texttt{MiniMax-M2.5} with SGLang, the other two with vLLM) with their default chat templates
and reasoning, and receive one text per request through the chat API; each text is sampled three times at
temperature 0.7 with up to 8{,}192 new tokens, and the answer is read from the final JSON object.
\texttt{Claude Sonnet 4.5} is queried in three independent sessions without tool access, each receiving all 68 texts
in one prompt under neutral ids in a shuffled order and judging each text independently. It is run through the
Claude Code agent interface, whose sampling parameters are fixed by the provider and not user-configurable. All 816 answers parse. We do not
use \texttt{Gemma-4-31B} as a judge because it wrote part of the rewrites. For accuracy, the chat-log texts
count as human-written by study design, although they are not individually verified; the human-rater accuracy
uses the same convention over the 85 answered judgments. Table~\ref{tab:partc-judges} reports
the results. All judges label most rewrites as human-written (83.3 to 89.2\%), more often than the raters
(59.1\%), and none separates the two sources clearly above chance (50.5 to 57.4\% accuracy).

\begin{table}[ht]
\centering
\caption{\textbf{Human-or-model judgments by humans and LLM judges} (\%). Part C, 68 texts (34 verified rewrites, 34 chat-log texts). Accuracy counts a rewrite labeled model-generated or a chat-log text labeled human-written as correct; 95\% Wilson intervals in brackets.}
\label{tab:partc-judges}
\small
\begin{tabular}{l|cc}
\toprule
Judge & Rewrites judged human & Accuracy \\
\midrule
Human raters & 59.1 [44.4, 72.3] & 55.3 [44.7, 65.4] \\
\texttt{Claude Sonnet 4.5} & 83.3 [74.9, 89.3] & 57.4 [50.5, 63.9] \\
\texttt{MiniMax-M2.5} & 89.2 [81.7, 93.9] & 50.5 [43.7, 57.3] \\
\texttt{Qwen3.6-35B-A3B} & 88.2 [80.6, 93.1] & 52.9 [46.1, 59.7] \\
\texttt{Muse-Glimmer-30B} & 83.3 [74.9, 89.3] & 54.9 [48.0, 61.6] \\
\bottomrule
\end{tabular}
\end{table}

\begin{table}[ht]
\centering
\caption{\textbf{Agreement between the two submissions of a language} (exact and within-1 in \%). On shared items. Numeric ratings use
quadratic-weighted Cohen's $\kappa$; the human/model choice uses unweighted $\kappa$.}
\label{tab:human-agree}
\small
\begin{tabular}{ll|rrrr}
\toprule
Language & Dimension & Items & Exact & Within 1 & $\kappa$ \\
\midrule
German & A: Meaning preservation & 10 & 60.0 & 90.0 & 0.077 \\
German & B: Style fidelity & 10 & 40.0 & 100.0 & 0.455 \\
German & D: Linguistic plausibility & 10 & 60.0 & 80.0 & 0.405 \\
German & C: Human/model choice & 10 & 80.0 & -- & 0.583 \\
Vietnamese & A: Meaning preservation & 10 & 60.0 & 90.0 & 0.583 \\
Vietnamese & B: Style fidelity & 13 & 38.5 & 84.6 & 0.493 \\
Vietnamese & D: Linguistic plausibility & 12 & 50.0 & 75.0 & 0.545 \\
Vietnamese & C: Human/model choice & 10 & 60.0 & -- & 0.200 \\
\bottomrule
\end{tabular}
\end{table}

\paragraph{Findings.} Meaning preservation, style fidelity and plausibility are rated favorably overall (pooled
medians of 4), and most strongly for English, German and Ukrainian; no English or Ukrainian rating in Parts A, B
or D falls below 3. The Greek panel is the clear exception: five of ten meaning ratings and eleven of
eighteen style ratings are 1, which marks its samples for review, although the evidence comes from a single
submission. Quality also varies by dimension within a language: Vietnamese meaning preservation (4.00) is rated
higher than plausibility (3.21), and Croatian is rated lower on meaning and style (3.11 and 3.13). On texts
verified as model rewrites, raters selected the human-written option in 59.1\% of judgments. Because the
\textit{Yes} option also describes the text as authentic and readable, this is evidence of perceived
human-likeness, not a measure of detection accuracy or of indistinguishability from human writing; overall
a fully verified accuracy requires a human-source key for the remaining texts (Table~\ref{tab:partc-judges} uses the design labels). The Ukrainian panel was evaluated on the
revised selections of Parts A and B (informal and task-diverse rewrites), which replaced the earlier selection
before any Ukrainian submission; no before-after comparison is therefore possible. Part D received five
personal-information reports (four Bosnian, one Serbian), which we review manually rather than treat as confirmed
privacy issues.

\paragraph{Limitations of the interim study.} Most languages currently have a single submission, and panel size,
item selection, source language and rater severity confound cross-language comparisons, so we report no
population-level confidence intervals or significance tests. The study has no baseline arm, so it does not show
an improvement over plain translation or another persona method. Collection is ongoing: we are adding raters for
the same items, particularly for low-scoring panels, and completing the human-source key for Part C.

\subsection{Part C per Language and the Greek Panel}\label{app:partc}

Part C covers eight of the study's nine languages (the Bosnian submission has no Part C answers), with one or two
raters each. Skipped items are excluded; the human-written texts are WildChat user turns.
\begin{center}\small
\begin{tabular}{lcccc}
\toprule
Language & Raters & Ours judged human & WildChat judged human & Confidence (ours / WildChat) \\
\midrule
Bulgarian & 1 & 1/3 & 3/3 & 3.33 / 3.33 \\
Croatian & 1 & 5/5 & 4/4 & 4.00 / 4.50 \\
English & 1 & 3/5 & 3/5 & 4.00 / 3.20 \\
German & 2 & 5/10 & 7/10 & 3.90 / 3.80 \\
Greek & 1 & 0/1 & 1/1 & 5.00 / 3.00 \\
Serbian & 1 & 4/5 & 1/3 & 4.00 / 3.67 \\
Ukrainian & 1 & 3/5 & 3/5 & 3.40 / 3.40 \\
Vietnamese & 2 & 5/10 & 7/10 & 3.90 / 4.40 \\
\midrule
All & 10 & 26/44 (59\%) & 29/41 (71\%) & 3.86 / 3.83 \\
\bottomrule
\end{tabular}
\end{center}
With one or two raters per language, these rates are descriptive only. The low Greek ratings reflect a few items with
weak Greek generation: items without such defects score far higher (mean rating 3.88, against 1.25 to 1.86 for items
with garbled Greek, stray non-Greek tokens, invented cultural markers or an unfollowed style description). We list this
as a language-specific limitation, tied to the generator's capacity in that language.

\subsection{Scope of the Human Study in Context}\label{app:human:scope}

The study follows an IRB-approved protocol classified as minimal risk (\S\ref{app:human:irb}). Its instrument
was prepared for 60 languages and has four parts (A meaning preservation, B style fidelity, C human or model with
confidence, D linguistic plausibility); raters are unpaid volunteer experts in the AI field who rate only language
varieties they know well, give informed consent, and are identified by random participant IDs.

At the time of writing, the study has 11 submissions in nine languages (Bosnian, Bulgarian, Croatian, English,
German, Greek, Serbian, Ukrainian, Vietnamese) with 476 assigned item evaluations; German and Vietnamese have two
submissions each, the others one. The texts use three scripts (Latin, Cyrillic, Greek), and the languages come from
the Slavic, Germanic and Hellenic branches of Indo-European and from Austroasiatic. Only English and German are
high-resource; the others are mid- or lower-resource. Related human evaluations are compared below (IRB: ``no''
means not mentioned, ``n/a'' means no external raters).

\begin{center}
\small
\begin{tabularx}{\linewidth}{@{}>{\raggedright\arraybackslash}p{3.7cm}>{\raggedright\arraybackslash}p{2.7cm}>{\raggedright\arraybackslash}Xc@{}}
\toprule
Work & Languages & Raters & IRB \\
\midrule
\multicolumn{4}{@{}l}{\textit{Human studies of synthetic personas}} \\
PersonaHub \citep{chan2024personahub} & English & none for personas; 2 math experts checked 200 generated problems & n/a \\
SynthPAI \citep{yukhymenko2024synthpai} & English & 40 paid (Clickworker); 2 per comment on 1{,}000 comments & yes \\
PERSONA \citep{castricato2025persona} & English & 80 (Prolific) & no \\
Synthetic-Persona-Chat \citep{jandaghi2024faithful} & English & 3 per conversation pair (Mechanical Turk) & no \\
Persona Policies \citep{chopra2026personapolicies} & English & 20 (Prolific), 16 after filtering; 87 conversations & yes \\
PersonaLens \citep{zhao2025personalens} & English & 3 researchers; 100 dialogues & no \\
\midrule
\multicolumn{4}{@{}l}{\textit{Multilingual persona datasets}} \\
Nemotron-Personas \citep{meyer2025nemotronpersonas} & one or two per country release & none reported & n/a \\
Mental-health dialogues \citep{xu2026multilingualmh} & 4 & none (LLM judges only) & n/a \\
Wiki-Talkie \citep{fucci2026wikitalkie} & 5 (European) & no external raters; one author audited 10 users per language & n/a \\
IndicTalk \citep{gawande2026indictalk} & 9 (Indic) & 2 to 3 volunteer graduate students; 10 conversations per variety & no \\
SEADialogues \citep{alkautsar2025seadialogues} & 8 (Southeast Asian), human results for 4 & 3 paid per language & no \\
\midrule
Ours & 9 (3 scripts) & 1 to 2 volunteer experts per language & yes \\
\bottomrule
\end{tabularx}
\end{center}

Human studies of synthetic personas with larger rater pools are English-only and recruit paid workers on
crowdsourcing platforms (Clickworker, Prolific, Mechanical Turk). Multilingual persona datasets either report no
evaluation by external raters or evaluate languages of a single region. Among the synthetic-persona works we
reviewed, ours is the only human study that covers languages of more than one region, and the only multilingual
one that reports ethics approval.

A clear weakness remains: seven of the nine languages have a single rater, while IndicTalk and SEADialogues use two
or three, so our per-language results are indicative only. We did not recruit more raters because
native speakers of most of these languages are scarce on crowdsourcing platforms, and several paid native raters
per language exceed the budget and time frame of this work.

The claims of the paper therefore rest on the automatic evaluation: the strict intent label and its human check
(Appendices~\ref{app:strict} and~\ref{app:humancheck}), external judges (Appendix~\ref{app:judges}), judge-free
metrics (Table~\ref{tab:fidelity} and Appendix~\ref{app:refrows}), detector analyses (Appendix~\ref{app:turing})
and downstream training (Appendix~\ref{app:downstream}). The human study is supporting evidence;
Table~\ref{tab:human-origin} gives its Part C results and Appendix~\ref{app:partc} gives them per language.

\FloatBarrier

\subsection{Example Items}
\label{app:human:examples}

We show example items from the English form. Part C items are omitted because their origin labels are part of the
ongoing study.

\begin{promptbox}[Part A: meaning preservation]
\textbf{Original:} What role can technology play in facilitating Urdu language learning outside of South Asia?\\
\textbf{Rewrite:} hep i was wondering what role technology can play in helping people learn Urdu outside south asia\\[4pt]
\textbf{Original:} Write a Python program that takes two lists as input and determines if the first list is a subset
of the second list. The program should output either ``True'' or ``False'', depending on whether the first list is a
subset of the second list. Ensure that your program uses appropriate variable names, follows PEP 8 guidelines for
style, and includes relevant comments to explain the logic of the code.\\
\textbf{Rewrite:} can u write a python program that takes two lists as input and determines if the first list is a
subset of the second list? it just needs to output ``True'' or ``False'' depending on whether it is a subset :)
please make sure it uses appropriate variable names, follows pep 8 guidelines for style, and includes relevant
comments to explain the logic of the code.
\end{promptbox}

\begin{promptbox}[Part B: style fidelity]
\textbf{Style description:} Writes in all lowercase with minimal punctuation and occasional technical shorthand,
reflecting a fast-paced digital native style.\\
\textbf{Message:} can you do a deep dive review on that new ergo keyboard you used / focus on the features and how
the comfort holds up during long sessions / need info on the adjustments and customization options and how they
improve the experience / also talk about the build quality and form factor plus any extras like wrist rests or
touchpads / give some examples of specific tasks you did and how it performed / make it detailed enough to show the
strengths and limits so people know if it's worth the investment\\[4pt]
\textbf{Style description:} She types like she is writing a formal letter, using full capitalization and ending
every sentence with a period, often adding polite pleasantries.\\
\textbf{Message:} Hello. I have been thinking about how populations continue to grow even after birth rates slow
down, and I came across the term demographic momentum. It would be lovely if you could explain how demographic
momentum works in current global trends, especially with changes in technology and healthcare. Could you please
provide some examples of where it is having a big impact? Thank you.
\end{promptbox}
{\small Slashes mark line breaks in the original message.}

\begin{promptbox}[Part D: linguistic plausibility]
can u check why this script is throwing a timeout error? looks like a network glitch machan\\[4pt]
bro, my bike brake is making noise again. is there any way to fix it with basic tools?
\end{promptbox}

\FloatBarrier

\section{Per-attribute Augmentation Analysis}
\label{app:attribution}

Aggregate fidelity tells us \emph{that} persona conditioning works, but not \emph{which} part of a profile is doing the work or what it costs. We therefore run a controlled, one-attribute-at-a-time ablation that decomposes the augmentation into three separately measured aspects. To our knowledge, prior user-simulation and persona-conditioning work reports fidelity only in aggregate; a per-attribute decomposition of a persona-augmentation pipeline is, to our knowledge, new.

\begin{figure}[ht!]
  \centering
  \includegraphics[width=0.72\textwidth]{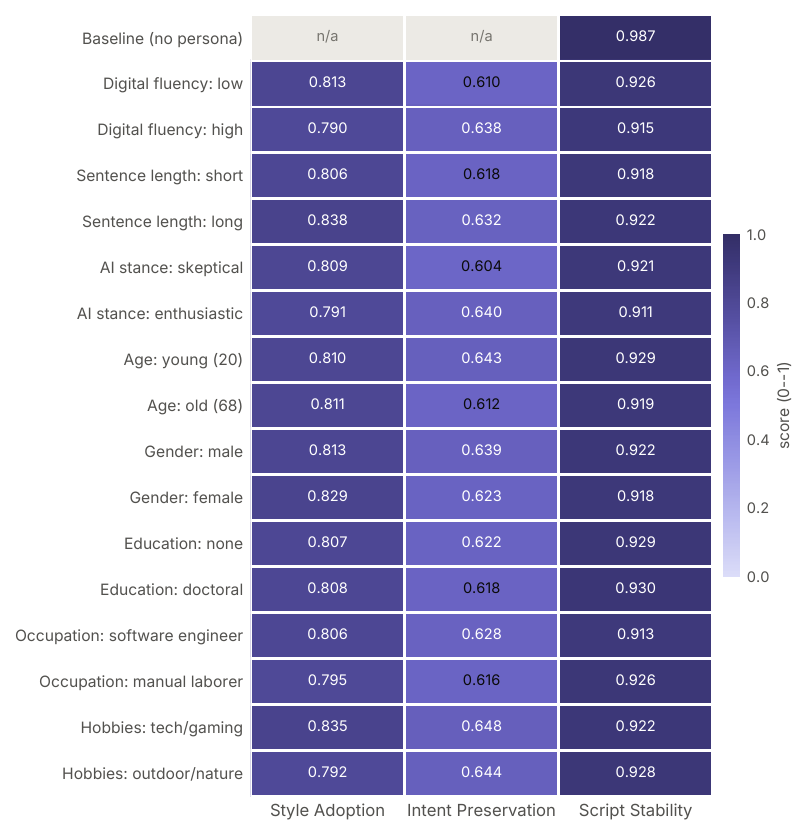}
  \caption{Every single-attribute condition scored on the three fidelity axes, averaged over the four rewriters, on one shared $0$--$1$ scale. The no-persona baseline has no rewrite, hence no style or intent score.}
  \label{fig:abl_heatmap}
\end{figure}

\paragraph{Setup.} From a stratified pool of up to 194 real source messages, we build \emph{single-attribute} personas: a no-persona baseline plus sixteen conditions that vary exactly one attribute and collapse into eight families (digital fluency, sentence length, AI stance, age, gender, education, occupation, hobbies). Each condition is rewritten by all four generators (\texttt{Gemma-4-31B}, \texttt{Ministral-3-14B}, \texttt{GLM-4.7-Flash}, \texttt{MiniMax-M2.5}), holding everything else fixed, for 12{,}845 valid rewrites. Every rewrite is scored on three orthogonal aspects: \emph{style adoption} (does the turn take on the attribute's register?), \emph{intent preservation} (is it still the same request in the same language?), and rule-based \emph{script stability}; the no-persona baseline fixes the script-stability ceiling at 0.987. Table~\ref{tab:attribution} reports the eight families, averaged over the four generators; Figure~\ref{fig:abl_heatmap} gives the full per-condition grid behind it.

\begin{table}[ht]
  \caption{\textbf{Per-attribute augmentation analysis} (share of rewrites, higher is better). One persona attribute varied at a time, averaged over four generators; script stability is relative to the no-persona baseline.}
  \label{tab:attribution}
  \centering
  \small
  \begin{tabular}{l|ccc}
    \toprule
    Persona attribute & Style adoption & Intent preservation & Script stability \\
    \midrule
    Digital fluency  & 0.801 & 0.624 & 0.921 \\
    Sentence length  & 0.822 & 0.625 & 0.920 \\
    AI stance        & 0.800 & 0.622 & 0.916 \\
    Age              & 0.810 & 0.627 & 0.924 \\
    Gender           & 0.821 & 0.631 & 0.920 \\
    Education        & 0.808 & 0.620 & 0.929 \\
    Occupation       & 0.801 & 0.622 & 0.920 \\
    Hobbies          & 0.814 & 0.646 & 0.925 \\
    \midrule
    Baseline (no persona) & n/a & n/a & 0.987 \\
    \bottomrule
  \end{tabular}
\end{table}

\paragraph{Reading the decomposition.} We present these numbers as a measurement, not an importance ranking. Across all eight families, style adoption sits in a narrow 0.80--0.82 band, intent preservation clusters near 0.62, and script stability decrements uniformly by roughly 0.06--0.07 from the no-persona baseline. We deliberately avoid ordering attributes by effect size: the between-attribute spread (at most $\approx$0.02 in style adoption) is smaller than the cross-generator and sampling variation, so no single attribute is separably ``more important.'' In the full per-condition grid (Figure~\ref{fig:abl_heatmap}), which puts all three axes on one $0$--$1$ scale, the columns sit at clearly different levels while the rows within a column barely move. The informative signal is the \emph{uniformity} itself, and it supports a negative statement only: within the precision of this probe, no background attribute is separably more influential than another, and conditioning on any of them costs a small, roughly constant script-stability decrement. It does not show that each attribute is individually effective, and none of the structured style keys is ablated here.

\begin{figure}[ht!]
  \centering
  \includegraphics[width=\textwidth]{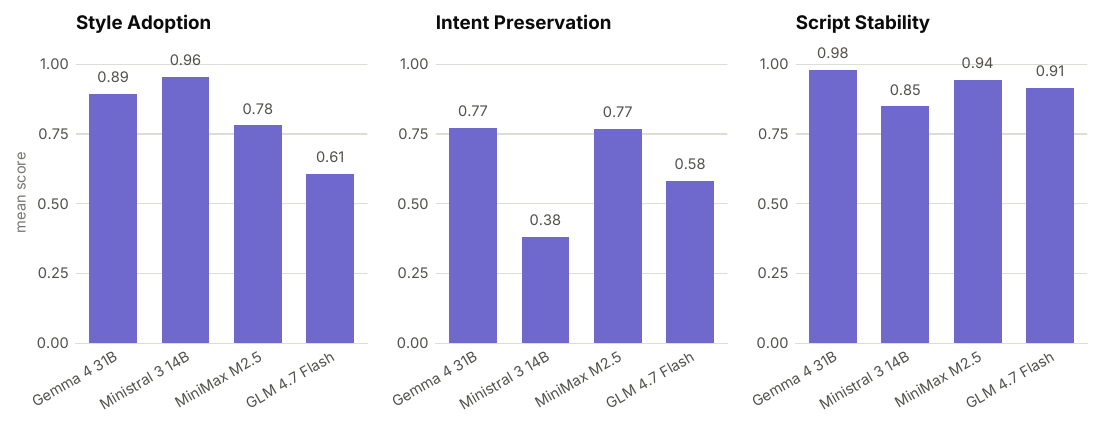}
  \caption{The same ablation summarized per rewriter, averaged over all sixteen attribute conditions.}
  \label{fig:abl_models}
\end{figure}

\paragraph{What does vary: the generator.} The variation that dominates is the rewriter, not the attribute, as Figure~\ref{fig:abl_models} makes plain: style adoption spans barely $0.02$ across the eight attribute families but $0.35$ across the four rewriters. On identical conditions it ranges from $\approx$0.61 (GLM) to $\approx$0.96 (Ministral), and it trades off against intent preservation (Ministral $\approx$0.38 vs.\ Gemma $\approx$0.77 and MiniMax $\approx$0.77), consistent with more aggressive restyling costing more intent. Ministral also has the lowest script stability ($0.85$ against Gemma's $0.98$), so the three panels tell one story: the most eager restyler is also the least faithful one, while Gemma is the balanced choice, which is the same responder gap that shows up at corpus scale in Section~\ref{sec:corpus}. This is exactly why the main-text comparison (Section~\ref{sec:eval}) fixes the rewriter to isolate style \emph{representation}, and why we report multi-judge agreement rather than trusting a single evaluator. The ablation here is a controlled probe on a stratified sample; it complements, rather than replaces, the corpus-scale metrics of Section~\ref{sec:corpus}.

\FloatBarrier

\section{Conversational Corpus}
\label{app:corpus}

\subsection{Stylized Corpus Details}
\label{app:corpusdet}

\begin{figure*}[ht]
\centering
\includegraphics[width=0.96\textwidth]{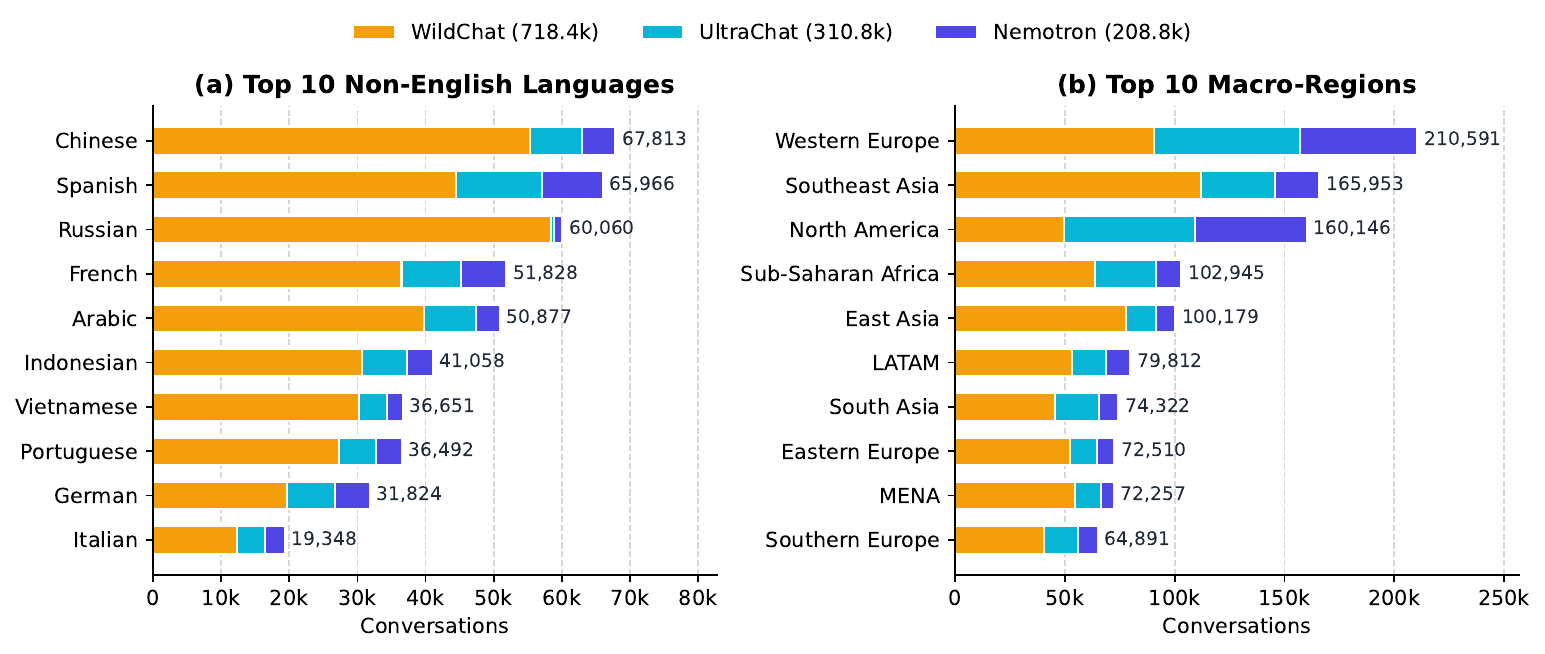}
\caption{Distribution of the released chat dataset: (a) top 10 non-English languages (detected language of the first user turn), and (b) top 10 persona macro-regions.}
\label{fig:app_lang_region_dist}
\end{figure*}

\paragraph{What the corpus looks like.} English is the largest language of the released corpus, with a long multilingual tail. Of the $\approx$1.24M conversations, 472{,}297 (38.2\%) start with an English user turn. The rest are spread over 84 other languages, led by Chinese (67.8k conversations), Spanish (66.0k), Russian (60.1k) and French (51.8k). All 13 macro-regions of the persona sampling grid appear, but not evenly: Western Europe (210.6k), Southeast Asia (166.0k), North America (160.1k) and Sub-Saharan Africa (102.9k) account for the largest shares, which is the skew we discuss in Section~\ref{sec:limitations}. This shape is not an artefact of one seed dataset. WildChat supplies most conversations in every top language, but UltraChat and Nemotron contribute to every language and region and supply most of the Western European, North American and Oceanian conversations. Figure~\ref{fig:app_lang_region_dist} shows both breakdowns.

\paragraph{When the pipeline leaves a turn alone.} About 94\% of released user turns are rewritten in the persona's voice. The remaining 6\% (146{,}643 turns) are kept unchanged on purpose. We spell out why, because a turn that was deliberately protected looks the same from the outside as one where the rewriter failed. There are five reasons (the remaining 0.2\% are rewriter calls that timed out):
\begin{itemize}[leftmargin=*,itemsep=1pt,topsep=2pt]
    \item \textbf{Truncation} (21.6\%, 31{,}716 turns): the rewrite came back much shorter than the original, which usually means content was lost rather than restyled.
    \item \textbf{Strict-format tasks} (14.4\%, 21{,}116): verbatim repetition, translation or quoted-output requests, where restyling would break exactly what the user asked for.
    \item \textbf{Lost protected content} (31.0\%, 45{,}453): every retry dropped a protected quote, code span or number.
    \item \textbf{Script or language change} (29.1\%, 42{,}607): the rewrite switched the script or language of the turn, or the persona's language did not match the script of the prompt.
    \item \textbf{No change} (3.7\%, 5{,}460): the rewriter returned the message as it was.
\end{itemize}
In each of these cases we prefer to ship the original turn over a broken request or a language mismatch.

One number may look surprisingly low: code-switching markers appear in only 1.95\% of user turns (measured on the 552{,}654 user turns of conversations with a model persona), although many more personas are described as code-switchers. This is intended. Linguistic quirks are applied with a fixed probability rather than at every opportunity, because a persona that uses every mannerism in every message reads as a caricature rather than a person (Section~\ref{sec:pipeline}). Each quirk is meant to be rare in a single turn and visible across the corpus.

\paragraph{How personas get matched to conversations.} Each seed conversation goes through a chain of matching strategies, and the first one that finds a compatible persona is used.
\begin{itemize}[leftmargin=*,itemsep=1pt,topsep=2pt]
    \item \textbf{Location} (288{,}196 conversations, 23.3\%): the prompt names a place. Capitalized tokens and names after prepositions (e.g.\ ``in Berlin'') are extracted with regular expressions and matched against each persona's sub-region, country, native language and dialect; prompts in non-Latin scripts are also mapped to the countries and languages that use that script.
    \item \textbf{Semantic embedding} (395{,}698, 32.0\%): the first user message and each persona's background and style are encoded with \texttt{all-MiniLM-L6-v2}, and the most similar persona is chosen if its cosine similarity is at least 0.15. This pairs a prompt with a persona whose occupation or interests fit the topic.
    \item \textbf{Language pool} (3{,}075, 0.2\%): a fallback that only guarantees a persona who speaks the prompt's language.
    \item \textbf{Cross-lingual} (551{,}007, 44.5\%): the separate cross-lingual mode, which ignores the prompt's language and samples a persona from the whole bank. It produces the cross-lingual 44\% of the corpus; without it, the language-matched set would stay almost entirely English.
\end{itemize}

\paragraph{What survives the sanity gate.} Of the 1{,}606{,}837 conversations that reached the sanity gate, 1{,}267{,}306 (78.9\%) pass every sanity check after the final assistant regeneration. After deduplication, a final filter on low-resource languages, a cap on cross-lingual Russian and a rule-based filter that removes 4{,}785 conversations in which a user turn restates the instructions of the rewrite prompt, 1{,}237{,}976 remain; these form the $\approx$1.24M release we recommend for training. At the turn level, 93.3\% of user turns pass (measured on the conversations with a model persona). Table~\ref{tab:app_sanity_breakdown} breaks the pass rate down, and two patterns stand out.

The first is the responder model. \texttt{Gemma-4-31B} passes 89.2\% of its conversations and \texttt{Ministral-3-14B} 68.0\%, so the smaller model accounts for about 74\% of all dropped conversations. The seed source matters much less at this stage (UltraChat 82.1\%, Nemotron 80.2\%, WildChat 77.2\%). The gap therefore reflects the responder's capacity, not a problem with persona conditioning itself.

The second pattern is smaller. Cross-lingual conversations pass less often than language-matched ones (75.2\% against 82.3\%), as expected when the user writes in a language the responder handles less confidently. We keep both parts in the release rather than quietly dropping the harder one.
\begin{table}[t]
  \caption{\textbf{Sanity pass rates by slice} (\%). Conversation-level sanity pass of the final assistant regeneration over all 1{,}606{,}837 conversations that reached the gate; responder = the model that generated the conversation.}
  \label{tab:app_sanity_breakdown}
  \centering
  \small
  \begin{tabular}{ll|rrr}
    \toprule
    \textbf{Dimension} & \textbf{Slice} & \textbf{Total} & \textbf{Sanity Pass} & \textbf{Pass Rate} \\
    \midrule
    \multirow{2}{*}{Responder Model} 
      & Gemma-4-31B & 823{,}676 & 734{,}927 & 89.2 \\
      & Ministral-3-14B & 783{,}161 & 532{,}379 & 68.0 \\
    \midrule
    \multirow{3}{*}{Seed Source} 
      & WildChat & 959{,}084 & 740{,}485 & 77.2 \\
      & UltraChat-200k & 384{,}191 & 315{,}508 & 82.1 \\
      & Nemotron-Instruction-Following & 263{,}562 & 211{,}313 & 80.2 \\
    \midrule
    \multirow{2}{*}{Persona Rewrite Type} 
      & Language-Matched & 838{,}034 & 689{,}348 & 82.3 \\
      & Cross-Lingual & 768{,}803 & 577{,}958 & 75.2 \\
    \midrule
    \textbf{Full Dataset} & \textbf{Overall} & \textbf{1{,}606{,}837} & \textbf{1{,}267{,}306} & \textbf{78.9} \\
    \bottomrule
  \end{tabular}
\end{table}

\paragraph{Dialect labels.}\label{app:dialects} Personas carry a free-text dialect label: the whole bank has 117{,}670 distinct labels, and the personas used in the conversations carry 81{,}388 distinct labels after case and whitespace normalization, so the same dialect can be phrased differently across personas. We group them into canonical varieties per language and keep a variety only when it belongs to the language of the conversation, which leaves 13{,}714 varieties: 1{,}639 have at least 100 user turns, 416 have at least 1{,}000, and 695 are carried by at least 100 distinct personas. Frequent labels include regional varieties such as Northern English, New Zealand English and Swiss German. These counts describe the personas that condition the rewrites; verifying how faithfully each dialect is realized in text would require native speakers for each variety, which we leave to future work.

\paragraph{Personal information.}\label{app:privacy} We checked 20{,}000 WildChat-seeded rewrites with Presidio \citep{microsoft2024presidio}. Rewrites do not add contact details: none of the 10{,}550 same-language pairs introduces an email address, phone number or card number that is absent from its seed. Details already present in a WildChat seed can carry over (pattern-detected entities in 12.2\% of seeds and 11.3\% of rewrites), so this part of the corpus inherits WildChat's own anonymization.

\FloatBarrier

\subsection{Assistant-Side Model Personas}
\label{app:modelpersona}

This appendix documents the assistant-side model personas of Section~\ref{sec:modelpersona}: the style set, how styles are matched and used, the data-level pilot that motivated their use, and a downstream fine-tuning ablation. All judged numbers come from a single LLM judge (\texttt{MiniMax-M2.5}) and are proxies.

\paragraph{Style set.} Each of the 12 styles has a free-text description of how answers are presented, eight presentation axes (register, structure, density, warmth, explanation, jargon, certainty, emoji), two or three optional signature touches, and content safeguards (e.g.\ never drop a requested detail for brevity, never agree with an incorrect premise). Two rules hold for every style: keep every fact, step, number and requested detail the answer needs, and never invent facts, sources or statistics to fit the style. Table~\ref{tab:mp-styles} lists the styles by their one-line summary together with their share of the release. Style names are used only for bookkeeping and never appear in a prompt. In a 1{,}000-conversation pilot, removing them from the prompt left style adoption (68.8\% without names, 66.4\% with) and judged fit (47.6 to 50.8 pairwise) unchanged within noise, and it removes any chance of a label leaking into the replies: 52 released conversations in which a reply named its own style were regenerated without names.

\begin{table}[ht]
  \caption{\textbf{The 12 assistant response styles} (share in \%), ordered by their share of the released corpus ($n = 234{,}723$ conversations).}
  \label{tab:mp-styles}
  \centering\footnotesize
  \setlength{\tabcolsep}{3pt}
  \begin{tabular}{p{4.9cm}p{2.1cm}p{3.1cm}rr}
    \toprule
    Summary & Register & Structure & Convs. & Share \\
    \midrule
    Talks like a knowledgeable friend: relaxed, contractions, light humour. & casual & conversational prose & 52{,}450 & 22.3 \\
    Answers first, in compact expert prose, with no preamble or sign-off. & neutral-professional & prose & 48{,}802 & 20.8 \\
    Explains through analogies and vivid concrete examples in flowing prose. & conversational & flowing prose & 24{,}872 & 10.6 \\
    Maximally compact layout: bullets, fragments allowed, zero filler. & neutral & tight bullets or one short paragraph & 21{,}621 & 9.2 \\
    Separates facts from judgement and makes assumptions explicit. & neutral-professional & labelled parts (established, assumptions, how to check) & 20{,}731 & 8.8 \\
    Formal written register with full sentences and a concluding summary. & formal & paragraphs with topic sentences & 16{,}795 & 7.2 \\
    Organises every answer into headed sections, numbered steps and code first. & neutral & short headers, numbered steps, tables where comparisons appear & 16{,}575 & 7.1 \\
    Short sentences, everyday words and one concrete example. & neutral & short paragraphs, occasional simple bullets & 8{,}659 & 3.7 \\
    Walks through things step by step in plain words, explaining each term once. & friendly-neutral & short paragraphs or numbered steps & 8{,}026 & 3.4 \\
    Energetic, tech-savvy and informal, with snappy bullets. & casual & snappy bullets and short code snippets & 7{,}628 & 3.2 \\
    Supportive, action-oriented and second person, without flattery. & friendly & short paragraphs and action lists & 5{,}029 & 2.1 \\
    Gentle, steady tone suited to stressful or personal topics. & gentle & short prose paragraphs & 3{,}535 & 1.5 \\
    \bottomrule
  \end{tabular}
\end{table}

\paragraph{Matching and generation.} The matcher sees a reduced user card with the persona's AI stance, digital fluency, sentence-length tendency, question framing, code-switching frequency and structured typing keys, together with the first user message; occupation, gender, age, location, languages and the free-text chatting style (which often names a language or region) are withheld, so the choice cannot key on demographics. Candidate styles are shown in a per-conversation random order, and the top-ranked style is used for the whole conversation. During assistant regeneration, the style section is appended to the paper's persona system prompt on every turn, with an explicit priority order: explicit user instructions about format, length, tone or language first, then the content rules, correctness and safety, and only then the style. A signature touch is offered with $p = 0.25$ per reply (25.0\% of released assistant turns), and the continuity adaptation step keeps the style. As with the user persona, neither the style nor its matching is stored in the training example. The match distribution is skewed (Table~\ref{tab:mp-styles}): the two most frequent styles cover 43\% of conversations and the least frequent 1.5\%.

\paragraph{Design pilots.} Before settling on the release setup, we tested three design questions on 604 conversations (1{,}489 assistant turns) with an earlier version of the style set without signature touches (Table~\ref{tab:mp-design}). Content agreement is measured against the reply generated with the user persona only; instruction compliance is measured on the 94 later turns with an explicit format or length instruction. First, \emph{matching carries signal}: in a pairwise judgment of which reply suits the user better, the matched style beats a random style 60.0 to 39.9. Second, \emph{generating with the style beats restyling afterwards on adoption}, while restyling an existing reply keeps its content almost perfectly; we generate with the style because the release regenerates the assistant side anyway, and keep restyling as an option for data whose answers must not change. Third, instruction compliance on later turns is unaffected in every arm within the precision of this sample. As in the main pilot, the matched style ties the user-persona-only reply on judged fit (49.7 to 50.1).

\begin{table}[ht]
  \caption{\textbf{Model persona design pilot} (\%, higher is better; bold: best per column). 604 conversations; content agreement vs.\ the user-persona-only reply; compliance on 94 turns with explicit instructions. 95\% intervals at most $\pm$2.5 (per-turn) and $\pm$10 points (compliance).}
  \label{tab:mp-design}
  \centering\small
  \begin{tabular}{l|cccc}
    \toprule
    Assistant side & Style adopt. & Same content & Instr.\ compliance & Sanity pass \\
    \midrule
    Regenerated, user persona only & 35.1 & 44.6 & 84.0 & 88.2 \\
    Matched style, generated & \textbf{75.7} & 37.6 & 87.2 & 87.7 \\
    Random style, generated & 53.3 & 35.8 & \textbf{89.2} & 87.7 \\
    Matched style, restyled afterwards & 67.1 & \textbf{96.9} & 88.2 & \textbf{92.3} \\
    \bottomrule
  \end{tabular}
\end{table}

\paragraph{Data-level pilot.} On 1{,}000 conversations with fixed user turns, we regenerate the assistant side four ways with the same responder per conversation and compare them (Table~\ref{tab:mp-pilot}). Style adoption asks whether a reply is recognisably presented in the conversation's matched style; for the default and background-only arms this measures how often the reply happens to fit that style anyway. Content agreement compares each reply with the default reply to the same turn; the noise floor, two independent default samples, is 59.6\%. Instruction compliance is measured on later turns whose user message contains an explicit format or length instruction.

\begin{table}[ht]
  \caption{\textbf{Data-level model persona pilot} (\%, except median words). 1{,}000 conversations, about 2{,}500 assistant turns, \texttt{MiniMax-M2.5} judge; 95\% intervals at most $\pm$1.9 points for per-turn rates.}
  \label{tab:mp-pilot}
  \centering\small
  \begin{tabular}{l|cccc}
    \toprule
    Assistant conditioning & Style adopt. & Same content & Instr.\ compliance & Median words \\
    \midrule
    None (default) & 19.4 & (ref.) & 86.1 & 521 \\
    User persona (paper) & 25.4 & 36.9 & 87.6 & 266 \\
    Model persona only & 71.1 & 40.0 & 86.4 & 313 \\
    User persona + model persona & 66.8 & 35.9 & 87.3 & 271 \\
    \bottomrule
  \end{tabular}
\end{table}

The model persona makes the intended style clearly recognizable without reducing instruction compliance, and conditioning on the user persona shortens replies to about half the default length. In a pairwise judgment of which reply suits the user better, shown the reduced user card, the combination of user persona and model persona beats the default 61 to 39 and the model persona alone 58 to 41, but ties the user persona alone (49 to 51). The model persona therefore adds a controllable assistant style rather than a better fit to the user.

\paragraph{Downstream fine-tuning ablation.} We regenerate the assistant side of the corpus three ways with identical user turns: (A) no assistant system prompt, (B) the paper's user-persona system prompt, and (C) the user persona plus a matched model persona, as in the release. We fine-tune \texttt{Llama-3.1-8B} (full fine-tuning, one epoch, learning rate $10^{-5}$, global batch 128) and \texttt{Apertus-8B-2509} \citep{swissai2025apertus} (base model, same budget, packed multi-turn sequences with loss on every assistant turn) on the same 100k conversations per arm. For evaluation, 300 held-out requests are each rewritten by 12 user personas that span digital fluency, AI stance, capitalization, message length and typos, so the request is fixed and only the user's voice changes; replies are decoded greedily. We measure within-request correlations between a user-message feature and the matching reply feature, style diversity as the mean pairwise distance of the 12 replies to one request in a content-independent style embedding \citep{wegmann2022style}, content stability as LaBSE similarity to the reply given to the unstyled request, and a pairwise judge of which reply is presented better for the user, with a length-controlled (LC) win rate estimated at equal reply length by logistic regression on the log length ratio.

\begin{table}[ht]
  \caption{\textbf{Adaptation to the user's style after fine-tuning.} Correlations are within-request (user-message feature to reply feature); diversity and stability as described in the text. Arms A to C as in the notation; IT is the instruction-tuned release.}
  \label{tab:mp-sft-adapt}
  \centering\small
  \setlength{\tabcolsep}{4pt}
  \resizebox{\linewidth}{!}{%
  \begin{tabular}{ll|cccc|cc}
    \toprule
    & & \multicolumn{4}{c|}{Within-request correlation} & & \\
    \cmidrule(lr){3-6}
    Family & Stage & formality & lowercase$\to$formality & length & emoji & style diversity & content stability \\
    \midrule
    \textbf{Llama-3.1-8B} & IT & 0.18 & $-$0.16 & $-$0.06 & 0.12 & 0.254 & 0.866 \\
    \cmidrule(lr){1-8}
    & SFT A (no conditioning) & 0.51 & $-$0.45 & $-$0.03 & 0.40 & 0.187 & 0.827 \\
    & SFT B (user persona) & 0.37 & $-$0.33 & 0.11 & 0.40 & 0.293 & 0.775 \\
    & SFT C (+ model persona) & 0.43 & $-$0.41 & 0.26 & 0.49 & 0.313 & 0.777 \\
    \midrule
    \textbf{Apertus-8B} & SFT A (no conditioning) & 0.49 & $-$0.43 & $-$0.01 & 0.28 & 0.185 & 0.824 \\
    & SFT B (user persona) & 0.45 & $-$0.38 & 0.04 & 0.29 & 0.380 & 0.819 \\
    & SFT C (+ model persona) & 0.45 & $-$0.37 & 0.05 & 0.38 & 0.275 & 0.818 \\
    \bottomrule
  \end{tabular}}
\end{table}

\begin{table}[ht]
  \caption{\textbf{Judge-free reply features after fine-tuning} (headers, bullets and emoji: \% of replies). Replies of the fine-tuned models to the style probes.}
  \label{tab:mp-sft-features}
  \centering\small
  \resizebox{\linewidth}{!}{%
  \begin{tabular}{ll|cccccc}
    \toprule
    Family & Stage & Median words & Headers & Bullets & Emoji & Formality & Contractions / 100 words \\
    \midrule
    \textbf{Llama-3.1-8B} & IT & 426 & 6.5 & 67.9 & 0.1 & 0.71 & 1.75 \\
    \cmidrule(lr){1-8}
    & SFT A & 497 & 71.5 & 71.9 & 17.4 & 0.64 & 0.92 \\
    & SFT B & 195 & 12.9 & 43.0 & 1.4 & 0.77 & 0.47 \\
    & SFT C & 210 & 8.2 & 45.6 & 2.0 & 0.73 & 0.95 \\
    \midrule
    \textbf{Apertus-8B} & SFT A & 666 & 75.1 & 70.9 & 4.4 & 0.68 & 0.71 \\
    & SFT B & 293 & 10.2 & 47.5 & 0.7 & 0.74 & 0.52 \\
    & SFT C & 299 & 7.4 & 39.7 & 1.1 & 0.73 & 0.96 \\
    \bottomrule
  \end{tabular}}
\end{table}

\begin{table}[ht]
  \caption{\textbf{Pairwise judge after fine-tuning} (win rate of the first model, \%). Raw win rate with 95\% confidence interval and length-controlled (LC) win rate; about 3{,}500 decisive comparisons per pair.}
  \label{tab:mp-sft-judge}
  \centering\small
  \begin{tabular}{l|cc}
    \toprule
    Comparison & Raw win & LC win \\
    \midrule
    Llama B vs.\ A & 57.0 (55.4--58.7) & 48.8 \\
    Llama C vs.\ A & 54.9 (53.3--56.6) & 42.4 \\
    Llama C vs.\ B & 48.1 (46.4--49.8) & 48.6 \\
    Llama C vs.\ \texttt{Llama-3.1-8B-Instruct} & 79.4 (78.1--80.7) & 68.6 \\
    Apertus B vs.\ A & 66.3 (64.7--67.8) & 52.1 \\
    Apertus C vs.\ A & 57.1 (55.5--58.7) & 44.8 \\
    Apertus C vs.\ B & 42.0 (40.3--43.6) & 42.6 \\
    \bottomrule
  \end{tabular}
\end{table}

\begin{figure}[ht]
\centering
\includegraphics[width=0.72\linewidth]{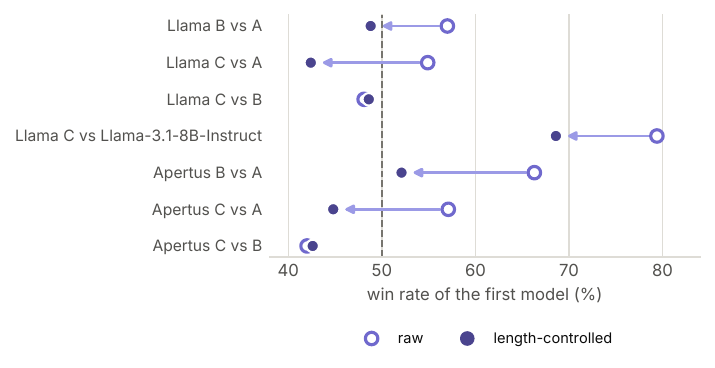}
\caption{Pairwise win rates after fine-tuning, raw (open) and length-controlled (filled).}
\label{fig:mp-sft-lc}
\end{figure}

Three observations follow (Tables~\ref{tab:mp-sft-adapt}, \ref{tab:mp-sft-features} and~\ref{tab:mp-sft-judge}, Figure~\ref{fig:mp-sft-lc}). Formality is the mean score of a formality classifier \citep{babakov2023formality}. First, every model fine-tuned on persona-styled conversations tracks the user's formality and emoji use far more than \texttt{Llama-3.1-8B-Instruct}, in both model families, and is clearly preferred over it. Second, without assistant conditioning (A) the models inherit the teacher's defaults: replies two to two and a half times longer, headers in about three quarters of them, and the lowest style diversity. Conditioning the assistant (B, C) yields shorter, less formatted and more varied replies that the judge prefers, but the preference disappears or reverses once length is controlled; since B and C replies are about half as long, the equal-length estimate is an extrapolation. Third, the model persona does not beat the user-persona prompt on judged fit (a tie for Llama, B preferred for Apertus); its benefit is the controllable style variation of Table~\ref{tab:mp-pilot}, and downstream Llama C adapts length and emoji most and is the most diverse, while Apertus C is less diverse than Apertus B. These results use one seed per arm, a single judge and no capability benchmarks, and the fine-tuning data share the user side, so they compare assistant conditioning only.

\paragraph{Conversations without a user persona.} Model personas do not depend on the user side, so they can also restyle the assistant side of an existing dataset. In this mode, the matcher sees only the user's own messages (the first three), and the assistant turns are regenerated with the style section as the only system prompt, with the same continuity loop and sanity gate. On 1{,}000 source conversations (UltraChat and Nemotron, 2{,}570 assistant turns), matching succeeds for every conversation, and Table~\ref{tab:mp-nouser} compares the variants. Regeneration with the style raises style adoption from 25.7\% to 79.5\% at a cost of about five points of sanity pass; restyling the source replies afterwards keeps their content but reaches lower adoption and falls back to the original reply on 35\% of turns because a protected span was lost.

\begin{table}[ht]
  \caption{\textbf{Model personas without a user persona} (\%, except median words; bold: best per column). 1{,}000 source conversations; content agreement of the first assistant turn vs.\ the source reply. 95\% intervals at most $\pm$1.9 (style) and $\pm$3.1 points (other rates).}
  \label{tab:mp-nouser}
  \centering\small
  \begin{tabular}{l|cccc}
    \toprule
    Assistant side & Sanity pass & Style adopt. & Same content & Median words \\
    \midrule
    Source replies & n/a & 17.4 & (ref.) & 344 \\
    Regenerated, no style & \textbf{82.2} & 25.7 & 42.4 & 583 \\
    Regenerated with style & 77.4 & \textbf{79.5} & 25.4 & 371 \\
    Source replies restyled & n/a & 57.5 & \textbf{93.0} & 417 \\
    \bottomrule
  \end{tabular}
\end{table}

A limitation of this mode appears when the source conversation carries its own system prompt. On 200 single-turn conversations with a checkable system constraint (at most three sentences, a numbered list, under 80 words, or no Markdown), compliance is 86\% without a style, but 58\% with the style placed below the system prompt and 66\% with the system prompt placed last as an explicit override. The style competes with explicit format and length constraints even when told to yield, so conversations with a system prompt are left unaugmented.

\subsection{Downstream Fine-tuning on the Released Corpus}
\label{app:downstream}

The ablation above varies only the assistant side. Here we ask what the released conversations, with their persona-styled user turns, do for a model trained on them, and which part of the pipeline is responsible.

\paragraph{Setup.} We fine-tune \texttt{Apertus-v1.5-8B} base \citep{apertus2026apertus15opentext} for one epoch on four versions of the same 79{,}817 language-matched conversations from the training split, sampled with 5\% of the personas held out for evaluation. The arms differ only in their turns: (A0) the original seed conversation; (A0r) the original user turns with the assistant turns regenerated by the same responder, with no persona, the same continuity loop and the same sanity gate; (A1) the persona-styled user turns with the original assistant turns; and (A2) the released conversation, with styled user turns and assistant turns regenerated for that user. Conversations whose A0r regeneration failed the sanity gate, or that exceed 3{,}800 tokens in any arm, are removed from all arms. Training uses Megatron-LM with packed 4{,}096-token sequences, loss on every assistant turn, global batch 128, and a learning rate of $10^{-5}$ decayed to $10^{-6}$ with a cosine schedule after 3\% warmup.

\paragraph{Evaluation.} We report clean capability with lm-evaluation-harness (thinking disabled), robustness as the drop from the clean prompt to a perturbed one on IFEval, GSM8K and MBPP, and a pairwise comparison on real user messages. Perturbations are either rule-based (lowercasing, punctuation removal, light and heavy typos, abbreviations, terse and combined variants) or persona rewrites of the benchmark prompt by three held-out personas (\texttt{Gemma-4-31B} rewriter, kept only if \texttt{MiniMax-M2.5} confirms that the task is unchanged, 99.6\% of rewrites). For the pairwise comparison, we sample 1{,}000 first user turns from WildChat seeds that no arm was trained on (500 English, 500 in other languages); two judges from outside the pipeline (\texttt{Qwen3.6-35B-A3B}, \texttt{Muse-Glimmer-30B}) judge each pair of replies in both orders, and we report a length-controlled win rate as in Appendix~\ref{app:modelpersona}. Intervals are 95\% paired bootstrap intervals over items.

\begin{table}[ht]
  \caption{\textbf{Clean capability after fine-tuning} \texttt{Apertus-v1.5-8B} base on each arm (\%, lm-evaluation-harness, thinking disabled). Mean over three training seeds (standard deviation at most 1.6 points). Median assistant-turn length of the training data in characters.}
  \label{tab:downstream-clean}
  \centering
  \small
  \resizebox{\linewidth}{!}{%
  \begin{tabular}{l|r|cccccc}
    \toprule
    Arm & Assistant chars & MMLU & ARC-C & GSM8K & IFEval strict & IFEval loose & MBPP \\
    \midrule
    A0, original & 1{,}131 & 63.7 & 71.4 & 70.1 & 67.9 & 69.7 & 48.7 \\
    A0r, regenerated, no persona & 2{,}618 & 61.3 & 73.0 & 67.7 & 65.1 & 68.7 & 48.5 \\
    A1, styled user, original assistant & 1{,}131 & 63.1 & 72.2 & 69.6 & 64.1 & 65.9 & 47.5 \\
    A2, released & 956 & 64.4 & 72.4 & 67.0 & 69.7 & 72.2 & 48.8 \\
    \bottomrule
  \end{tabular}}
\end{table}

\begin{table}[ht]
  \caption{\textbf{Paired comparisons between arms}, training seed 1234. IFEval strict accuracy scored with our IFEval checker, which also scores the perturbed prompts: difference on clean prompts and difference in the robustness gap (clean minus perturbed, positive: the first arm loses more), in points with 95\% intervals over items; length-controlled (LC) WildChat win rate of the first arm (\%) per judge.}
  \label{tab:downstream-paired}
  \centering
  \small
  \resizebox{\linewidth}{!}{%
  \begin{tabular}{l|ccc|cc}
    \toprule
    & \multicolumn{3}{c|}{IFEval} & \multicolumn{2}{c}{WildChat LC win rate} \\
    \cmidrule(lr){2-4}\cmidrule(lr){5-6}
    Comparison & Clean difference & Rule-based gap difference & Persona gap difference & Qwen3.6 & Muse \\
    \midrule
    A1 vs.\ A0 & $-$4.1 ($-$7.2, $-$0.9) & $-$1.0 ($-$3.9, 1.8) & 0.0 ($-$3.6, 3.7) & -- & -- \\
    A2 vs.\ A0r & 5.0 (1.1, 8.9) & 1.7 ($-$1.4, 4.8) & 2.5 ($-$1.9, 6.8) & 46.5 (42.9, 49.7) & 53.8 (49.8, 57.5) \\
    A2 vs.\ A1 & 6.7 (2.8, 10.4) & 1.3 ($-$1.5, 4.1) & 1.9 ($-$2.0, 5.9) & 57.8 (55.2, 60.6) & 59.5 (56.7, 62.3) \\
    A2 vs.\ A0 & 2.6 ($-$0.9, 6.1) & 0.3 ($-$2.4, 3.0) & 2.0 ($-$1.8, 5.9) & 54.6 (51.9, 57.3) & 55.5 (52.8, 58.3) \\
    \bottomrule
  \end{tabular}}
\end{table}

\paragraph{Results.} Table~\ref{tab:downstream-clean} and Table~\ref{tab:downstream-paired} show three findings. First, styled user turns help only together with assistant turns written for them. Training on styled user turns with the original replies (A1) lowers IFEval against the original data, while the released conversations (A2) score highest on IFEval and clearly beat A1. The A2 advantage over A1 on IFEval holds across three training seeds (7.2, 6.5 and 3.3 points with lm-evaluation-harness; 5.7 [1.9, 9.6] with intervals over items and seeds), and on real WildChat prompts both external judges prefer A2 replies over A1 replies at equal length for every seed (57.3 to 60.0\%). Second, the released conversations match or improve on the original data: against A0, A2 gains 1.8 points of IFEval strict accuracy on average over seeds and loses 3.1 points on GSM8K, the only benchmark where it is consistently behind, and both judges prefer A2 replies over A0 replies at equal length for every seed (54.6 to 57.5\%). Third, none of the arms measurably changes robustness to messy prompts: rule-based gaps stay small for every arm, and no gap difference between arms excludes zero. Persona rewrites of benchmark prompts cost every arm more than rule-based noise. Regenerating the assistant side without a persona (A0r) more than doubles the reply length, so the comparisons against A0r depend on the length correction, and the two judges disagree on A2 against A0r at equal length for every seed. Over three training seeds, A2 gains 4.6 points [1.1, 8.2] of IFEval strict accuracy over A0r and A1 loses 3.8 points [$-$6.7, $-$0.9] against A0 (lm-evaluation-harness, intervals over items and seeds). These results come from one base model.

\paragraph{Strict subset.} Retraining A2 only on conversations in which every user turn is labelled same request (Appendix~\ref{app:strict}), at equal size and with three seeds, leaves every benchmark unchanged (IFEval $+$0.3 [$-$2.2, 2.8]), and the preference over the original data remains (53.8 to 57.4\%).

\paragraph{Adaptation of reply style.} We sent 300 held-out requests from 12 user personas (three seeds) to models trained on our data (A2) and on the original data (A0). A2 adapts its reply style to the user more, while content stays as stable (Table~\ref{tab:downstream-adapt}). This controlled study is our main downstream evidence; the comparison with \texttt{Llama-3.1-8B-Instruct} in Appendix~\ref{app:modelpersona} is context only.

\begin{table}[ht]
  \caption{\textbf{Adaptation of reply style} after fine-tuning on the released (A2) and original (A0) conversations, 300 held-out requests from 12 user personas, three seeds.}
  \label{tab:downstream-adapt}
  \centering
  \small
  \begin{tabular}{l|cc}
    \toprule
    Metric & A2 & A0 \\
    \midrule
    Formality correlation & 0.463 & 0.378 \\
    Emoji mirroring & 0.392 & 0.125 \\
    Style diversity across personas & 0.325 & 0.227 \\
    Content stability & 0.826 & 0.829 \\
    \bottomrule
  \end{tabular}
\end{table}

\paragraph{Cross-lingual conversations.} We repeat the comparison for the cross-lingual setting with two arms built from the same English seeds and target personas: (X1) every user turn translated into the persona's language by the pipeline's persona-blind translation step, with the assistant turns regenerated without a persona as in A0r; and (X2) the released cross-lingual conversation, with styled user turns and assistant turns regenerated for that user. Conversations whose X1 translation or regeneration failed, or that exceed 3{,}800 tokens in either arm, are removed from both, leaving 40{,}528 conversations; training follows the setup above with one seed. Table~\ref{tab:downstream-x} adds multilingual benchmarks. X2 is ahead on most benchmarks, with the largest gain on multilingual instruction following (Multi-IF, 7.3 points), and behind on GSM8K. On WildChat prompts, X2 replies are half as long as X1 replies; both judges prefer X2 on raw votes, but at equal length Muse is at parity (52.1\%, interval 48.0 to 55.9) and Qwen3.6 prefers X1 (44.3\%, interval 40.7 to 47.9), so the open-ended comparison is inconclusive given the large length difference.

\begin{table}[ht]
  \caption{\textbf{Clean capability after fine-tuning on cross-lingual arms} (\%, lm-evaluation-harness, thinking disabled, one seed). X1: translated user turns with a regenerated assistant; X2: released conversations. Multi-IF: prompt-level strict accuracy; INCLUDE: 44 languages.}
  \label{tab:downstream-x}
  \centering
  \small
  \begin{tabular}{l|cc}
    \toprule
    Benchmark & X1 & X2 \\
    \midrule
    MMLU & 61.9 & 63.2 \\
    ARC-C & 70.0 & 71.5 \\
    GSM8K & 68.8 & 67.3 \\
    IFEval strict & 69.3 & 70.4 \\
    IFEval loose & 72.8 & 74.1 \\
    MBPP & 48.4 & 50.6 \\
    \midrule
    Multi-IF & 59.7 & 67.0 \\
    MGSM-en & 50.4 & 53.3 \\
    Global-MMLU & 56.6 & 56.6 \\
    INCLUDE & 56.1 & 57.1 \\
    \bottomrule
  \end{tabular}
\end{table}

\FloatBarrier

\end{document}